\documentclass[3p,times,final,preprint,review,12pt,authoryear]{elsarticle} 

\usepackage{amssymb}
\usepackage{amsmath}

 \usepackage{lineno}
 
\usepackage{algorithm}
\usepackage{algorithmic}
\usepackage{multirow}
\usepackage{verbatim}
\usepackage{microtype}
\usepackage{enumitem}
\usepackage{booktabs} 
\usepackage{color,graphicx}
\usepackage{comment}

\usepackage{amsmath,dsfont,bm,mathtools,amsthm,amssymb } 
\usepackage{thmtools, thm-restate}

\DeclareMathOperator*{\argmin}{argmin}

 \newtheorem{definition}{Definition}
 
\usepackage{multirow}
\usepackage{booktabs}
\usepackage{subcaption}
\usepackage{adjustbox}
\newcommand{\ouralgo}{$\texttt{p-FClus}$}

\newcommand{\kFed}{$\texttt{k}$-$\texttt{FED}$}
\newcommand{\MFC}{$\texttt{MFC}$}
\newcommand{\Clus}{$\texttt{CentClus}$}
\newcommand{\sota}{$\texttt{SOTA}$}

\usepackage{tikz}
\definecolor{babypink}{rgb}{0.96, 0.76, 0.76}
\definecolor{bubblegum}{rgb}{0.99, 0.76, 0.8}

\newcommand{\Mu}{\boldsymbol{\mu}}
\newcommand{\Var}{\boldsymbol{\sigma}}
\newcommand{\Max}{\boldsymbol{\max}}

\usepackage{cancel}

\newcommand{\INDSTATE}[1][1]{\STATE\hspace{#1\algorithmicindent}}

\newcommand{\nonl}{\renewcommand{\nl}{\let\nl\oldnl}}
\usetikzlibrary{tikzmark,calc}

\usepackage{multirow}

\usepackage{pgfplots}
\pgfplotsset{width=10cm, compat=1.9}
\usepackage{bm}
\usetikzlibrary{calc}

\journal{Journal}

\begin{document}

\begin{frontmatter}



\title{Fair Federated Data Clustering through Personalization: Bridging the Gap between Diverse Data Distributions}


\author[label1]{Shivam Gupta\corref{cor1} } 
\ead{shivam.20csz0004@iitrpr.ac.in}
\author[label1]{Tarushi} 
\ead{2020csb1135@iitrpr.ac.in}
\author[label1]{Tsering Wangzes}
\ead{2020csb1136@iitrpr.ac.in}
\author[label1]{Shweta Jain}
\ead{shwetajain@iitrpr.ac.in}
\affiliation[label1]{organization={Indian Institute of Technology Ropar},
        state={Punjab},
     country={India}}
\cortext[cor1]{Corresponding author}

        

\begin{abstract}
The rapid growth of data from edge devices has catalyzed the performance of machine learning algorithms. However, the data generated resides at client devices thus there are  majorly two challenge faced by traditional machine learning paradigms - centralization of data for training and secondly for most the generated data the class labels are missing and there is very poor incentives to clients to manually label their data owing to high cost and lack of expertise. To overcome these issues, there have been initial attempts to handle unlabelled data in a privacy preserving distributed manner using unsupervised federated data clustering. The goal is partition the data available on clients into $k$ partitions (called clusters) without actual exchange of data. Most of the existing algorithms are highly dependent on data distribution patterns across clients or are computationally expensive. Furthermore, due to presence of skewed nature of data across clients in most of practical scenarios existing models might result in clients suffering high clustering cost making them reluctant to participate in federated process. To this, we are first to introduce the idea of personalization in federated clustering. The goal is achieve balance between achieving lower clustering cost and at same time achieving uniform cost across clients. We propose \ouralgo\ that addresses these goal in a single round of communication between server and clients. We validate the efficacy of \ouralgo\ against variety of federated datasets showcasing it's data independence nature, applicability to any finite $\ell$-norm,  while simultaneously achieving lower cost and variance.
\end{abstract}

\begin{graphicalabstract}
\includegraphics[scale=0.155]{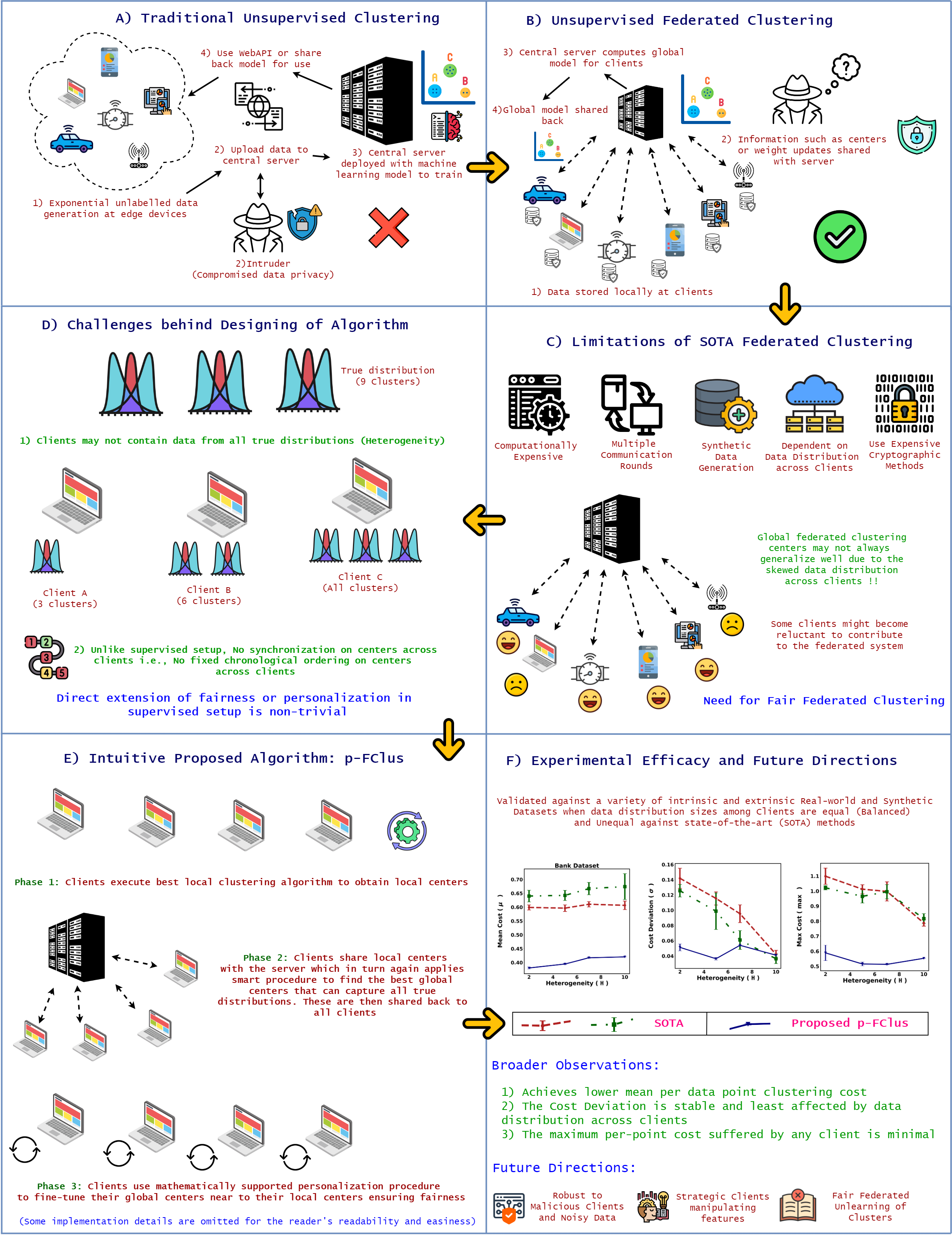}
\end{graphicalabstract}

\begin{highlights}

\item Unsupervised privacy-preserving federated clustering for increasing unlabelled data

\item Existing methods have high-cost deviations across clients thus they may leave system

\item Our method results in centers with lower deviation (i.e, first personalized solution)

\item We give  single round communication method independent of data division among clients
    
\item Works for any finite value of l-norm and undergoes rigorous experimental validation

\end{highlights}


\end{frontmatter}



\section{Introduction}
In the age of rapid technological advancement, the proliferation of devices, such as smartphones, tablets, Internet of Things (IoT) sensors, and edge devices, has become an inherent aspect of our daily lives (\cite{brecko2022federated}). This surge in technological devices has resulted in an exponential growth of data (\cite{manyika2011big}), a phenomenon that has significantly impacted many domains, especially machine learning (ML) (\cite{gupta2021epilnet,gupta9neural,ranga2020automated,ranga2022pathological}). The increasing data is catalysing the evolution and success of ML algorithms, as many algorithms perform better with more data (\cite{osisanwo2017supervised}). Furthermore, the abundance of data facilitates training more complex, robust and generalised models and underscores the notion that `data is the new oil'. However, it is essential to note that traditional machine learning methods necessitate centralised data collection for model training, typically on a server. Nonetheless, most generated data exists outside the servers and data centers, i.e., on edge devices. The data on these edge devices is private, thus sparking rising concerns from alliances, governments, and researchers about data privacy and protection during collection at centralised repositories (\cite{shen2022distributed}).

In response to this challenge, Federated Learning (FL) is emerging as a promising solution for collaborative  ML model training without centralized data collection. Pioneered by Google Inc. in 2016, FL revolutionizes the traditional model training process into a novel distributed paradigm (\cite{FedAvg}). In this approach, each participating client (or edge device) receives an initial  model from server trained on publicly available data. As more private data is generated at each client, they are tasked with training their respective models (referred to as local models) on this private data. Subsequently, instead of sharing the complete updated models or local data with the server, clients share model parameter updates. The server, after gathering updates from all clients\footnote{A few literature work opts for a subset of clients to improve throughput and efficiency}, aggregates and computes global model updates, which are then shared back with clients for future use. While the existing literature encompasses various aggregation policies, the simplest involves computing the weighted average of model parameters (called FedAvg \cite{FedAvg}) based on the local dataset size of each client. As communication involves solely sharing model updates rather than actual data, the risk of recovering the original data from these updates is minimized. Although a few works suggest that FedAvg encounters such issues, the literature offers numerous more robust, privacy-enhanced methods (\cite{li2020federated,li2021ditto,deng2020distributionally,hu2024fedmmd,wu2024fedel}). Consequently, one can assert that FL is emerging as a burgeoning paradigm for harnessing the hidden potentials inherent within the growing data landscape.

Within the field of FL, two prominent frameworks, namely Horizontal Federated Learning (HFL) and Vertical Federated Learning (VFL), have gathered significant attention over the past few years (\cite{yang2019federated}).  These frameworks differ in how data is split among participating clients.  In the HFL framework, data instances on client devices share the same set of attributes and labels (\cite{yang2019federated}). On the contrary,  in VFL, clients exhibit a larger overlap in data instances but a less similar set of attributes (feature set and labels) (\cite{yang2019federated}).  Although the literature extensively studies supervised federated learning, where the feature set and labels are available in data instances (\cite{fedmix,fedssc,osisanwo2017supervised}). However, in practical, real-life scenarios often face challenges,  as data available on the clients may lack labels (\cite{van2020towards}). A few possible factors include lack of motivation, incentives, or expertise to label their own data.  For instance, in scenarios like the clustering sentiments of social media posts,  clients might be reluctant to invest efforts in labelling their posts as happy or sad (\cite{muhammad2020sentiment,al2022building}). Furthermore, even if clients are willing to label their data, they might not fully label all local data, potentially due to a lack of expertise in labelling.  This challenge is evident in emerging trends like wearable fitness bands, where clients may lack accurate knowledge to label themselves as medically fit or unhealthy based on readings from different sensors (such as heart rate, blood oxygen level, and sugar level) (\cite{krishnan2022self}). This raises the key question: How can one find a global model in a federated environment when confronted with the presence of unlabelled data?

To answer this, a few initial attempts have been made to handle unlabelled data in various domains such as Speech Enhancement (\cite{tzinis2021separate}), Intrusion Detection (\cite{de2023generalizing}), Healthcare (\cite{ng2021federated}), Driving Style Recognition (\cite{lu2023federated}), Synthetic Data Generation (\cite{GAN1}), Recommender Systems (\cite{gupta2023eqbal,vyas2023federated}) and Autonomous Driving (\cite{SSL3}). The solution approaches used in these directions can be categorized into two types. Firstly, the methods that assume the availability of a limited amount of labelled data at clients, i.e., Federated Semi-Supervised Learning (FedSSL). These methods involve using (a) pre-trained models to annotate clients' local data (\cite{SSL1,jin2023federated}), (b) providing pseudo-labels (\cite{SSL2,SSL5}), and (c) fine-tuning with labelled data (\cite{SSL4,SSL3,he2021ssfl}). Secondly, the more challenging and intriguing problem is to tackle situations where no labelled data is available, known as Federated Unsupervised Learning (FedUL). As in centralized unsupervised learning (\cite{golalipour2021clustering}), clustering is one of the powerful tools in FedUL. It involves dividing the data available on clients into $k$ partitions (called clusters) and finds applications in numerous domains (\cite{gupta2023groupa,gupta2024online}), one of which is as follows -  Consider a situation where multiple banks want to cluster their users' transaction data to differentiate legitimate transactions from fraudulent ones. In such a situation, certain banks may even have limited samples of fraudulent transactions, and due to security and privacy concerns, banks are prohibited from sharing their data with each other. Therefore, having a federated data clustering model can enable banks to reap the benefits of collective learning (\cite{FedFraud,reddy2024deep,zhang2023consumer}). The existing solutions in supervised FL cannot be directly mapped to FedUL clustering as it primarily involves the following challenges:
\begin{enumerate}[label=(\alph*)]

    \item  Each client may not contain data from all $k$ partitions. 
    
    \item As in supervised FL, there does not exist a chronological ordering\footnote{Unlike supervised learning, where fixed ordering or numbering of weights exists across clients, no such numbering exists for centers in unsupervised clustering.} (or synchronisation) on centers, so mapping centers for an averaging function is non-trivial and can lead to bad initialization (\cite{GreedyCentroid}).

\end{enumerate}


To overcome the above challenges, researchers have undertaken various studies in existing literature.  An initial attempt is in (\cite{kFED} (\kFed)), which extends the centralized method from (\cite{awasthi2012improved}) to the federated setting in two steps. First, apply the centralized method locally to clients, and second, share the local information to the server before reapplying therein. While \kFed\ demonstrates single-round communication efficiency, it faces difficulties in certain scenarios as method is highly dependent on nature of data distribution across clients (as evident from our experimental study, Section \ref{sec:Exp} as well). (\cite{MFC}) (\MFC) improvises the method by incorporating multiple rounds of communication. Some other directions explore synthetic data generation or utilize encryption techniques, albeit at the cost of increased computational expenses. 
A common limitation across all these methods is that the best local centers may be distant from the global centers, leading to high-cost deviations across clients. The main reason for this high cost is that the global centers cannot generalize well, potentially due to non-identically and independently (non-iid) distributed data across clients, sometimes resulting in skewed distributions. This can lead to an inductive effect and the potential reluctance of clients to contribute to the federated system. Thus, there is a pressing need for an algorithm that not only solves federated data clustering to find centers that generalize well but also ensures that the centers provided to clients are not too far from local ones, ensuring lower cost deviation across clients and long-term commitment to the system. 

For example, in a banking scenario, banks may need to adjust a global clustering model to suit local factors such as user intelligence, fraudulent behaviour, income levels, and fraud amounts. Therefore, there is need to develop personalized clustering (close to local) models but, at the same time, leverage the benefits of collective learning and achieve uniformly good performance across almost all participating clients. We are the first to address this open direction in federated data clustering and propose \ouralgo\ 
 (\textbf{p}ersonalized-\textbf{F}ederated \textbf{Clus}tering). The algorithm handles the non-triviality, as mentioned before, of missing chronological ordering (or synchronisation) on cluster centers by leveraging the concept of data point-wise gradient updates and nearest local (or global) center mappings. Broadly, the algorithm primarily involves three steps: firstly, finding the initial local cluster models, which are then used to build a collaborative global model. The last step involves specialized unsupervised fine-tuning of clients' local data to achieve separate (or individual) personalized models. To sum up, our contributions are as follows-\\
\textbf{Contributions}:
\begin{itemize}

    \item The proposed method \ouralgo\ results in cluster centers that exhibit lower or comparable cost deviation across clients, leading to a fairer and more personalized solution. The method is the first attempt to provide personalization in federated data clustering. 
    
    \item Further, \ouralgo\ also achieves a lower clustering objective cost in a single communication round between the server and clients, independent of the nature of data distribution (or division) among clients. Thus, \ouralgo\  captures the benefits of both state-of-the-art (\sota) \kFed\ (single shot) and \MFC\ (data distribution independence), resulting in an efficient algorithm. 
    
    \item The methods works for any finite value of $\ell$-norm and especially for $\ell$ value of two ($k$-means), one ($k$-medoids) unlike \sota\ that are applicable only to $k$-means.
    
    \item \ouralgo\ has undergone rigorous analysis using intrinsic and extrinsic federated clustering datasets to showcase its efficacy against \sota\ methods.
   
\end{itemize}

\textbf{Organization:} The rest of the paper is organized as follows: Section \ref{sec:Related} revisits the current literature on FedUL with a primary focus on federated data clustering. Section \ref{sec:prelim} provides an account of the different notations and definitions that will help readers better understand the paper. These notations and definitions will be used throughout the paper to familiarize readers with the proposed algorithm, which we call \ouralgo\ in Section \ref{sec:Algo}. Next, Section \ref{sec:Exp} discusses the experimental setup and the datasets used for validating the efficacy of \ouralgo\ against state-of-the-art (\sota) algorithms on different metrics. Finally, Section \ref{sec:concl} concludes the work with possible directions to work upon. 






\section{Related Work}
\label{sec:Related} 
ML encompasses a wide range of learning paradigms such as supervised, semi-supervised, and unsupervised learning (\cite{MLbookShalev2014understanding}). Among these approaches, supervised and semi-supervised methods require access to labelled data. On the contrary, unsupervised learning helps uncover patterns in unlabeled data (\cite{wang2022federated,giampieri2022unsupervised,gupta2023group,yang2024improved,rodriguez2024novel,rengasamy2022k,izakian2015fuzzy}). Traditional ML approaches typically demand centralized data access, potentially raising privacy issues when handling sensitive information (\cite{PrivacyAndFair,djenouri2024enhancing}). Federated Learning (FL), an emerging paradigm, tackles such challenges by facilitating collaborative model training across distributed  (or decentralized) devices without the need to share raw data (\cite{DecenFedAvg,FedAvg}). Abundant literature is available on supervised  FL (\cite{fedssc,fedmix,federatedTraffic,secureFL,decentralizedFL,adjei2024cov,federatedTL,lobner2022enhancing,banerjee2020multi,hsu2020federated}). On the other hand, work in FedUL is still in its nascent stages, but is gaining momentum, especially considering the increasing availability of unlabeled data (\cite{qi2020small}). FedUL expands to include federated clustering wherein data points are distributed across clients, and all clients collaborate to identify (say $k$) distinct structures known as clusters, each represented by a data point referred to as cluster's center. These clusters must be discovered using each client's locally available data points. A global model consisting of $k$ distinct centers is then learned using local information from clients, and shared back with clients to partition their local data points. The method ensures data privacy by avoiding data sharing and only exchanging partial information other than data points (\cite{kFED}).

\cite{kFED} makes an initial effort to partition the data points and proposes an algorithm which they call \kFed. The algorithm builds upon the (\cite{awasthi2012improved} aka (Awasthi)), assuming the centers are well separated and clusters follow gaussian distribution properties.  \kFed\ executes Awasthi locally on each device to find $k$ local centers, which are then communicated to the server for computing the final clustering. The server then employs a farthest heuristic similar to the offline $k$-center approach\footnote{https://cseweb.ucsd.edu/~dasgupta/291-unsup/}.  Despite its single-round communication efficiency, our experiments show that the cost of clustering with  \kFed\ can be considerably high for some set of clients, potentially leading to a lack of motivation for continued participation. (\cite{GreedyCentroid}) proposes a slightly enhanced greedy centroid-based initialization for \kFed\,  which surpasses centralized $k$-means in specific scenarios.

\cite{MFC} (\MFC) has recently enhanced \kFed\ by reducing the one-shot communication load on clients and leveraging the advantage of multiple communication rounds with minimal information exchange. The approach also addresses the challenge of data distribution dependence as seen in \kFed\ and, as a by-product result, promotes a more equitable distribution of clustering costs across clients. Therefore, both \kFed\ and \MFC\ are among the closest baselines to validate the efficacy of our proposed approach.

Some works in this direction approach the problem by framing it as a generative data synthesis challenge and leveraging concepts from Generative Adversarial Networks (GANs) (\cite{GanSimi,GAN,GANchung2022federated,wijesinghe2023ufed,GAN1}). The broader picture involves training multiple GANs locally at clients and utilizing their parameters to construct a global GAN model. This global GAN model is employed to generate synthetic data and further identify $k$ distinct cluster centers. These centers are subsequently communicated back to clients to partition their local data points. These approaches differ from ours, as we work directly with original data points and aim to identify the best possible centers. \cite{SedFC} also pursues a parallel approach to develop privacy-preserving distributed clustering by incorporating concepts from cryptography. The proposed method initially computes local center updates and then shares encrypted information using lagrange encoding back to the server. Subsequently, the server aggregates all secret distance codes from the clients and performs subsequent communication updates. While the algorithm harnesses the advantages of encryption-decryption to safeguard data privacy, such techniques entail substantial computation overhead and communication costs, thereby hindering the scalability of the approach. Similarly, \cite{BCFL} employs federated data clustering within the blockchain's committee-based consensus protocol. However, the additional overheads counterbalance the performance improvement. 

It is important to note that no existing work in federated data clustering has specifically focused on addressing the challenge of cost distribution spread across clients and fostering a more equitable clustering as a primary goal by leveraging the principles of personalization from a supervised setting (\cite{li2021ditto}). Furthermore, extending these principles to an unsupervised setting is non-trivial due to the lack of chronological ordering\footnote{Unlike supervised learning, where fixed ordering or numbering of weights exists across clients, no such numbering exists for centers in unsupervised clustering for direct averaging.} on centers and the varying data division across clients. We will address this direction in the present paper. 

\textbf{Other Works}: Recent studies have investigated federated data clustering in the soft (or better called fuzzy) clustering paradigm, wherein a data point can have membership to more than a single cluster (\cite{FedFCM,ArchiFedFCM,hu2023efficientFCM,FrameFedClusAnaly}). These algorithms are highly dependent on rounding methods used for deployment in real-world applications. Therefore, these works differ slightly from ours as we focus on hard assignments. Furthermore, (\cite{DPCBased, FedClus,rehman2022divide}) extends a variant of DBSCAN for the federated setting, where clustering relies on characteristics such as data point density in space. Typically, these methods struggle with high-dimensional data and lack control over a number of clusters. In contrast, the primary focus of the current work is to extend $k$-centroid clustering to the federated setting. A specific enhancement involves clustering image datasets in a federated environment by leveraging the additional advantages of incorporating the latent representation of these images using encoders. These approaches increase a major portion of the computational load onto the client devices, which are now tasked with training encoders (utilizing backbone networks like ResNet18 (\cite{Resnet})). Furthermore, heavy communication bandwidth is required between clients and servers as it involves sharing information about centres and encoder parameters (\cite{CCFC,CCFCPlusPlus}). Similarly,  (\cite{FedGrEM,GMMBased}) investigates federated data clustering in Gaussian Mixture Models (GMM) and Expectation Maximation (EM) algorithms, respectively. Similarly, \cite{kPFED} proposes a decentralized and cloud framework based on the federated averaging method. Furthermore,  (\cite{GMMBased,kPFED}) also highlights the need for personalized models for all clients in probabilistic and peer-to-peer networks respectively. It's essential to note that our focus lies in clustering data in a federated setting, which is altogether different from federated client clustering. The latter entails smartly selecting a subset of clients for model updates (\cite{islam2024fedclust,partialParticipation3,partialParticipation1,partialParticipation4,partialParticipation2,partialParticipation5}).

\section{Preliminaries}
\label{sec:prelim} 
Let $X \subseteq \mathbb{R}^{h}$ be a set of data points that are distributed among $Z$ clients in a federated setting. Let the data points on any client $z \in [Z]$ be $X^{(z)}$. 
Each data point in $X^{(z)}$ is a $h$-dimensional real-valued feature vector. We assume that these data points are embedded in a metric space having distance metric $d: X \times X \to \mathbb{R}^{+}$  that measures dissimilarity between data points using any $\ell$-norm represented as $||\cdot||_\ell$. Note that the data points $X$ belong to $[k]$ different true distributions, and the goal of any clustering algorithm is to partition the data points spread across clients into a set of disjoint sets (called clusters) represented by the set of global centers denoted by set $C^{g} = \{c_1^{g},  c_2^{g}, \ldots,  c_k^{g}\}$. The computation of finding these global centers involves initially computing the best local centers that partition the local data $X^{(z)}$ (for any $z \in [Z]$) into $k$ disjoint sets represented by $C^{(z)} = \{ c_1^{(z)},  c_2^{(z)}, \ldots,  c_k^{(z)}\}$. We denote the local assignment function at each client by $\phi^{(z)}: X^{(z)} \to C^{(z)}$. Note that the data points on any client $z$ may not be sampled from all $[k]$ true distributions, and this idea is captured using the notion of heterogeneity in federated settings. Formally, it is defined as follows:
\begin{definition}[Heterogeneity] Given $k$, the heterogeneity level ($\mathtt{H}$) determines the maximum number of distributions the data points $X^{(z)}$ on a client $z \in [Z]$ belongs to, i.e., $\mathtt{H} \le k$.
\end{definition}
In practice, determining the exact level of heterogeneity ($\mathtt{H}$) on a client is often not feasible. Consequently, a common approach in federated data clustering literature is to compute $k\ (\ge \mathtt{H})$  partitions on each client (\cite{kFED, MFC}). These local partitions are not arbitrary selections but are the one that minimizes the following local objective cost:

\begin{definition}[Objective Cost] Given $k$, $\cup_{z \in [Z]} X^{(z)}$, and distance metric $d: X \times X \to \mathbb{R}^+$ with norm value $\ell$ the global objective cost $L_\ell$ over all clients $Z$ and local objective cost $L_{\ell}^{(z)}$ of client  $z$  of $(k,\ell )$-clustering in a federated setting with a set of global centers $C^{g}$ is computed as follows:
\label{def:objCOst}
\begin{align*}
L_\ell^{(z)}(C^g) =  \sum_{x \in X^{(z)} } \sum_{c^{g} \in C^{g}}   \mathbb{I}\left( \phi^{(z)}(x) = c^g\right)\ \big( d(x, c^g) \big)^\ell
\end{align*}
\begin{equation}
L_\ell(C^g) = \sum_{z\in [Z]} L_\ell^{(z)}(C^g)
\label{def:LocalObj}
\end{equation}
\end{definition}

In a federated setup, comparing methods based on the \textbf{mean objective cost per data point} is often more realistic than the total objective cost across clients. The primary reason is that dataset sizes across clients can differ significantly in federated settings. Thus, evaluating the per-point cost incurred by clients when using global centers makes more sense. Mathematically, this can be formulated as follows:

 \begin{equation}
        \Mu^{(z)}(C^g) =  \frac{L^{(z)}_{\ell}(C^g) }{|X^{(z)}|}  
    \end{equation}
    
     \begin{equation}
        \Mu(C^g)  =  \frac{1}{Z} \sum_{z \in [Z]} \Mu^{(z)}(C^g)
        \label{eqn:perPointCost}
    \end{equation}

where $\Mu^{(z)}(C^g)$ is the mean cost per data point on any client $z$ and $\Mu$ is the mean objective cost per data point over all clients using center set as $C^g$. Also, further note that in a federated setting, the objective cost suffered by any client $z$ can significantly differ from that of other clients, owing to the fact that data points from $\mathtt{H}\le k$ distributions can be distributed (or generated) in a highly skewed manner among (or at) clients. Therefore, if the global centers deviate too much from the best local centers, clients might feel reluctant to contribute to the federated environment to learn a better global center representation. Thus, the aim is to not solely focus on minimizing objective cost (or per point cost) but rather to find a $(k, \ell)$-clustering in the federated setting that is fair for all clients, i.e., one which achieves near uniform cost across all clients. We formally define such a clustering as follows:




\begin{definition}[Fair Federated Clustering] Given that data points are sampled from $k$  true clusters and are distributed over $Z$ clients. Then, for any two set of federated global centers $C^{g}_1$ (produced by Algorithm 1) and $C^{g}_2$ (produced by Algorithm 2), we say that $C^{g}_1$ is more fair than $C^{g}_2$ if the \textbf{cost deviation per point} ($\sigma$) is lower for  $C^{g}_1$ than $C^{g}_2$. Here $\sigma$ over centers $C^{g}_i$ for $i \in \{1, 2\}$ is given as follows:
 \begin{equation}
        \Var(C^{g}_i) =  \sqrt{\frac{ \sum_{z \in [Z]} \left(\Mu^{(z)}(C^{g}_i) -\Mu(C^{g}_i)\right)^2  }{Z} }
\label{eqn:stdDevCost}    
\end{equation}

\end{definition}    

On similar We now examine the proposed algorithm, which addresses federated data clustering and identifies centers that are relatively close to the local centers by introducing the concept of personalization (\cite{li2021ditto,ye2024upfl}), thereby promoting fair data clustering among clients.
Towards the end of the next section, we point out the main working mechanism behind the algorithm that helped to solve the non-trivial extension of achieving personalization from supervised literature to unsupervised federated data clustering. 

\section{\ouralgo: personalized Federated Clustering}
\label{sec:Algo} 
We propose a novel algorithm called \ouralgo\ (\textbf{p}ersonalized-\textbf{F}ederated \textbf{Clus}tering). The algorithm comprises three phases, which are explained in subsequent sections below. The complete pseudo-code for \ouralgo\ is described in Algorithm \ref{algo:pFedCLusMain}, and its implementation is available on the public repository\footnote{https://github.com/P-FClus/p-FClus}.

\begin{algorithm}[h!]
\caption{\ouralgo: {\textbf{p}ersonalized-\textbf{F}ederated \textbf{Clus}tering } }
\label{algo:pFedCLusMain}

\texttt{procedure} \textbf{ \ouralgo} $\left( \{X^{(z)}\}^{Z}_{z=1}, k, \ell, \eta, \lambda\right)$

\begin{algorithmic}[1] 

\STATE{$\forall\ z \in [Z]$ in parallel:} \\
  \INDSTATE  $C^{(z)} \gets$ \textbf{ clientInitialization}($X^{(z)}$, $k$, $\ell$)\\ 
  \STATE call procedure \textbf{Server}$\left(\left\{C^{(z)}\right\}_{z=1}^Z\right)$\\
  \STATE /* Each client receives set of global cluster centers $ C^{g}$ */ \\
  \STATE{$\forall\ z \in [Z]$ in parallel:} \\
  \INDSTATE /* Each client personalizes the $C^g$ using SGD with learning rate $\eta$ and fine-tuning level $\lambda$ */ \\
  \INDSTATE$C^{(z)}, \phi^{(z)} \gets $call procedure \textbf{ clientPersonalization}$\left(C^{(z)}, C^{g}, \eta, \lambda, X^{(z)}\right)$\\

  \STATE /* Global cluster centers received at all clients $ C^{g}$ from server has been personalized (fine-tuned) for use.*/ \\

  \RETURN $\left\{\phi^{(z)}, C^{(z)} : C^g \text{ ( personalized ) }\right\}_{z=1}^Z$


\end{algorithmic}
\label{algo:pFedCLusMain}
\end{algorithm}

\begin{algorithm}[h!]
\caption{\ouralgo's Client-side Initialization Procedure}
\label{alg:pFedclus1}

\texttt{procedure} \textbf{ clientInitialization}$\left(X^{(z)}, k, \ell \right)$

\begin{algorithmic}[1] 

  \STATE /* Apply ($k, \ell$)- clustering using (\cite{lloyd1982least}) for $\ell$ = $2$ ($k$-means) and (\cite{kMedianLP}) for $\ell$ = $1$ ($k$-mediod).   */\\
  \STATE  $C^{(z)} \gets$ \textbf{($k, \ell$ )-\text{clustering}}$(X^{(z)})$\\ 
  \RETURN  $C^{(z)}$


\end{algorithmic}
\label{alg:pFedclus1}
\end{algorithm}

\begin{algorithm}[h!]
\caption{\ouralgo's Server-side Procedure }
\label{alg:pFedclus2}

\texttt{procedure} \textbf{ Server}$\left(k, \ell, \left\{C^{(z)}\right\}_{z=1}^{Z}\right)$

\begin{algorithmic}[1] 
\STATE /*Post client initialization*/\\


\STATE $S \gets  \cup_{z \in [Z]} C^{(z)}$\\

 /*Apply ($k, \ell $)-clustering (\cite{lloyd1982least,kMedianLP} for $\ell$ value of $2, 1$ respectively) on $S$  to get $k$ global centers*/\\

\STATE $C^g \gets$ \textbf{($k, \ell $)-\text{clustering}}(S) \\

\STATE  In parallel  $\forall z \in [Z]$:\ \textbf{$\text{return}$} $ C^{g}$  to all clients \\

\end{algorithmic}

\label{alg:pFedclus2}
\end{algorithm}

\subsection{Client Initialization}
Initially, all the clients run an initialization procedure described in Algorithm \ref{alg:pFedclus1} in parallel. Each client $z \in [Z]$ executes a $\ell$-norm clustering algorithm to find a set of $k$ local cluster centers $C^{(z)}$. These centers can be computed using known heuristics or approximate algorithms, such as those described in (\cite{lloyd1982least}) for $\ell$-norm value of two ($k$-means) and in (\cite{kMedianLP}) for the $k$-medoid objective. These methods ensure that the centers obtained minimize the objective cost on the local datasets. After computing their respective local cluster centers, each client shares its set of local cluster centers $C^{(z)}$ with the server.

\subsection{Server execution}
The complete server side procedure is summarized in Algorithm \ref{alg:pFedclus2}. In line number $1$ to $2$ of procedure, after receiving the set of centers $C^{(z)}$ from all devices $z \in [Z]$, the server constructs a set $S$ by aggregating them, i.e., $S = C^{(1)} \cup C^{(2)} \cup \ldots \cup C^{(z)}$.  Subsequently, the server applies $(k, \ell)$-clustering algorithm (\cite{lloyd1982least,kMedianLP}) to this set $S$ to determine set of $k$ global centers $C^{g}$ in line $3$. These global centers are expected to minimize the objective cost across any of the clients. However, they can in some cases far from the current local data due to heterogeneity. As a result, these global centers are distributed back to all clients, allowing them to personalize their centers, thus leading to lower cost deviation across clients. 

\subsection{Client side personalization}
After receiving the set of global centers, all clients use their set of local centers to fine-tune global centers to form personalized centers using procedure available in Algorithm \ref{alg:pFedclus3}. The fine-tuning level $\lambda$ can be kept consistent across clients, or clients can vary it according to their preferences. For each $h$-dimensional data point in $x  \in X^{(z)}$, the client identifies the closest local ($c^{(z)} \in C^{(z)}$, line number $5$) and global center vector ($c^g \in C^{g}$, line number $6$) and fine-tunes the global ones by minimizing the following function for finite $\ell$:
\begin{equation}
P(x) = \underbrace{\frac{1}{2}  \big|\big| c^g - x\big|\big|_\ell}_\text{clustering cost} + \underbrace{\lambda \left( c^g- c^{(z)} \right)^2}_\text{regularization penalty} \text{ 
 \ \ (Personalization Objective) }
\label{eqn:PersonObjective}
\end{equation}

The above personalization objective (Equation \ref{eqn:PersonObjective}) emphasizes updating global centers to minimize the local cost (computed using the set of $C^{g}$ centers)  while ensuring that $C^{g}$ does not collapse to $C^{(z)}$ by addition of a regularization factor. The role of the regularization factor is to ensure that the global centers are not too much deviated by incorporating penalty terms in the form of $L2$-regularization.

\begin{algorithm}[h!]
\caption{\ouralgo's Client Personalization procedure }
\label{alg:pFedclus3}
\texttt{procedure} \textbf{ clientPersonalization}$\left(C^{(z)}, C^{g}, \eta, \lambda, X^{(z)}\right)$

\begin{algorithmic}[1] 
\STATE Initialize assignment function $\phi^{(z)} \gets \Phi$ (empty)
\INDSTATE[1] \textbf{while} { tuning steps $t$ or convergence} \textbf{do}

\INDSTATE[1] $t \gets t - 1$
\INDSTATE[1] \textbf{for}{ $x \in X^{(z)}$} \textbf{do }
\INDSTATE[2] $c^{(z)} \gets \argmin_{c^{(z)}  \in C^{(z)}} \left(d(x, c^{(z)})\right)$
\INDSTATE[2] $c^{g} \gets \argmin_{c^{g}  \in C^{g}} \left(d(x, c^{g})\right)$
\INDSTATE[2] $\eta \gets 1\bigg/\left|\mathbb{I} \left(\argmin_{c^{g}  \in C^{g}} d(x, c^{g}) = c^{g} \right)\right|$
\INDSTATE[2] $c^{g} \gets c^{g}  - \eta \Delta_{c^{(g)}}\left(P(x)\right)$  (Using Equation \ref{eqn:kmeansUpdateRule} or \ref{eqn:kmediodUpdateRule} and $\lambda$) 
 \ \ \ /*$C^{g}$ are personalized centers for client z*/\\
 \INDSTATE[2] $\phi^{(z)}[x] \gets c^{g}$
 \RETURN $C^{g},\phi^{(z)} $
\end{algorithmic}
\label{alg:pFedclus3}
\end{algorithm}



Now, for $k$-means, the norm $(\ell)$ takes the value of two and since minimizing euclidean distance ($2$-norm)  is the same as minimizing squared euclidean, therefore the derivative with respect to global center results in the following:
\begin{equation}
    \Delta_{c^{g}} (P(x))\big]_{\ell=2}=  (c^g - x) + 2\lambda (c^g- c^{(z)})
    \label{eqn:kmeansUpdateRule}
\end{equation}

In the case of $k$-medoids, the norm takes the value of one i.e., minimizing $L1$-distance and  we get the derivative value as follows:
\begin{equation}
    \Delta_{c^{g}} (P(x))\big]_{\ell=1}= 1/2 + 2\lambda (c^g- c^{(z)})
    \label{eqn:kmediodUpdateRule}
\end{equation}

Now, we can update the global center using Stochastic Gradient Descent (SGD) based personalization and find the updated global center which are not too far from local ones in line number $8$ of the procedure.
\begin{equation}
    c^{g} \gets c^{g} - \eta \Delta_{c^{g}} (P(x))
\end{equation}

\noindent where we set $\eta$ = $\left|\frac{1}{\mathbb{I} \left(\argmin_{c^{g}  \in C^{g}} d(x, c^{g}) = c^{g}\right)}\right|$ i.e, multiplicative inverse of the number of data points currently assigned to $c^g$ act as learning rate (line number $7$).

Note that in the special case of the $1$-norm, i.e., $k$-medoids, there is a constraint that the center should be a data point. However, during the personalization process, it can happen that the final obtained center is not a data point. In such cases for $k$-mediod objective, the nearest data point to the final center within the client's local data is chosen as the center.

\textbf{Novelty}: Therefore, to summarize, it is important to note that the non-trivial nature of extending the literature in supervised personalization (\cite{li2021ditto,ye2024upfl}) to unsupervised learning is handled by the intrinsic design of the \ouralgo\ procedure. Rather than demanding the need for having a synchronization (or chronological) ordering on centers across clients for direct averaging, unlike prior works, we use the data point-wise gradient update and use nearest local or global center mappings. We now validate the efficacy of the \ouralgo\ against state-of-the-art (\sota) methods.

\section{Experimental Result and Analysis}
\label{sec:Exp} 
We will now validate the performance of the proposed \ouralgo\footnote{https://github.com/P-FClus/p-FClus}\ against \sota\ approaches on different synthetic and \textit{benchmarking} real-world datasets used in clustering literature (\cite{chierichetti2018fair,bera2019fair,modha2003feature,ziko2021variational}).
These are as follows:
\begin{itemize}
    \item \textbf{Synthetic Datasets} (\texttt{Syn}) - The synthetic datasets generated can mainly be categorized into the following:
    \begin{itemize}
        \item \textbf{No Overlap} (\texttt{Syn-NO}): It contains data points from ten bi-variate gaussian distributions $\{\mathcal{N}_i(10i, 1)\}_{i=1}^{10}$, ensuring that the dataset has well-separated clusters.
        
        \item \textbf{Little Overlap} (\texttt{Syn-LO}): It consists of data points from ten bi-variate gaussian distributions arranged in a way that the consecutive pairs of distributions overlap only after two standard deviations. The standard deviation is set to two for all gaussian's. 
        
        \item \textbf{Overlapping} (\texttt{Syn-O}): It consists of data points from ten bi-variate gaussian distributions arranged such that in each consecutive pair of distributions, the mean of one distribution and three standard deviations of the other distribution in the pair touch each other.  The standard deviation is set to three for all gaussian's. The data generation code is available in the code repository$^6$.

    \end{itemize}

    \item \textbf{Real-world Datasets} - The real-world datasets used in the study can be further divided into two types. The first type comprises datasets that require pre-processing to make them ready for use in a federated environment i.e,  they are extrinsic in nature. The second type includes datasets that are inherently captured in a federated manner, where the data is naturally divided among clients.
    
\begin{itemize}
\item Non-Federated Datasets (Extrinsic) 
    \begin{itemize}

        \item \textbf{\texttt{Adult}}: The census record collection of $1994$ US citizens. It comprises $32562$ records with feature attributes under present study as age, fnlwgt, education\_num, capital\_gain, and hours\_per\_week. These attributes are consistent with prior works on clustering (\cite{ziko2021variational}). The dataset is openly available\footnote{https://archive.ics.uci.edu/ml/datasets/Adult}.  
        
        \item \textbf{\texttt{Bank}}:  A direct phone call marketing campaign data of banks in the Portugal region. It comprises $41108$ records containing information about consumers' age, duration, campaign, cons.price.idx, euribor3m, nr.employed as attributes. The features selected for experimentation align with previous literature, and the dataset is publicly available\footnote{https://archive.ics.uci.edu/ml/datasets/Bank+Marketing}.
        
        \item \textbf{\texttt{Diabetes}}: US clinical records collected over ten years. The features chosen are age and time\_in\_hospital, and is publicly available\footnote{https://archive.ics.uci.edu/ml/datasets/Diabetes+130-US+hospitals+for+years+1999-2008}. 

        \item \textbf{\texttt{FMNIST}}: Contains $60,000$ training images covering ten classes of fashion items, each at a resolution of $28 \times 28$ pixels and is publicly accessible\footnote{https://github.com/zalandoresearch/fashion-mnist/}.

    \end{itemize}

\item Federated Datasets (Intrinsic)
     \begin{itemize}
        \item \textbf{\texttt{FEMNIST}}: It is a handwritten character recognition dataset where each client corresponds to a writer from the EMNIST dataset\footnote{https://github.com/TalwalkarLab/leaf/tree/master/data/femnist}. The dataset's intrinsic heterogeneity is $\mathtt{H} = 62$, i.e., $62$ true distributions are distributed among $500$ clients during data collection.

        \item \textbf{\texttt{WISDM}}: It is a publicly available wireless sensor data mining dataset consisting of $1,098,207$ samples for activity recognition using mobile phone accelerometers\footnote{https://www.cis.fordham.edu/wisdm/dataset.php} for recognizing six ($\mathtt{H}=6$) activities: walking, jogging, going upstairs, going downstairs, sitting, and standing across $36$ clients.

     \end{itemize}

\end{itemize}

\end{itemize}

We compare \ouralgo\ on both $k$-means ($\ell$-norm of $2$) and $k$-mediod ($\ell$-norm of $1$) objectives (see Definition \ref{def:objCOst}) against the following baselines:

\begin{itemize}
    \item \textbf{Centralized Clustering} (\Clus) (\cite{SurveyClusteringNew}): An euclidean-based centralized version of $k$-centroid clustering. The aim is to minimize the objective cost using a norm value of two ($\ell=2$), resulting in the well-known Lloyd's heuristic (or simply $k$-means) (\cite{lloyd1982least}). When the norm value is one ($\ell=1$), a centralized variation of $k$-mediod clustering is achieved, where the set of centers is restricted to the data points in the dataset (\cite{kMedianLP}).

    \item \textbf{Oneshot Federated Clustering} (\kFed) (\cite{kFED}): The method is a federated data clustering approach that leverages the data heterogeneity among clients. The method executes Awasthi's $k$-means (\cite{awasthi2012improved}) locally on clients, and then clients share information about local centers with the server. The server then applies a variant of the farthest heuristic to select the best $k$ global centers. These global centers are then shared back with clients for local clustering. Note that the method works well only when the network has high heterogeneity. 
    
    \item \textbf{Multishot Federated Clustering} (\MFC) (\cite{MFC}): The method enhances the \kFed\ approach incorporating multiple communication rounds between the server and clients. These additional rounds help adapt the method to settings with lower and higher heterogeneity levels. The authors demonstrate that as a byproduct, \MFC\ achieves a certain level of fairness. Note that in additional rounds, the only shared information pertains to the maximum cost cluster center, not all local centers.  

\end{itemize}

An important point to note is that the \kFed\ and \MFC\ methods are intrinsically designed to work only for the $k$-means objective.  Thus, this limits comparing these methods to $k$-means version of our \ouralgo. To validate the performance of the $k$-median objective of \ouralgo, we compare it to the centralized setting. The metrics involved in comparing the efficacy include the following:
\begin{itemize}
    \item \textbf{Mean Cost per data point} ($\Mu$ $\downarrow$): It is the mean (or average) objective cost  experienced by each of the data points and is lower the better. It is computed as described in Equation \ref{eqn:perPointCost}.


    \item \textbf{Cost Deviation per data point} ($\Var$ $\downarrow$): It is a fairness metric that measures the standard deviation in per-point cost experienced by clients. The empirical value is estimated using Equation \ref{eqn:stdDevCost}.

    \item \textbf{Maximum Cost per data point} ($\Max$ $\downarrow$): It helps in estimating the worst per point cost that any client has to suffer and is computed as $\Max =  \max_{z \in [Z]} \Mu^{(z)}(C)$. A lower value indicates a more fairer clustering for clients. Note that we consider $C$ in all these metrics as personalized global centers for \ouralgo, and for \kFed, \MFC, we consider $C$ as the set of global centers computed by the method.

\end{itemize}

\subsection{Experimental Setup}
All experiments are conducted on an Intel Xeon $6246R$ processor with $280$GB of RAM, running Ubuntu $18$ and Python $3.8$. We report the results as the mean and standard deviation of five independent runs, with the seed chosen from the set $\{0, 300, 600, 900, 1200\}$. The complete reproducible code is available online$^{6}$. Next, we investigate the distribution of data among clients, focusing mainly on two different settings described below:
   \begin{figure*}[t!]
    \centering 
    \begin{tabular}{@{}c@{}c@{}c@{}c@{}}
    \includegraphics[width=0.24\textwidth]{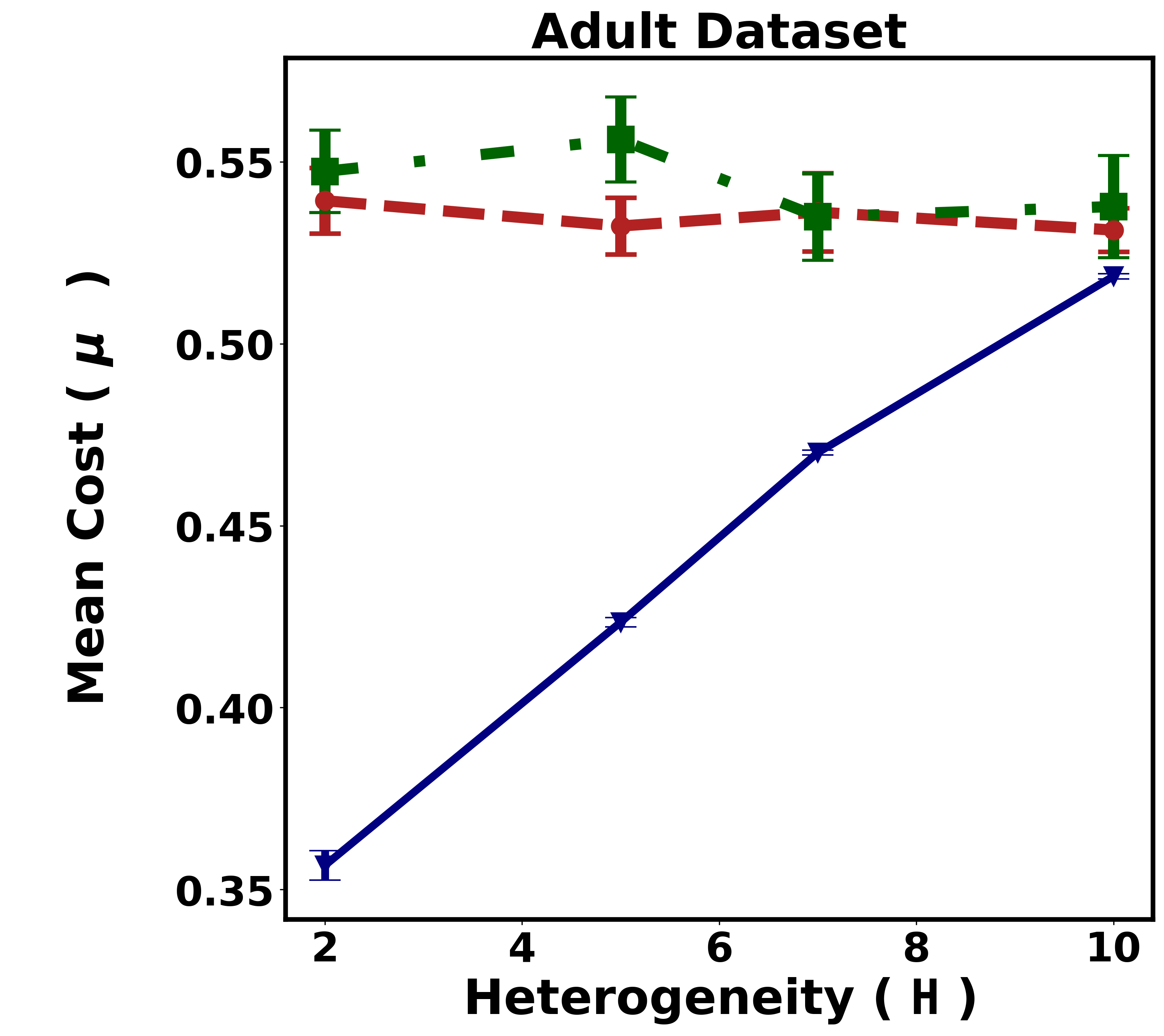}&\includegraphics[width=0.24\textwidth]{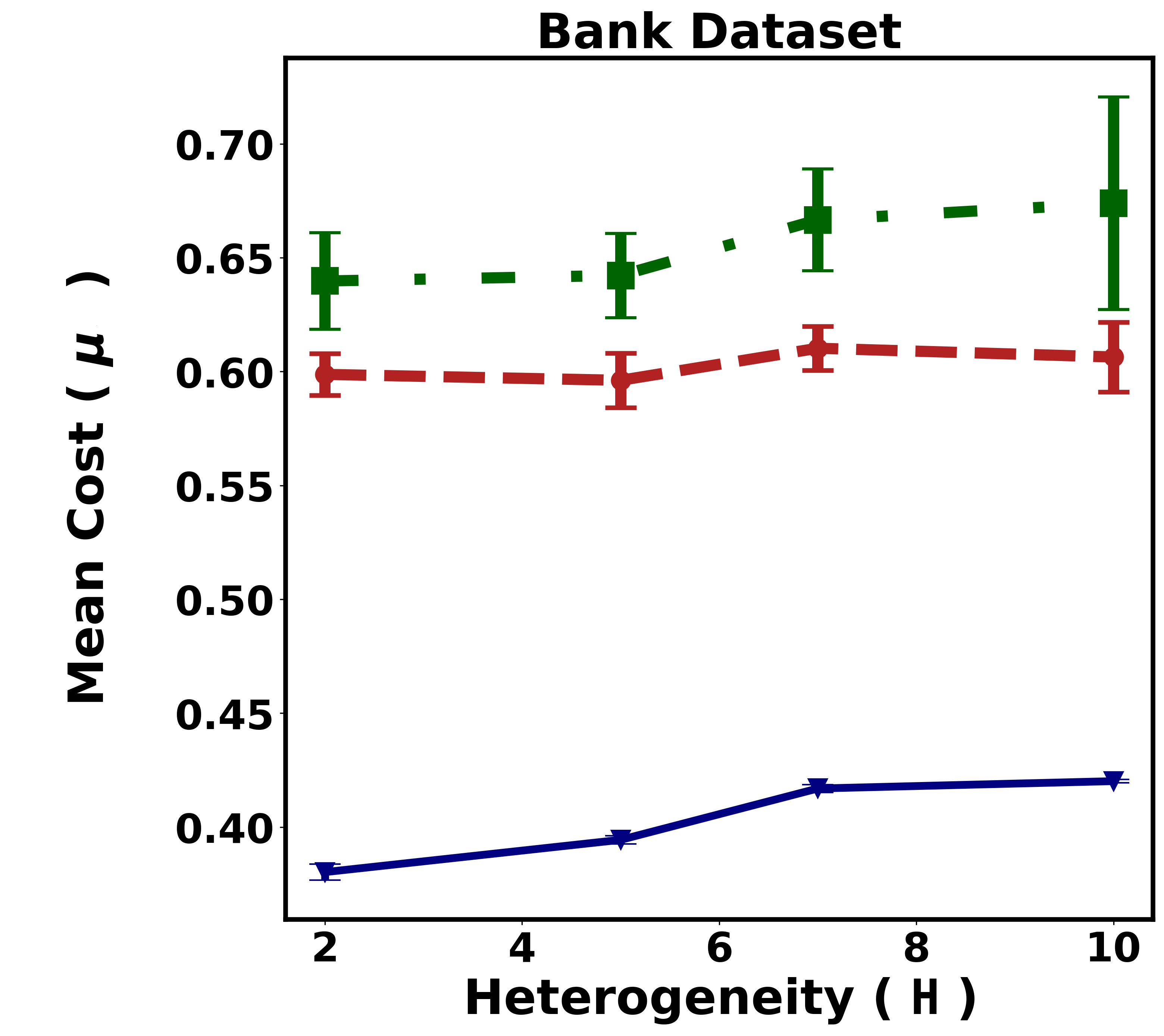}&\includegraphics[width=0.24\textwidth]{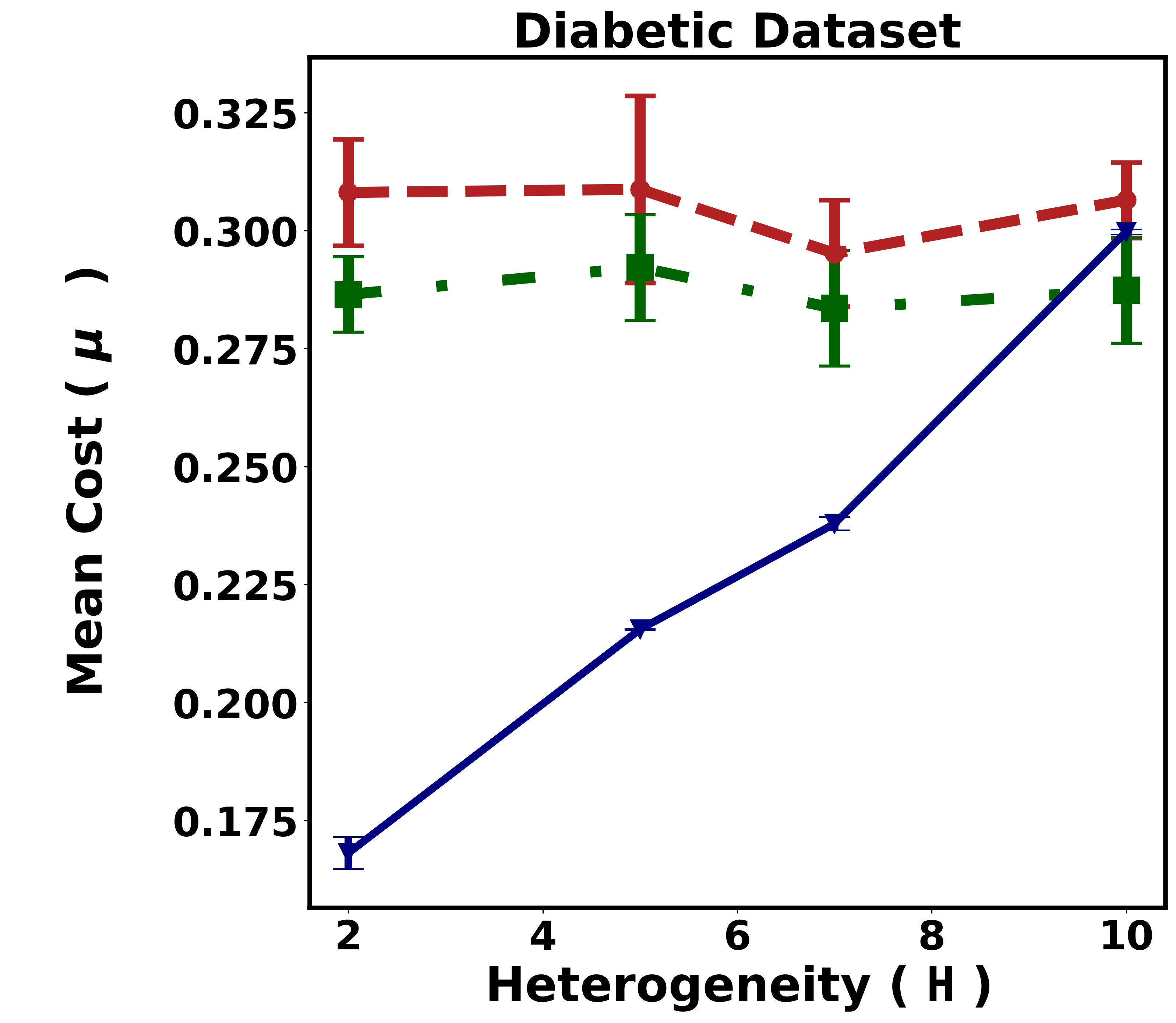}&\includegraphics[width=0.24\textwidth]{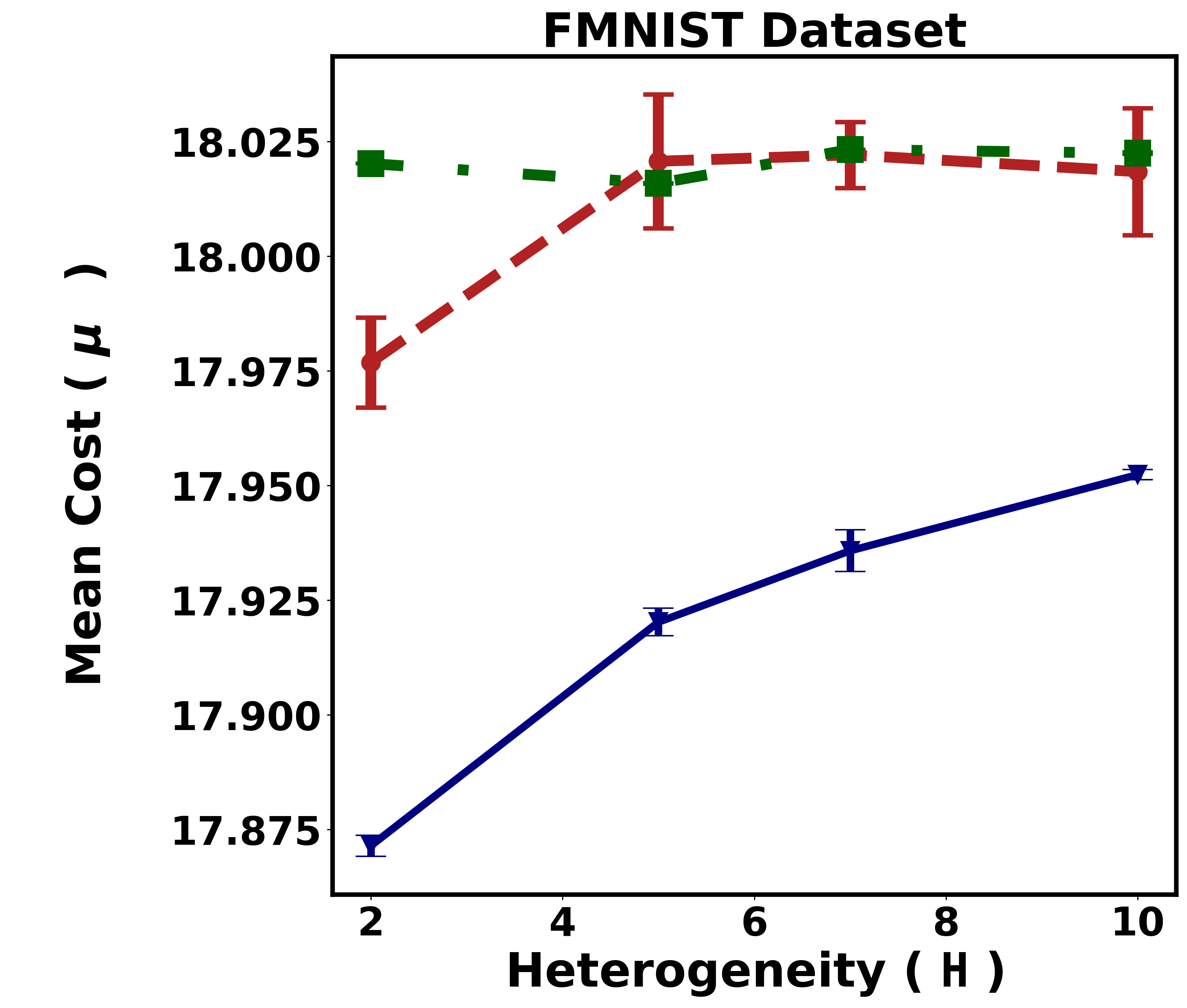}\\
    \includegraphics[width=0.24\textwidth]{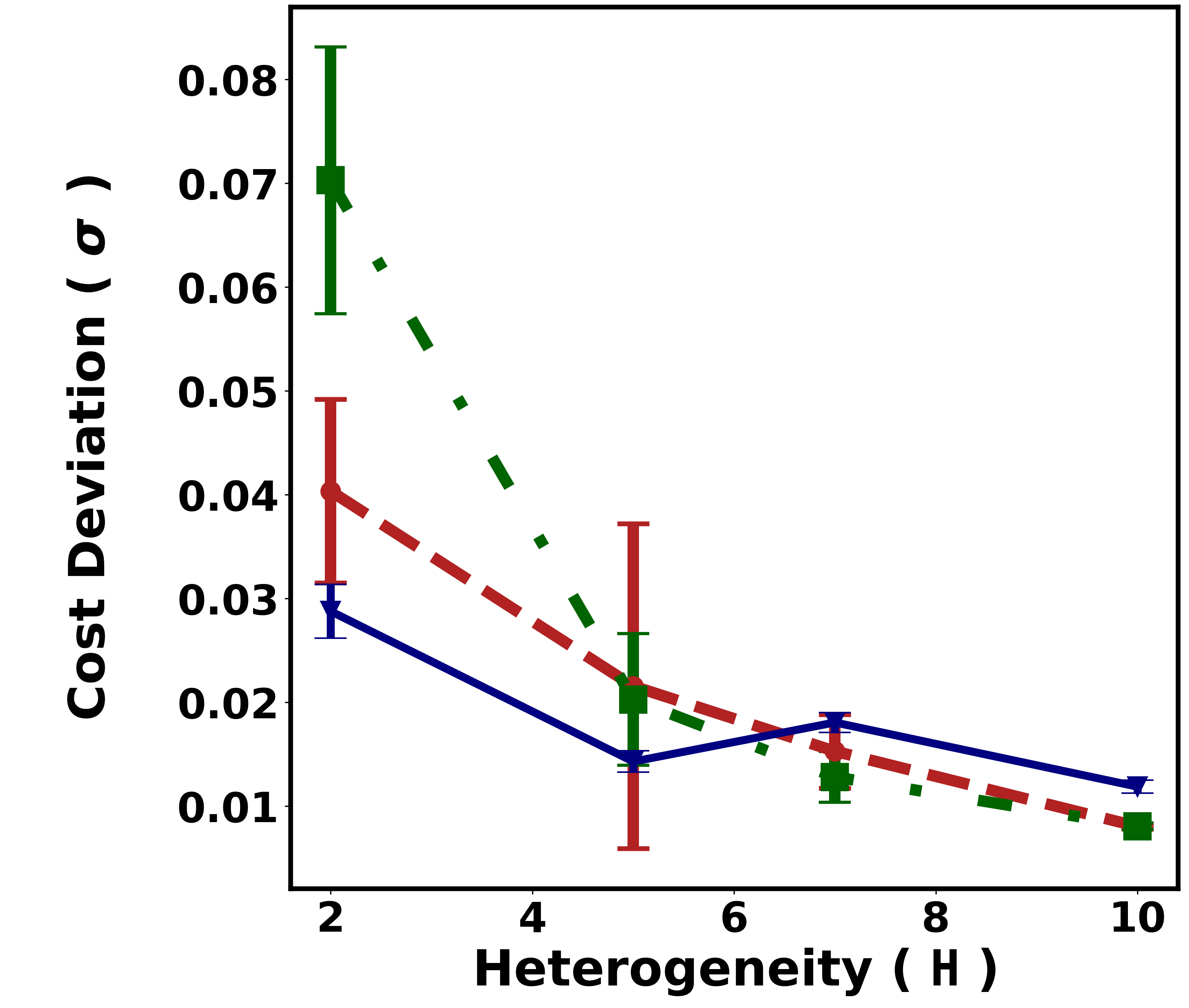}&\includegraphics[width=0.24\textwidth]{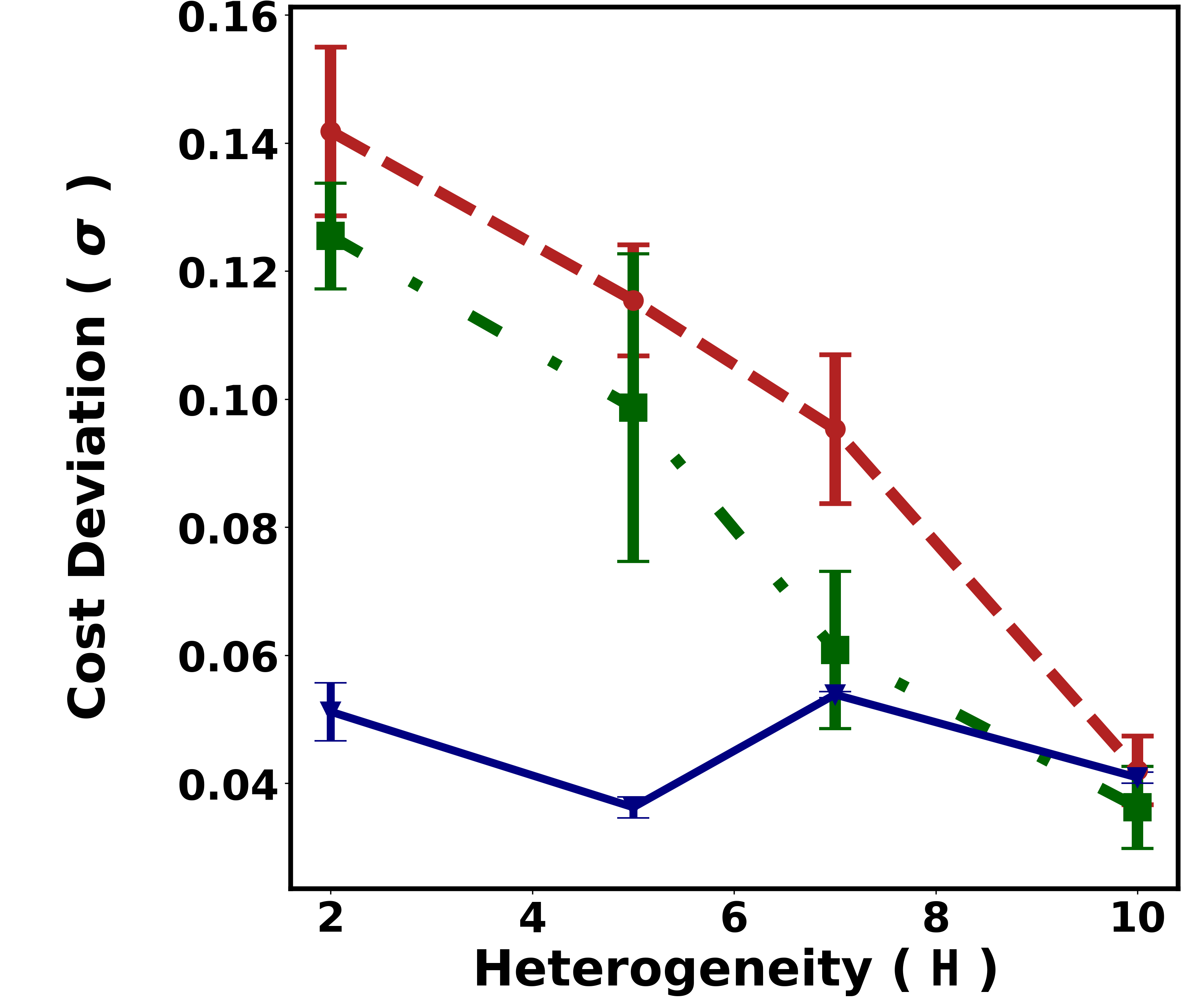}&\includegraphics[width=0.24\textwidth]{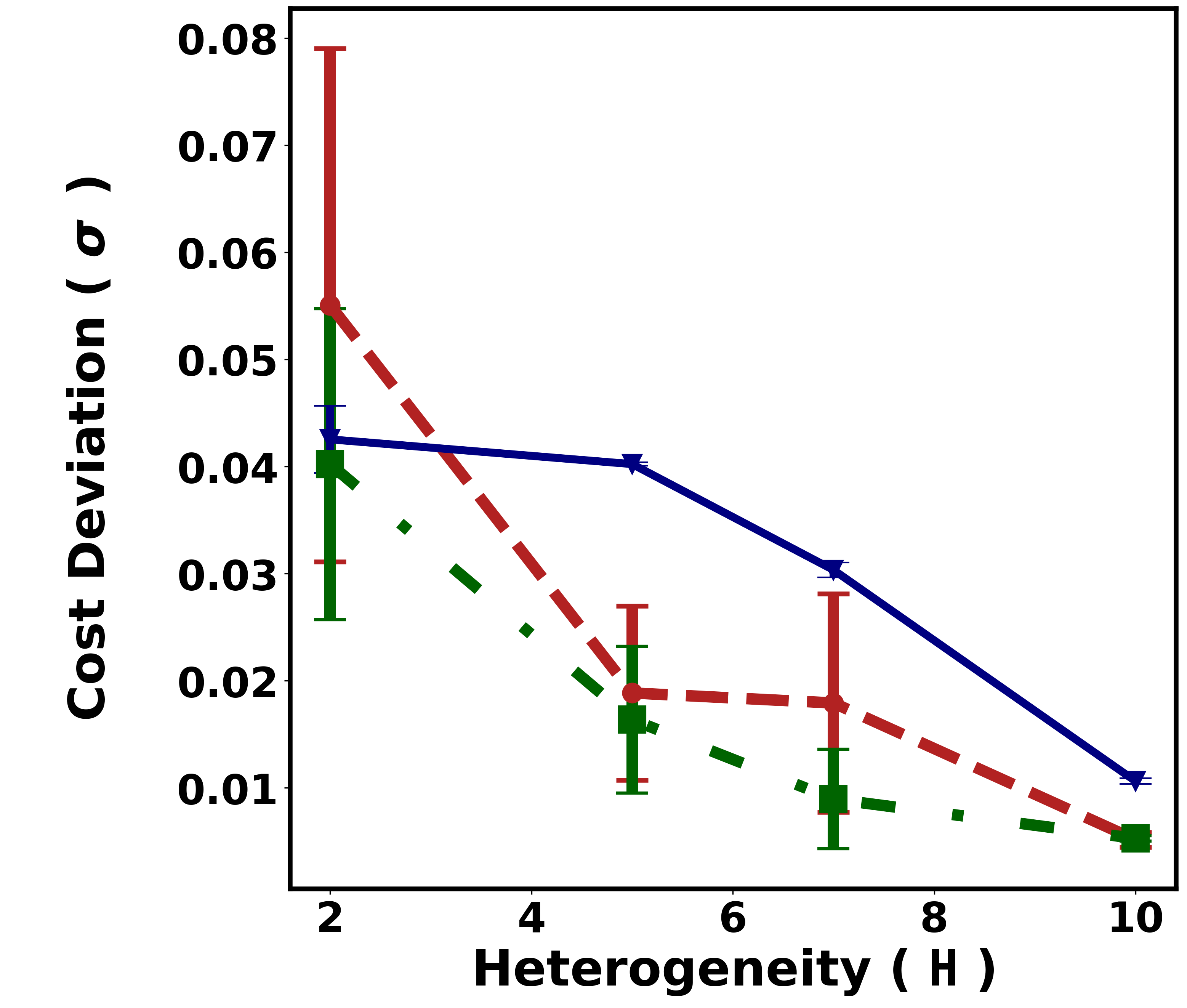}&\includegraphics[width=0.24\textwidth]{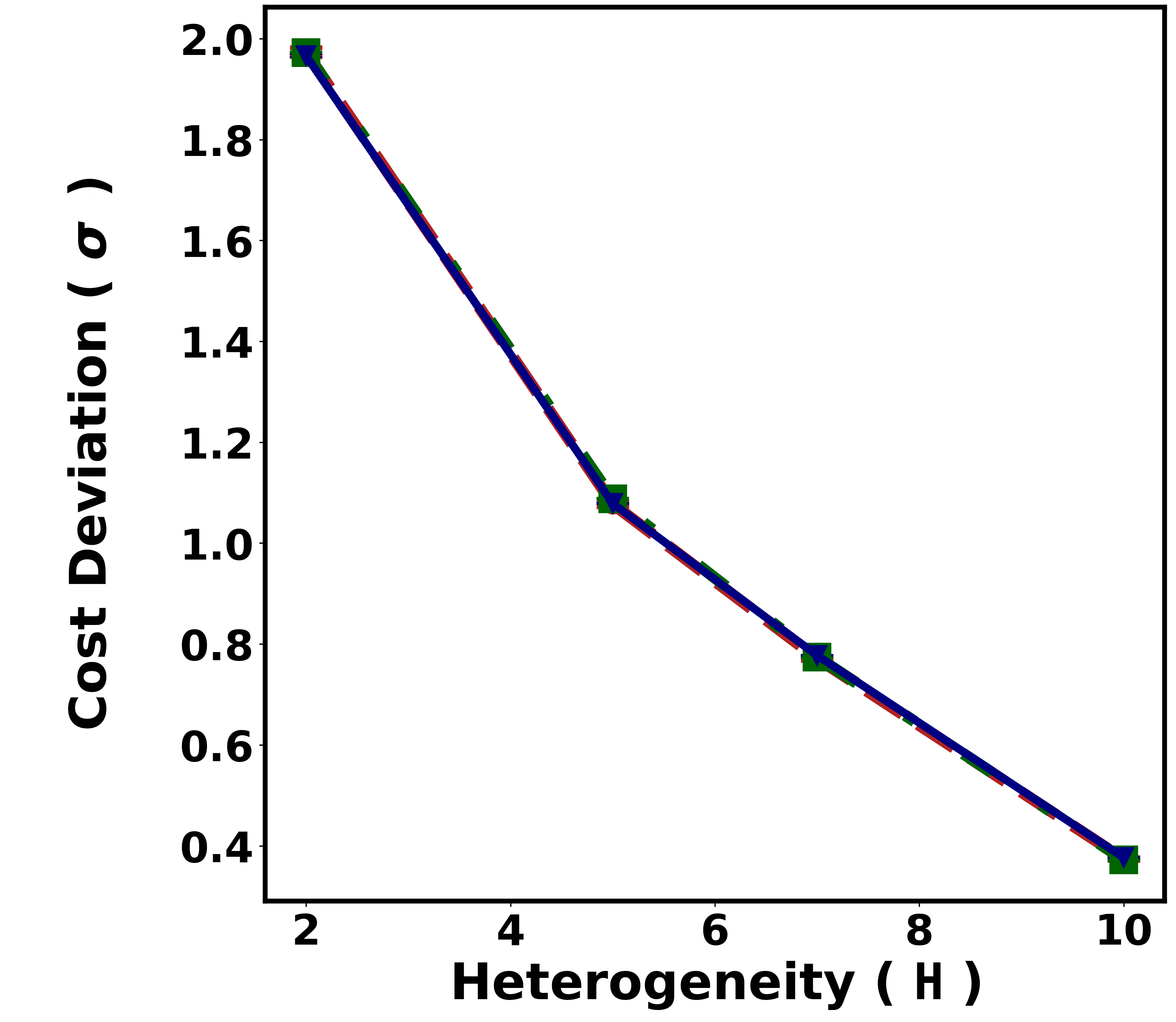}\\
    \includegraphics[width=0.24\textwidth]{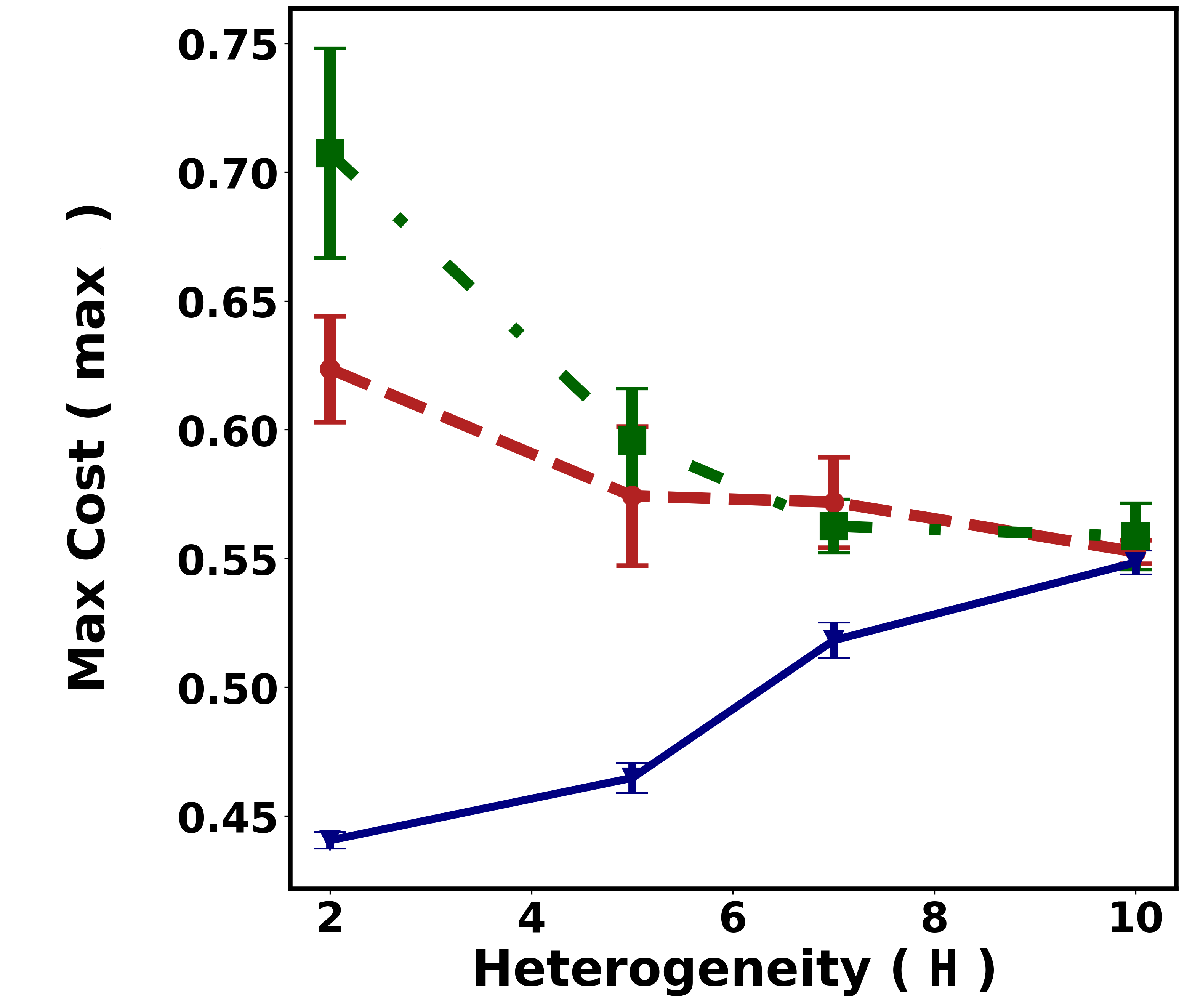}&\includegraphics[width=0.24\textwidth]{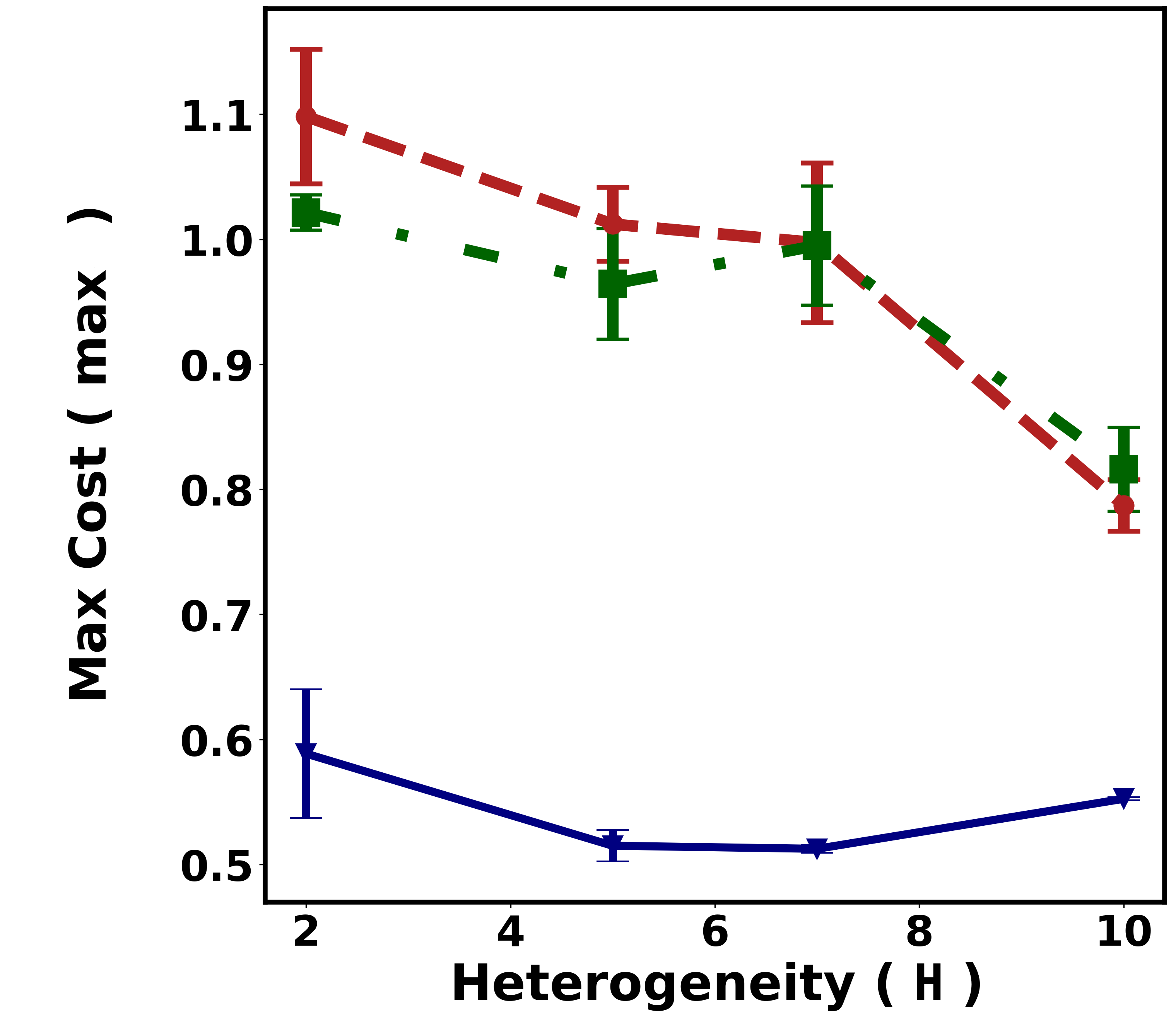}&\includegraphics[width=0.24\textwidth]{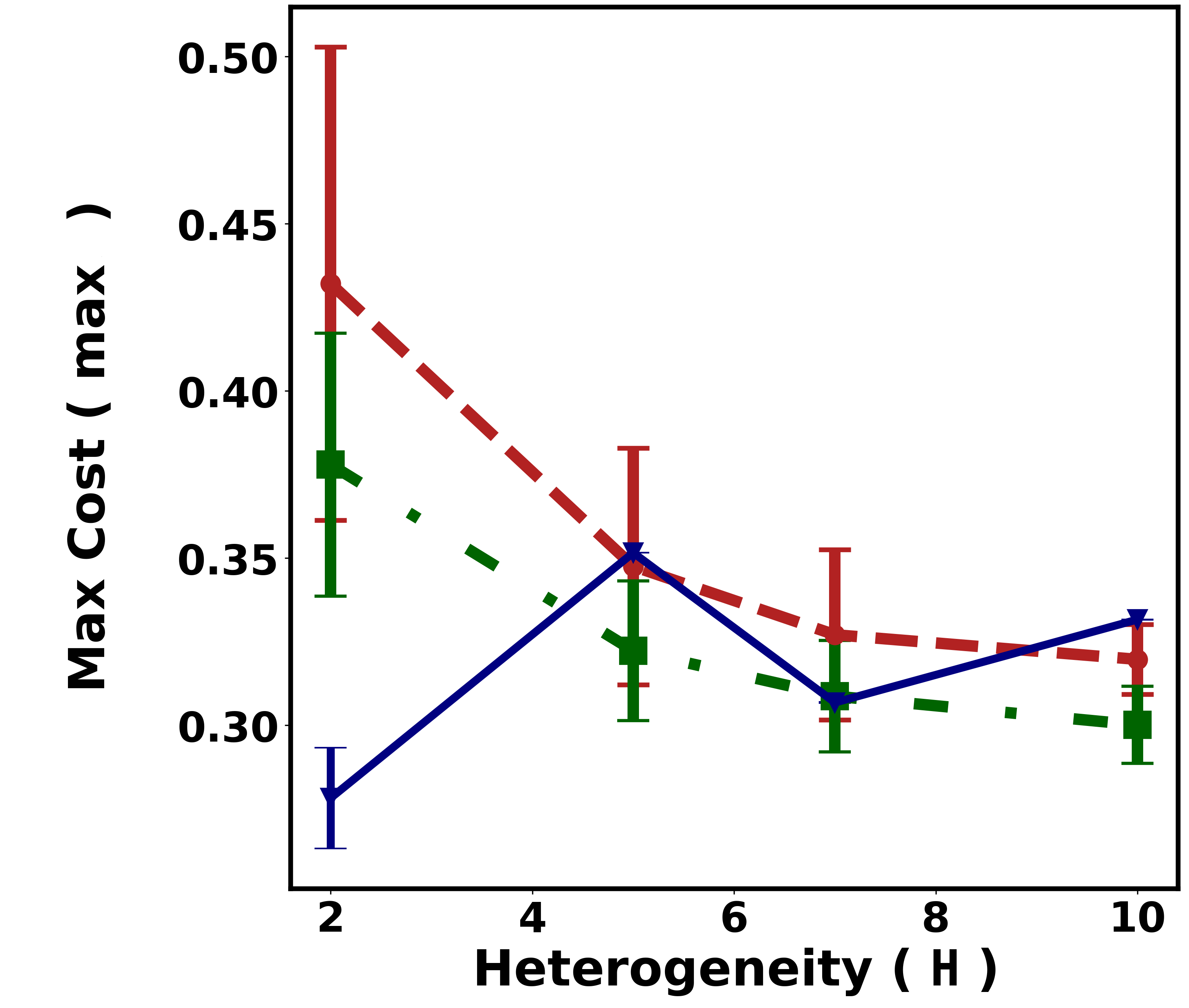}&\includegraphics[width=0.24\textwidth]{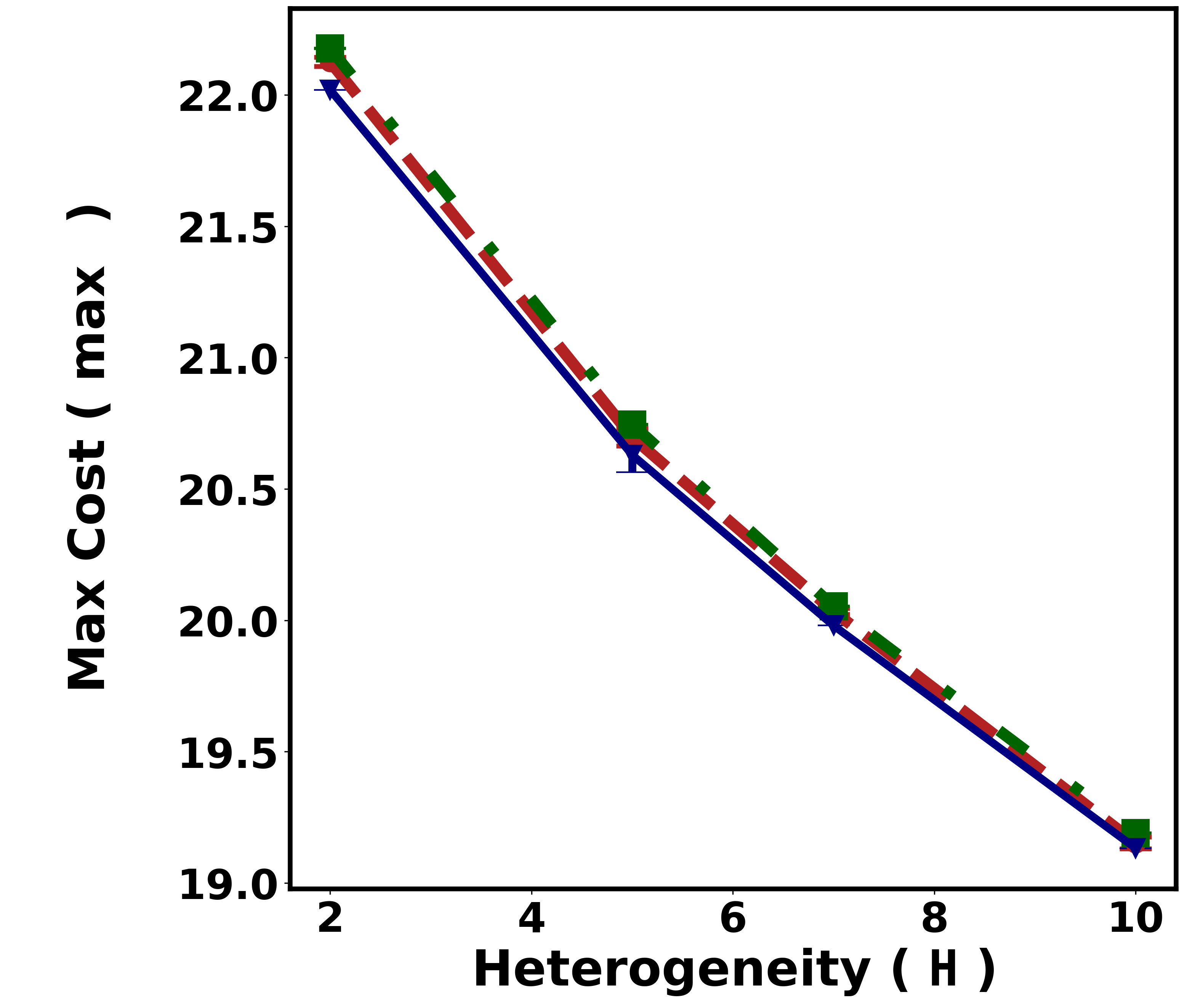} \\
     \multicolumn{4}{l}{\centering\includegraphics[width=0.75\textwidth]{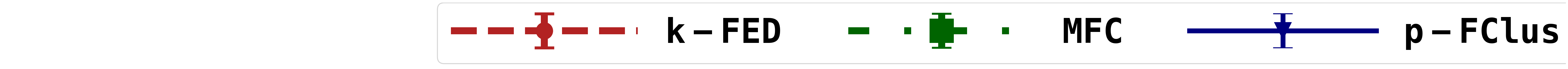}}
    \end{tabular}
      \caption{The plot shows the variation in evaluation metrics against proposed \ouralgo\ and \sota\ on $k$-means for varying heterogeneity levels on a \texttt{Balanced} data split across $100$ clients. Each column represents a dataset as specified at the top, and each row represents one metric under evaluation. Note that the \texttt{FMNIST} dataset is on $500$ clients. (Best viewed in color).}

        \label{fig:Balanced100clients}
    \end{figure*}

    \begin{figure*}[t!]
    \centering
    \begin{tabular}{@{}c@{}c@{}c@{}c@{}}
    \includegraphics[width=0.24\textwidth]{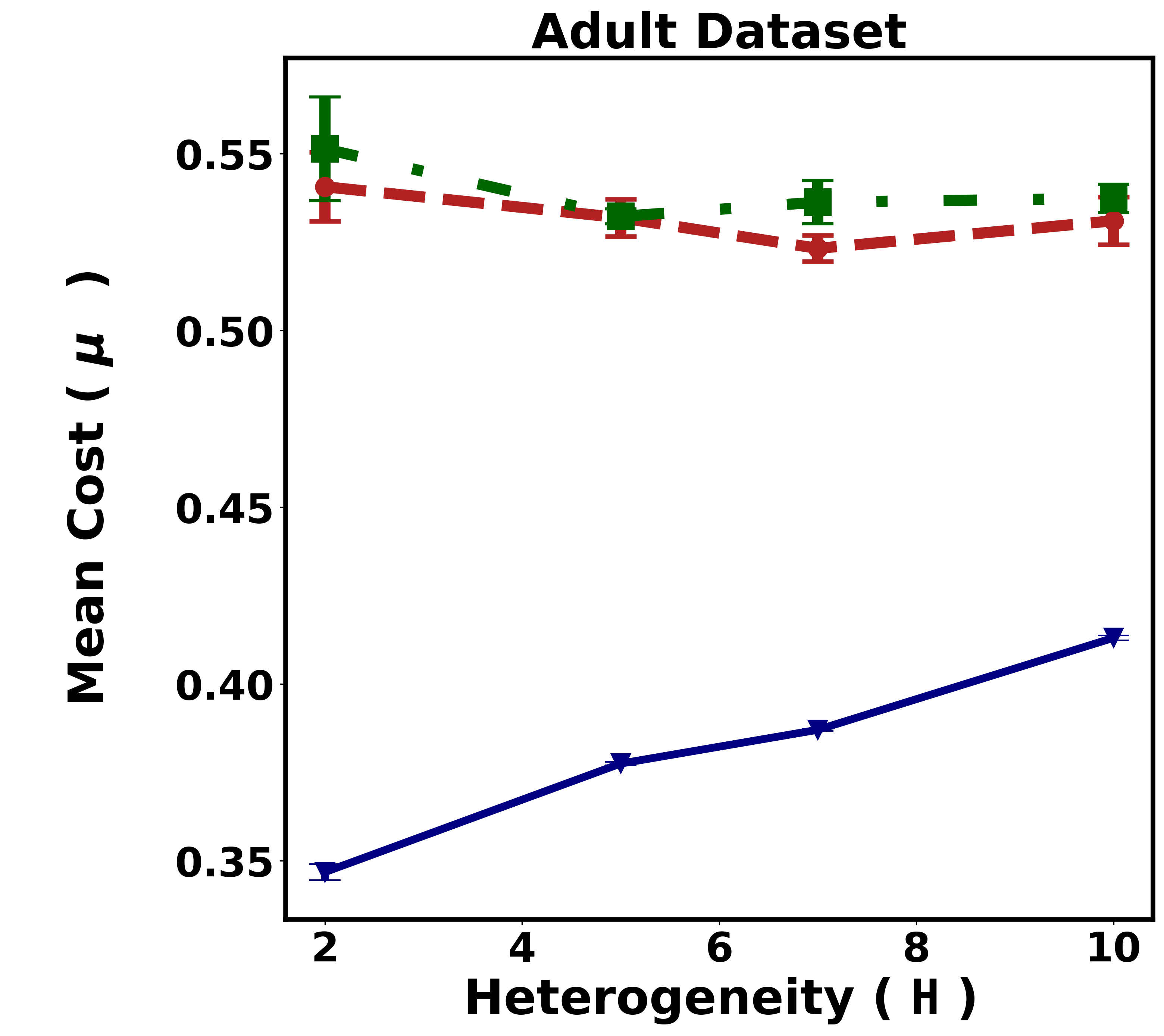}&\includegraphics[width=0.24\textwidth]{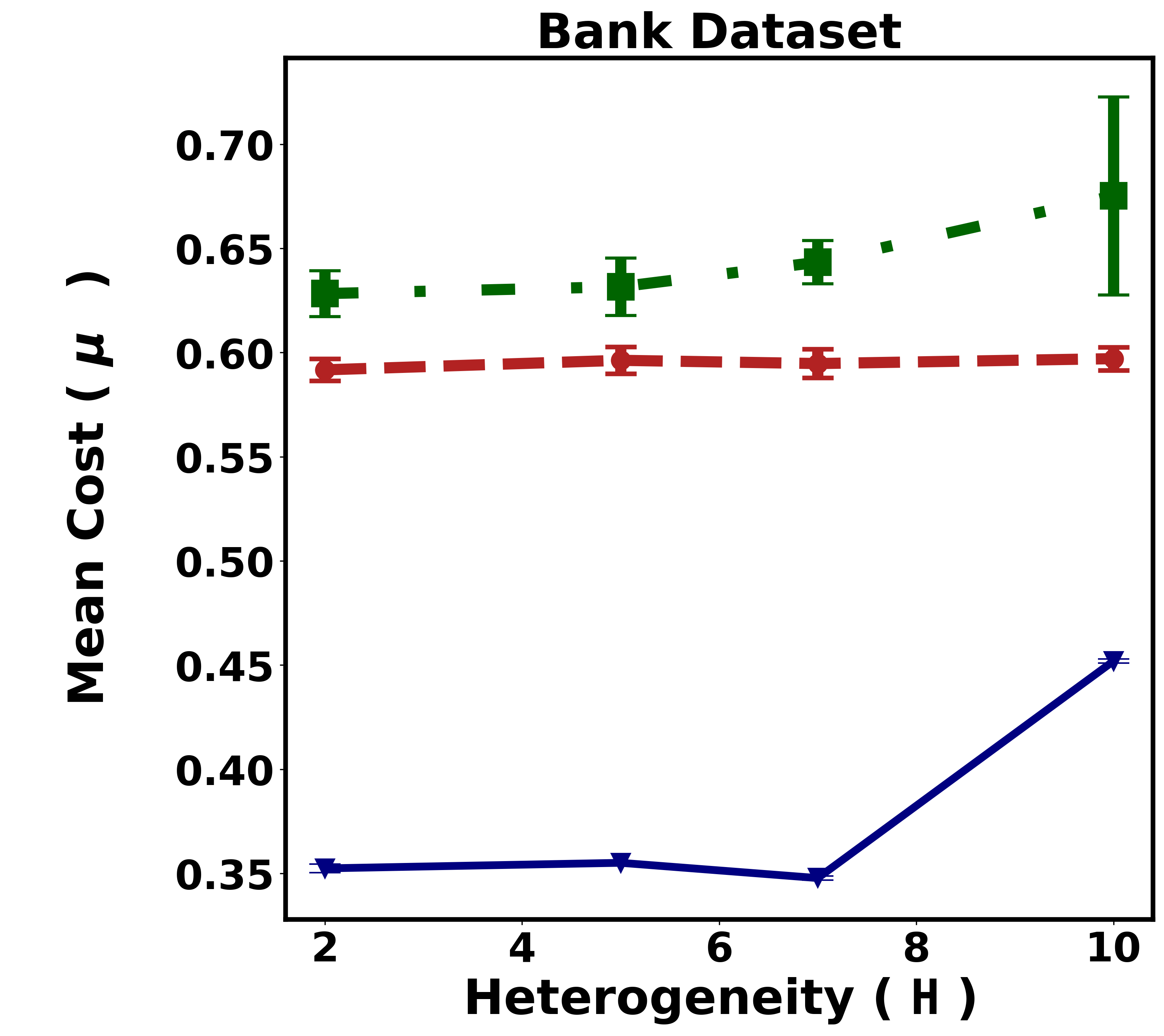}&\includegraphics[width=0.24\textwidth]{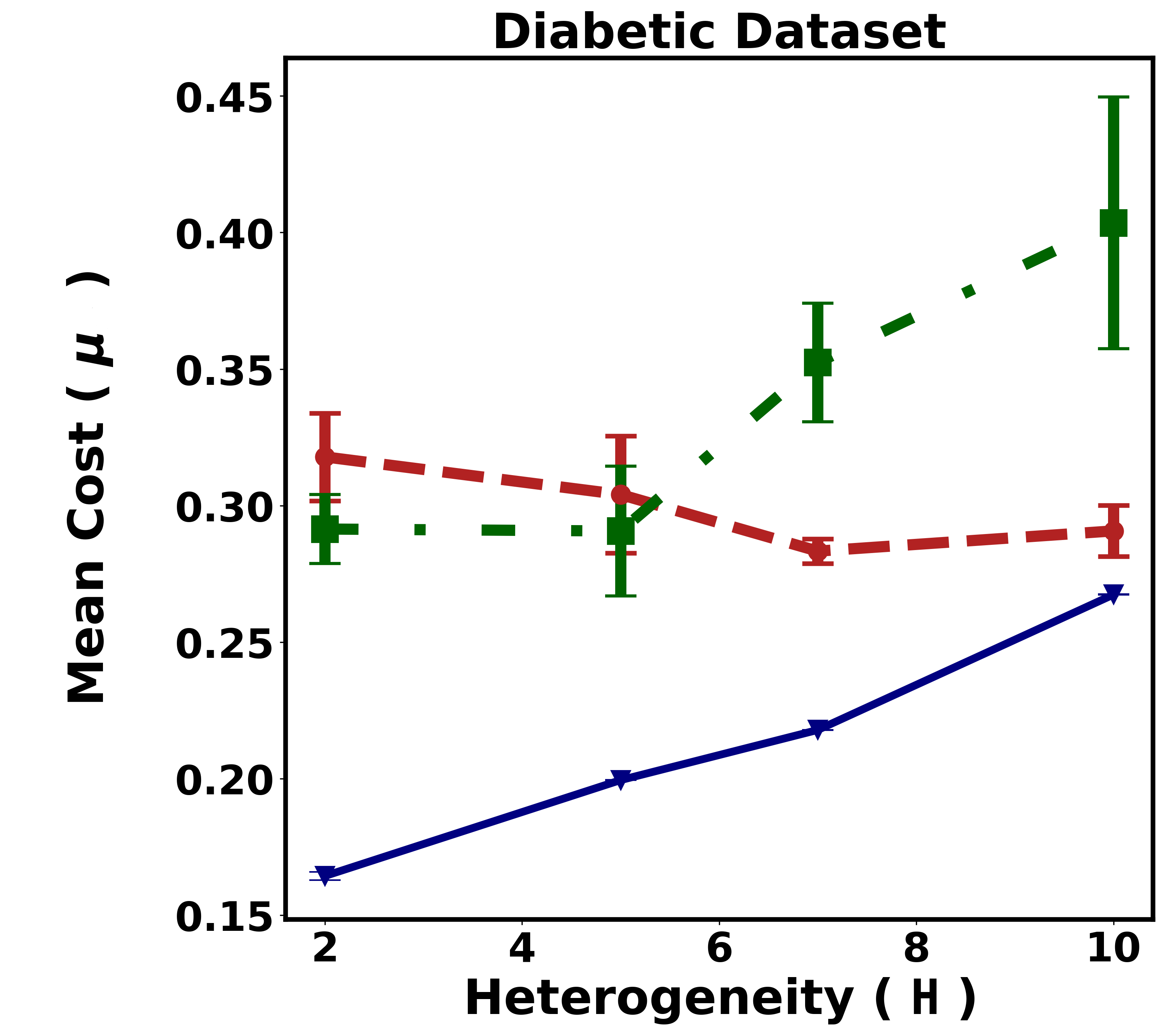}&\includegraphics[width=0.24\textwidth]{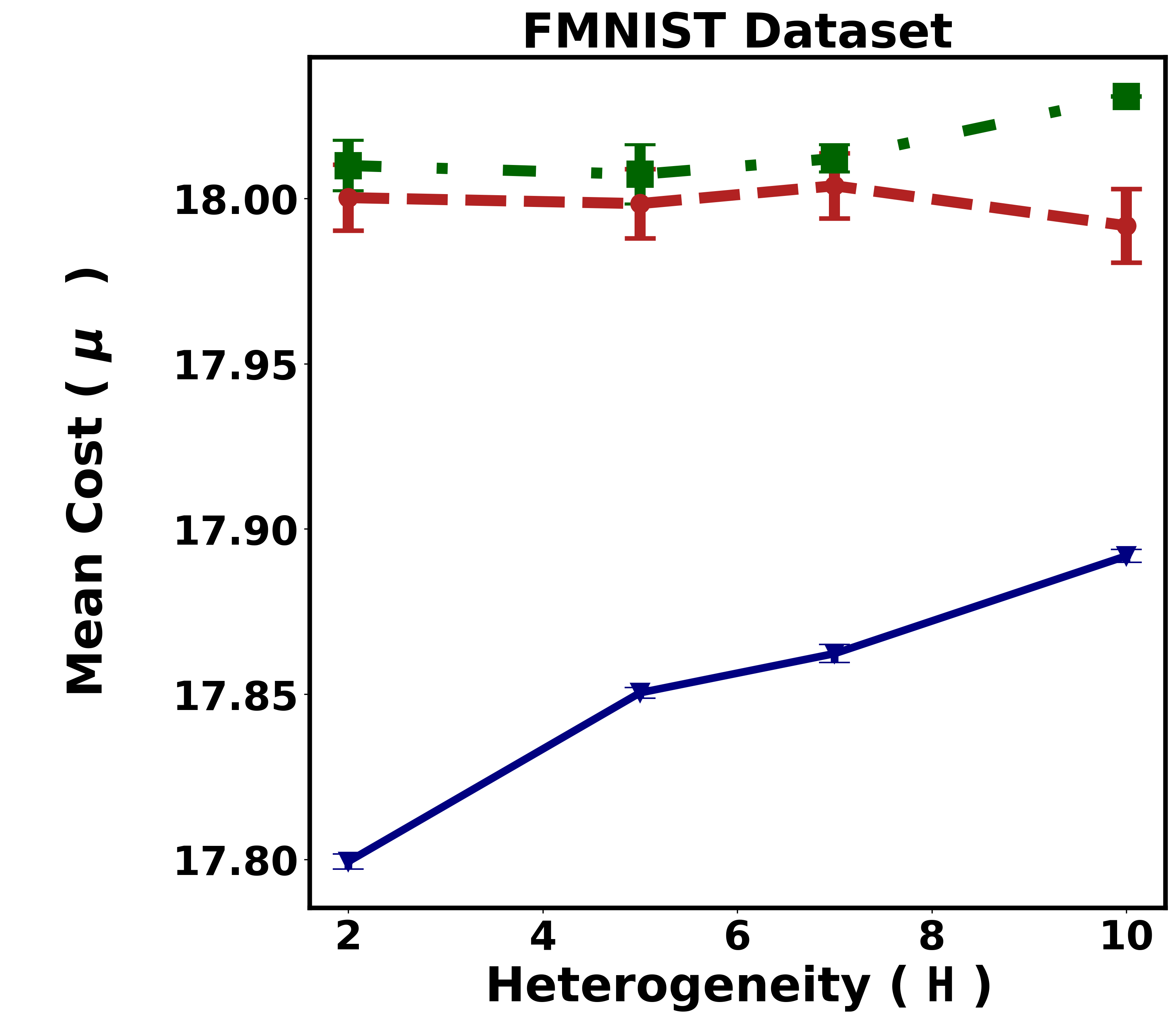}\\
    \includegraphics[width=0.24\textwidth]{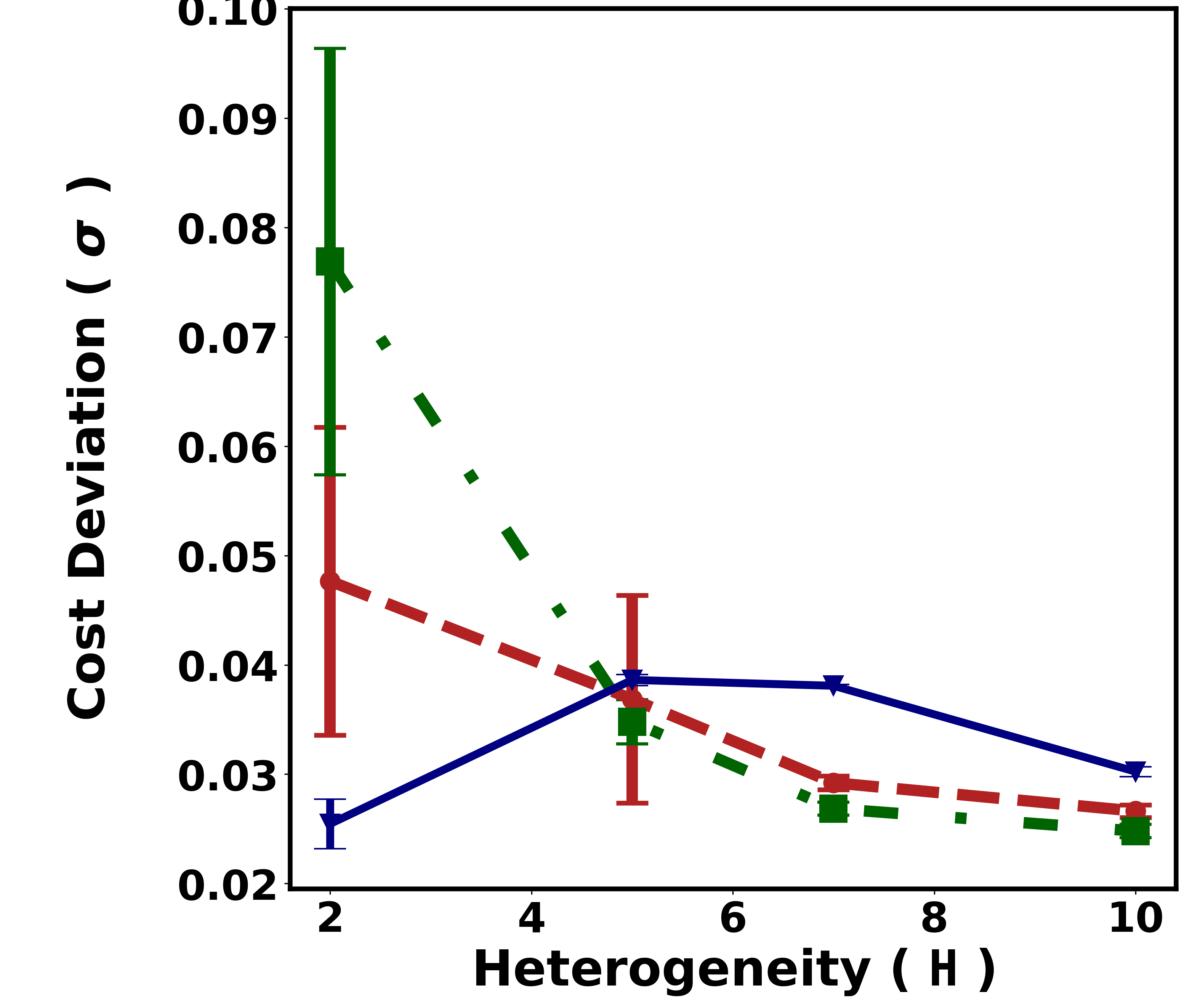}&\includegraphics[width=0.24\textwidth]{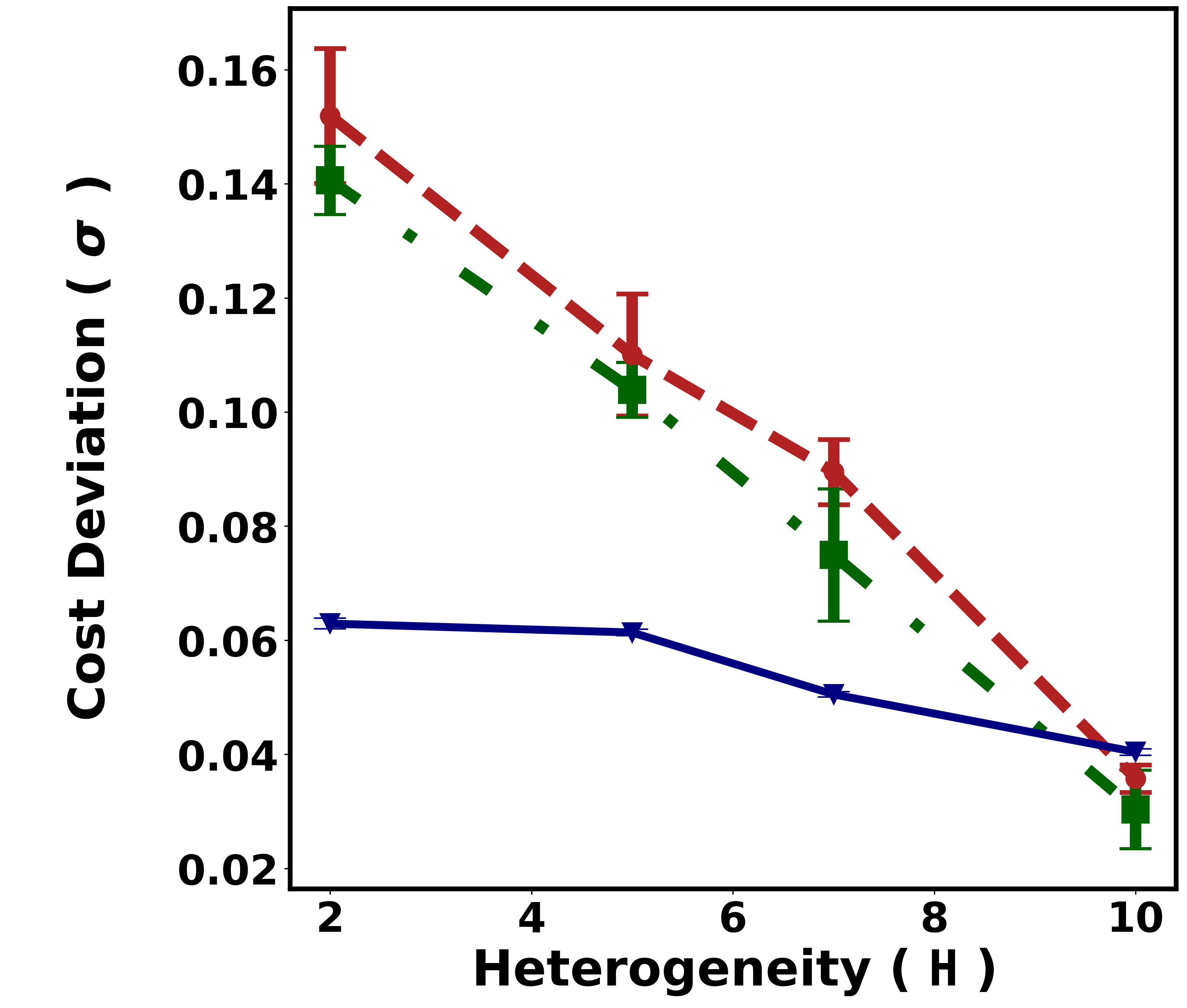}&\includegraphics[width=0.24\textwidth]{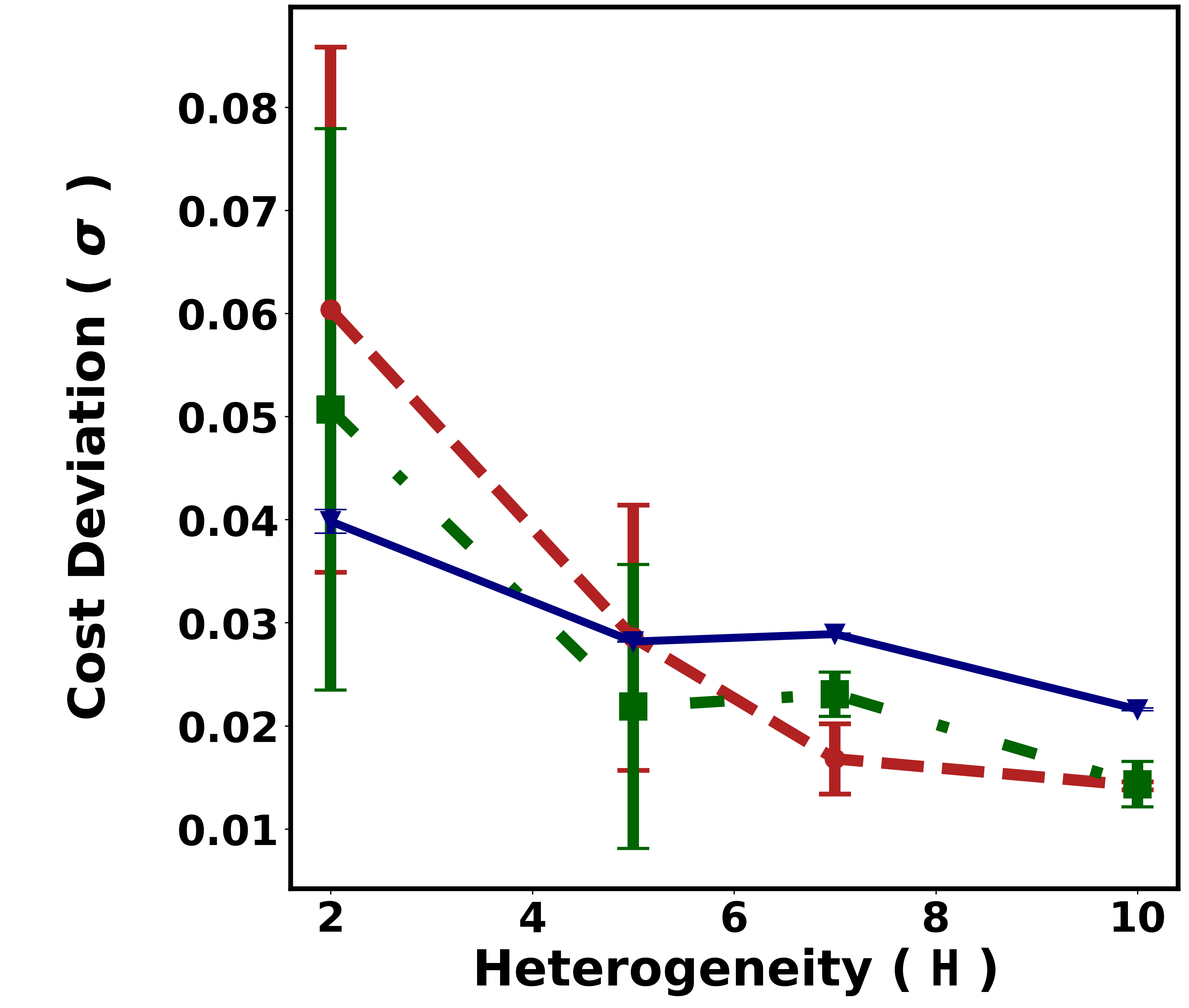}&\includegraphics[width=0.24\textwidth]{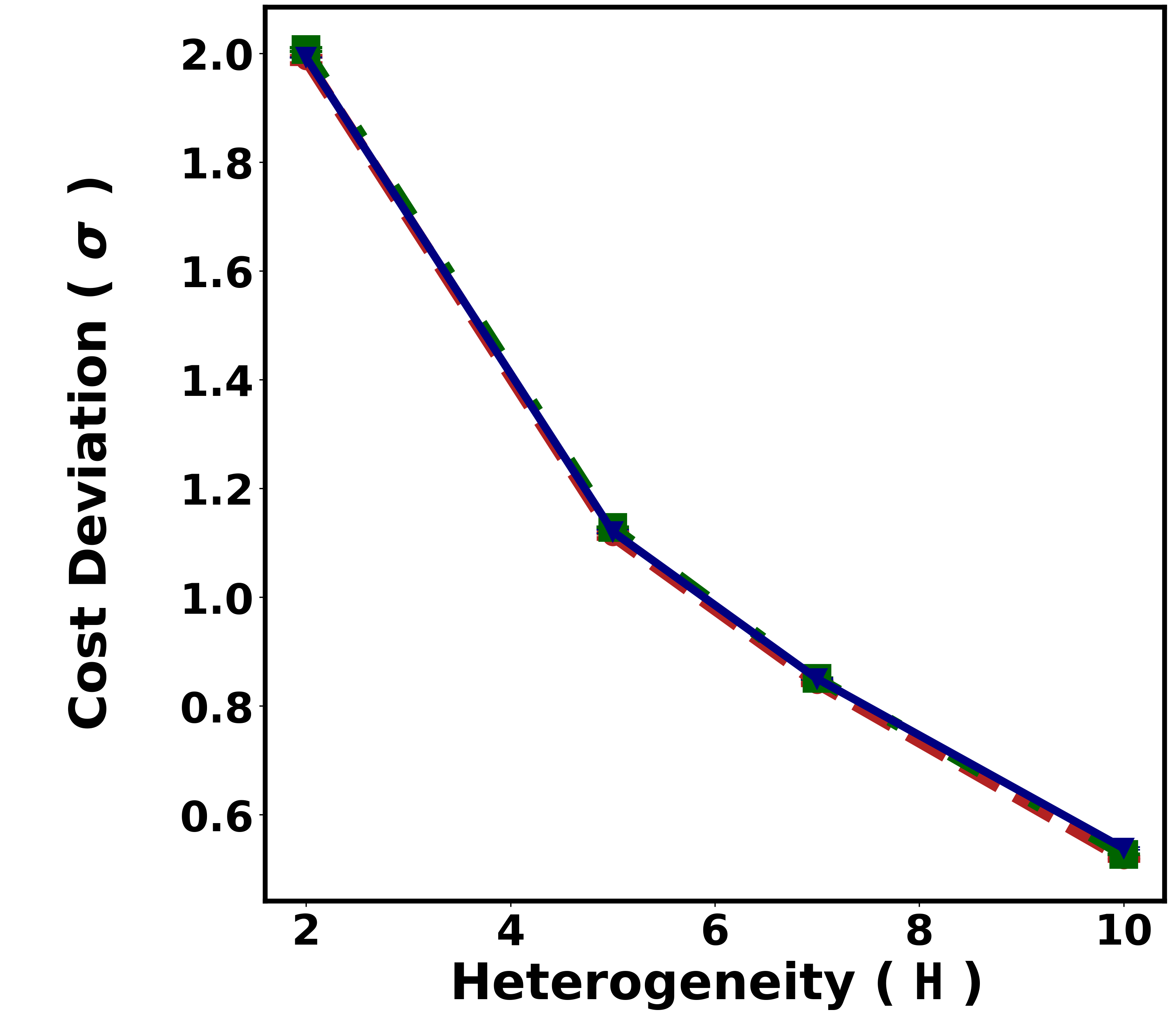}\\
    \includegraphics[width=0.24\textwidth]{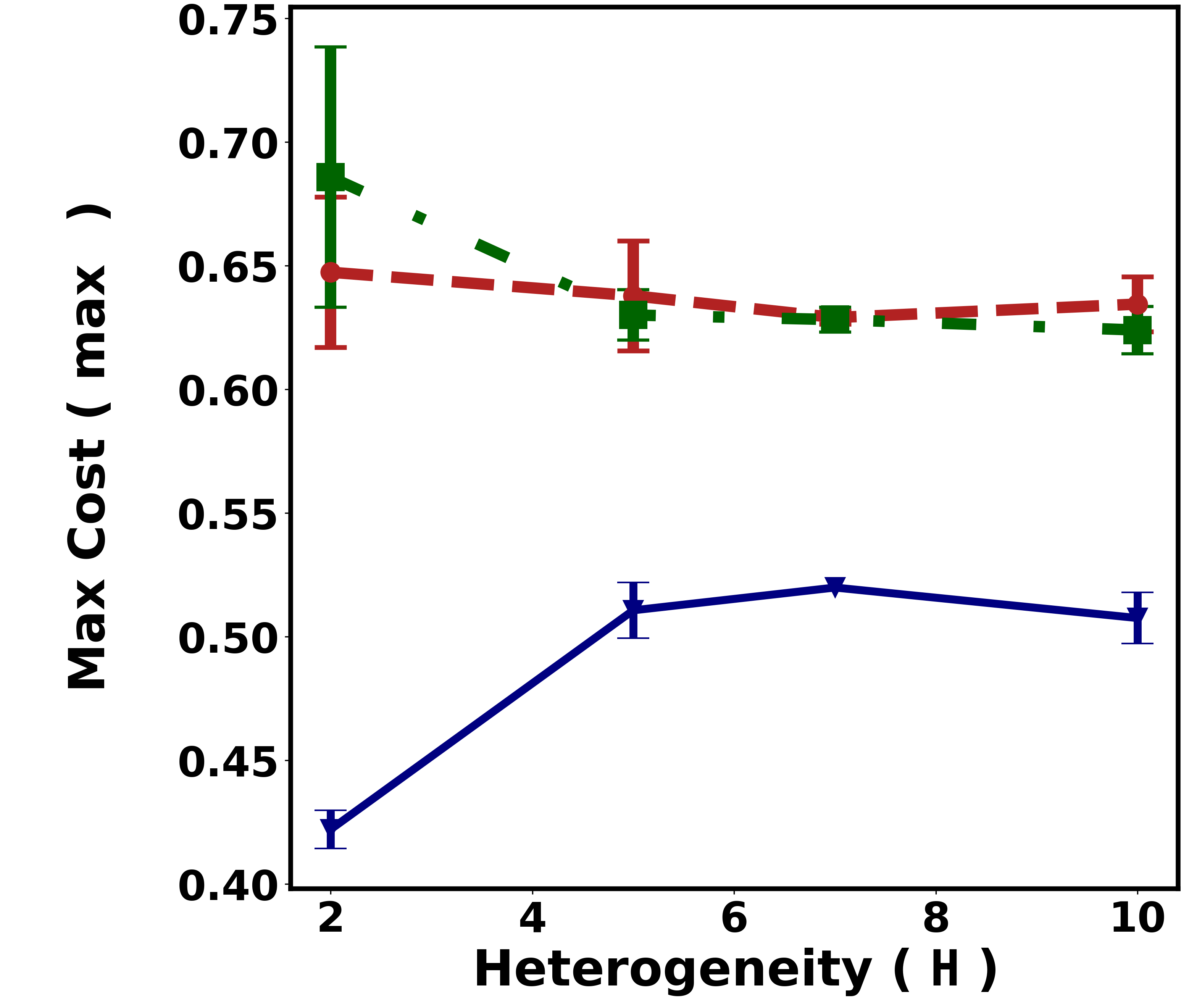}&\includegraphics[width=0.24\textwidth]{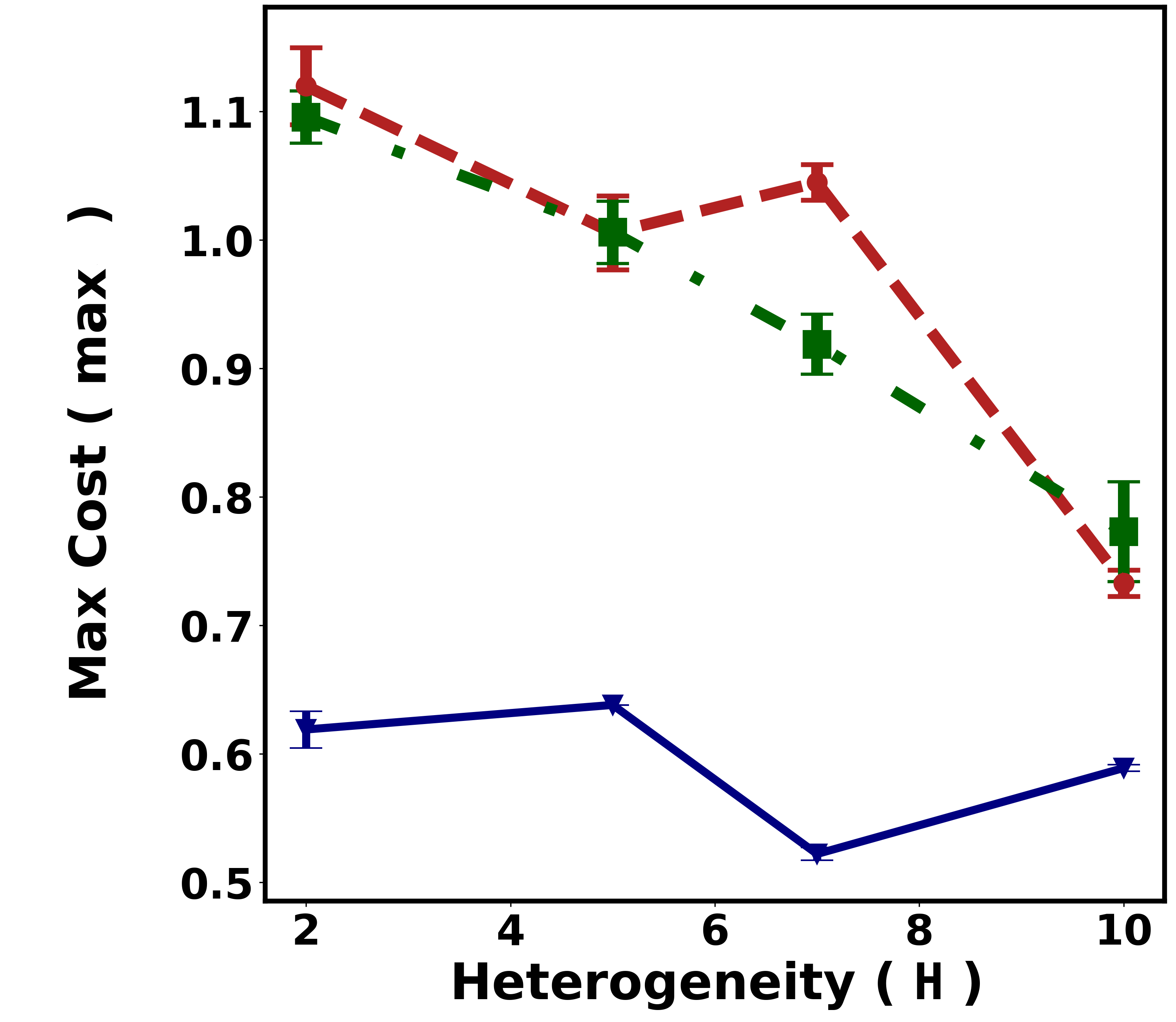}&\includegraphics[width=0.24\textwidth]{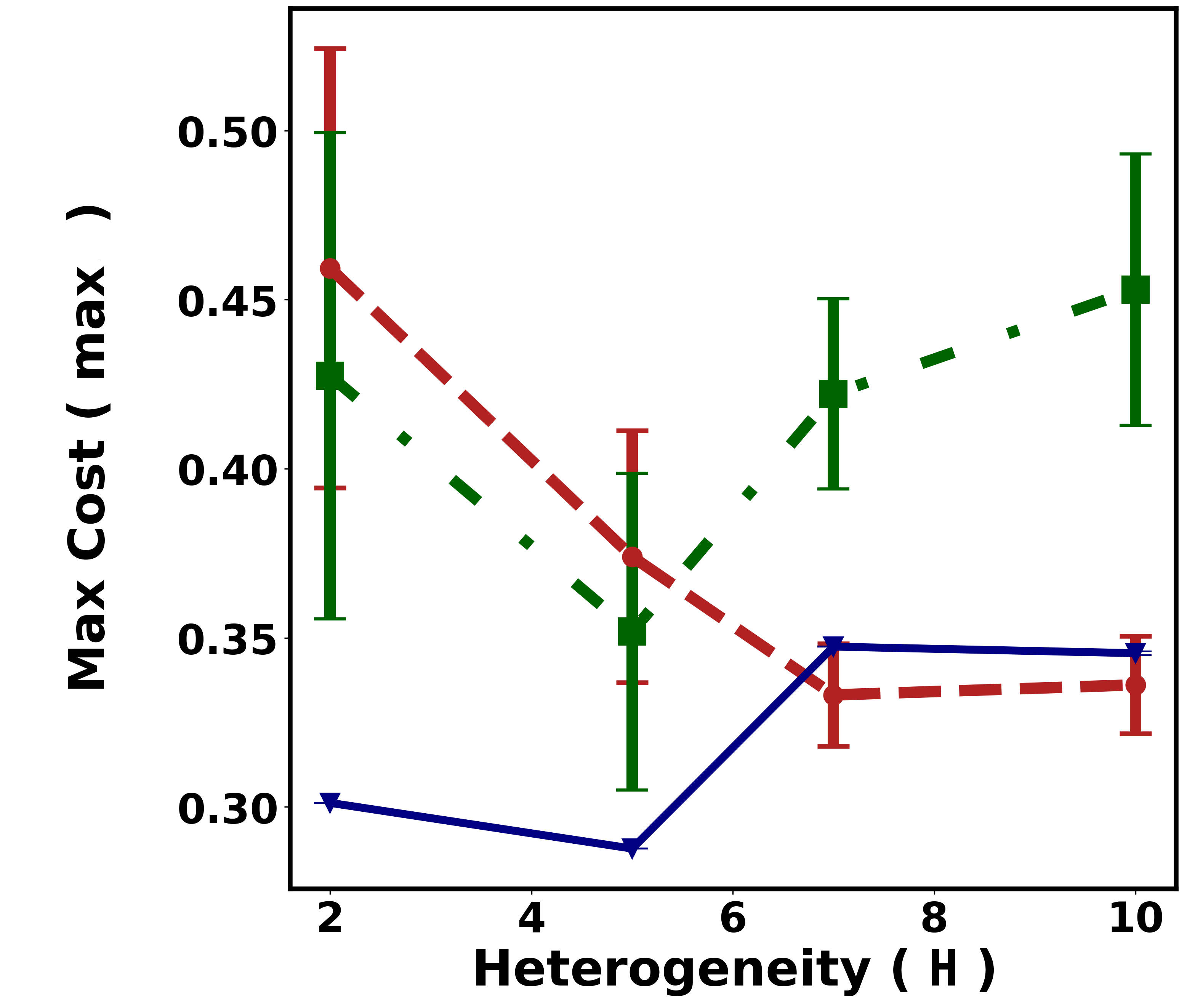}&\includegraphics[width=0.24\textwidth]{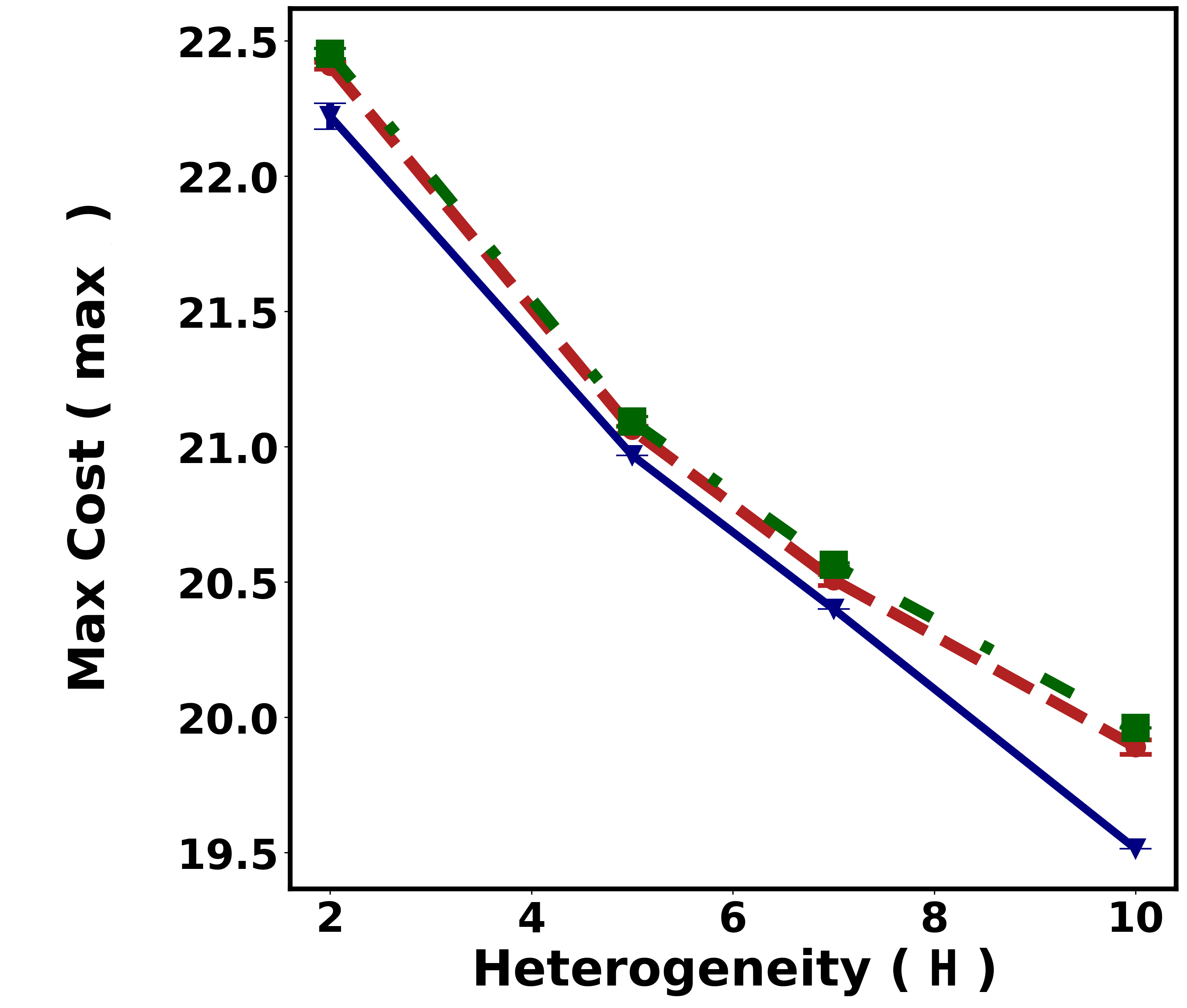} \\
           \multicolumn{4}{l}{\centering\includegraphics[width=0.75\textwidth]{images_final/label_long.png}}
    \end{tabular}
      \caption{The plot shows the variation in evaluation metrics against proposed \ouralgo\ and \sota\ on $k$-means objective for varying heterogeneity levels on a \texttt{Balanced} data split across $1000$ clients. Each column represents a specific dataset as specified at the top, and each row represents one metric under evaluation. (Best viewed in color).}
        \label{fig:Balanced1000clients}
    \end{figure*}


\begin{enumerate}
    \item \texttt{Balanced} or (Equal) distribution: In this setting, the random data points from each $\mathcal{D}_k$ distribution are equally divided among clients. Note that in scenarios where the total number of data points from $\mathcal{D}_k$ is not divisible by the number of clients, each client will have nearly equal data points or may have one less data point compared to other clients.
    
    \item \texttt{Unequal} distribution: In this setting, each client can have a different number of data points from each of the $\mathcal{D}_k$ distributions. Instead of arbitrarily dividing the data points among clients, we intelligently distribute the data points to capture scenarios where some clients may have significantly fewer data points from certain $D_k$, resulting in an overall skewed division among clients. The code repository contains scripts for generating the same$^6$.
\end{enumerate}

    \begin{figure*}[t!]
    \centering
    \begin{tabular}{@{}c@{}c@{}c@{}c@{}}
   \includegraphics[width=0.24\textwidth]{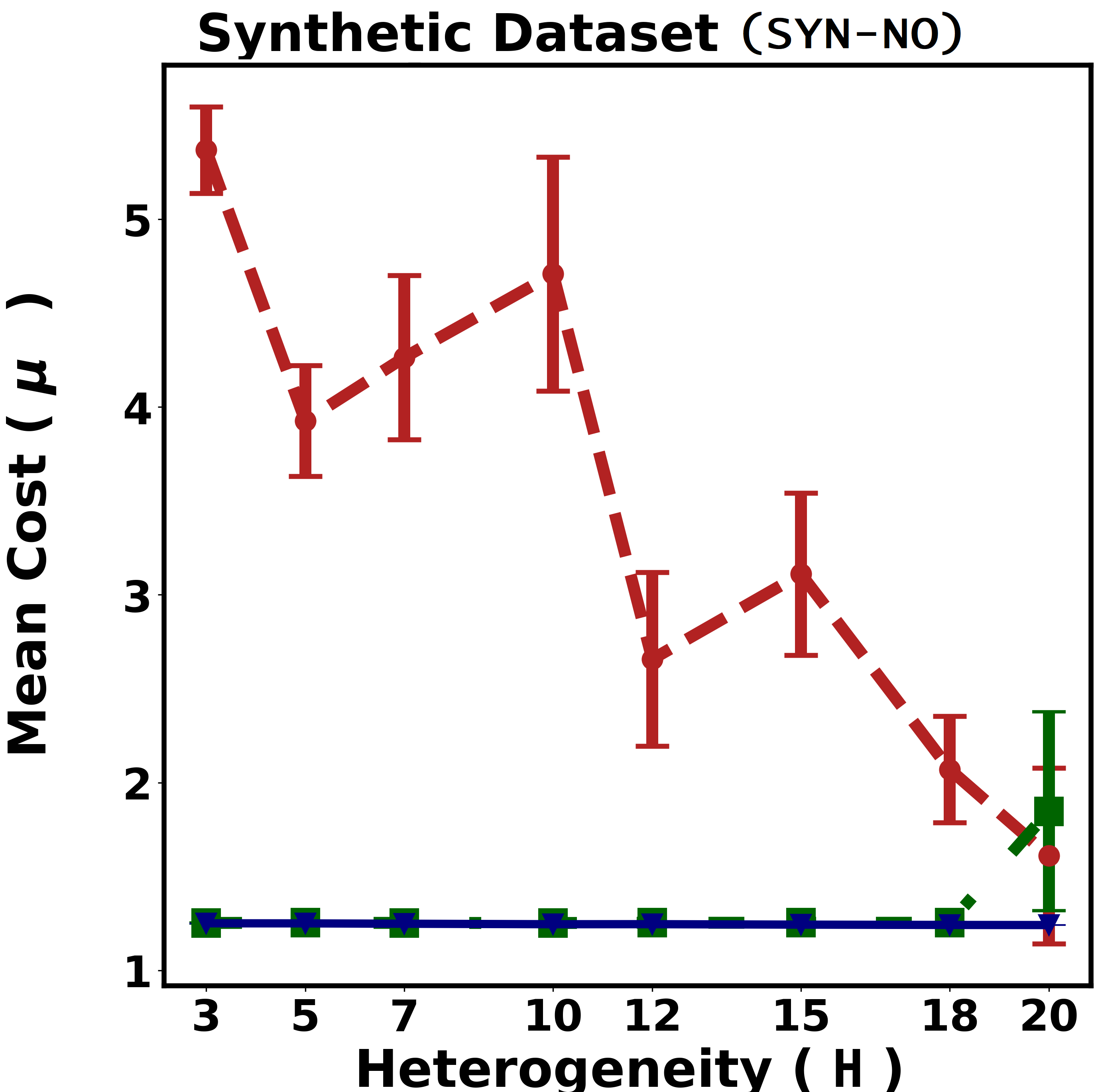}&\hspace {4pt}\includegraphics[width=0.24\textwidth]{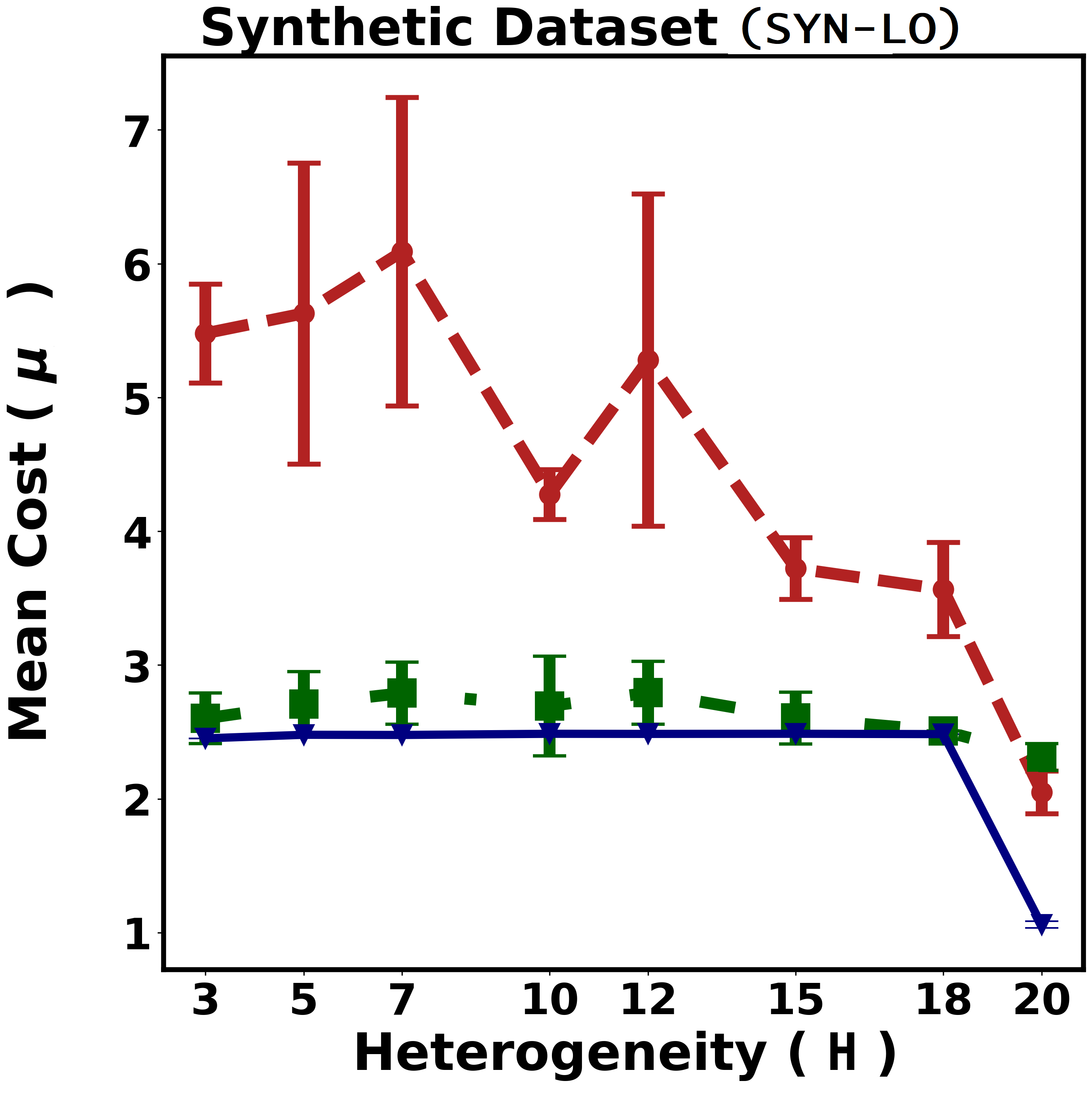}&\includegraphics[width=0.25\textwidth]{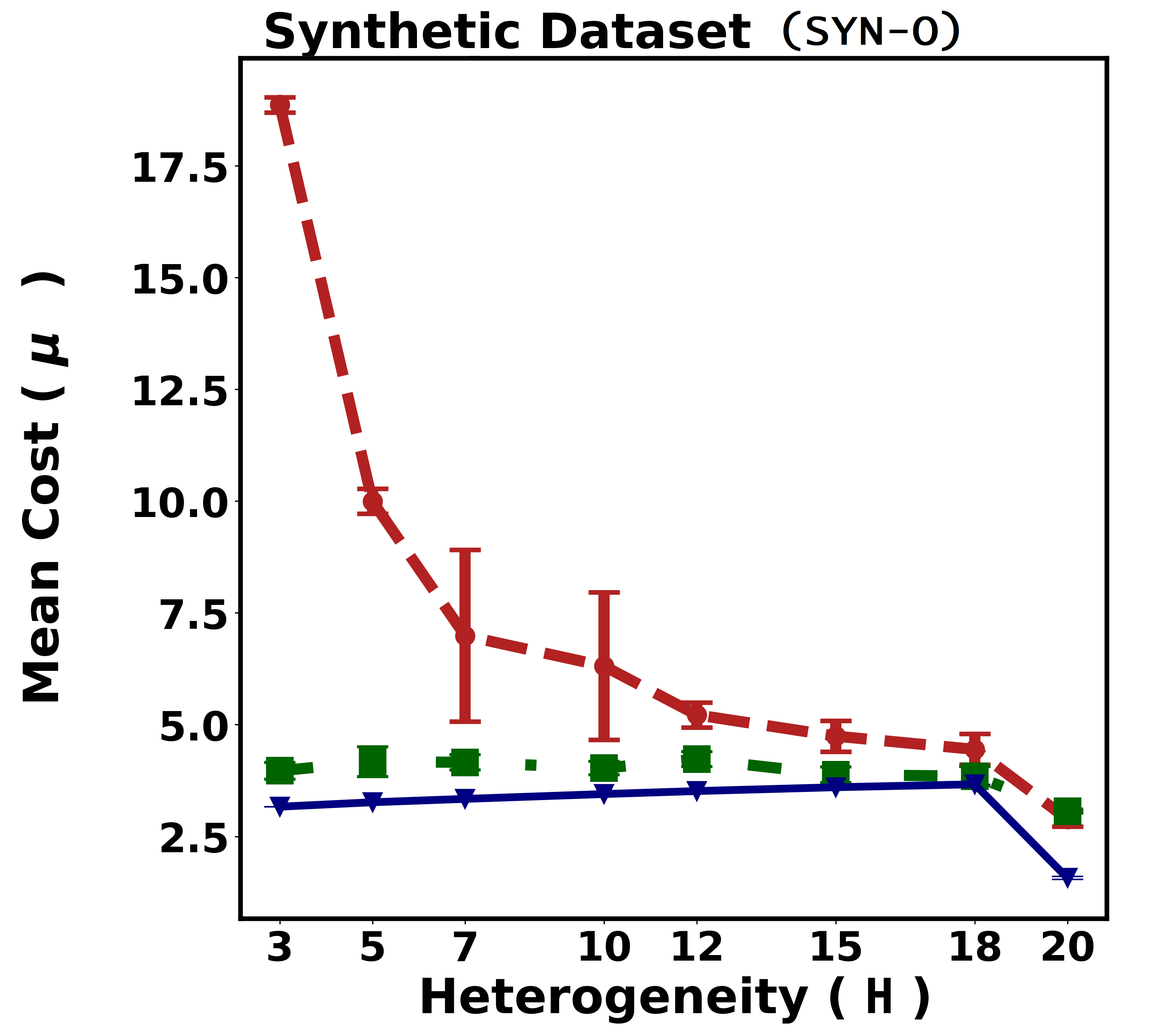}\\
    \includegraphics[width=0.24\textwidth]{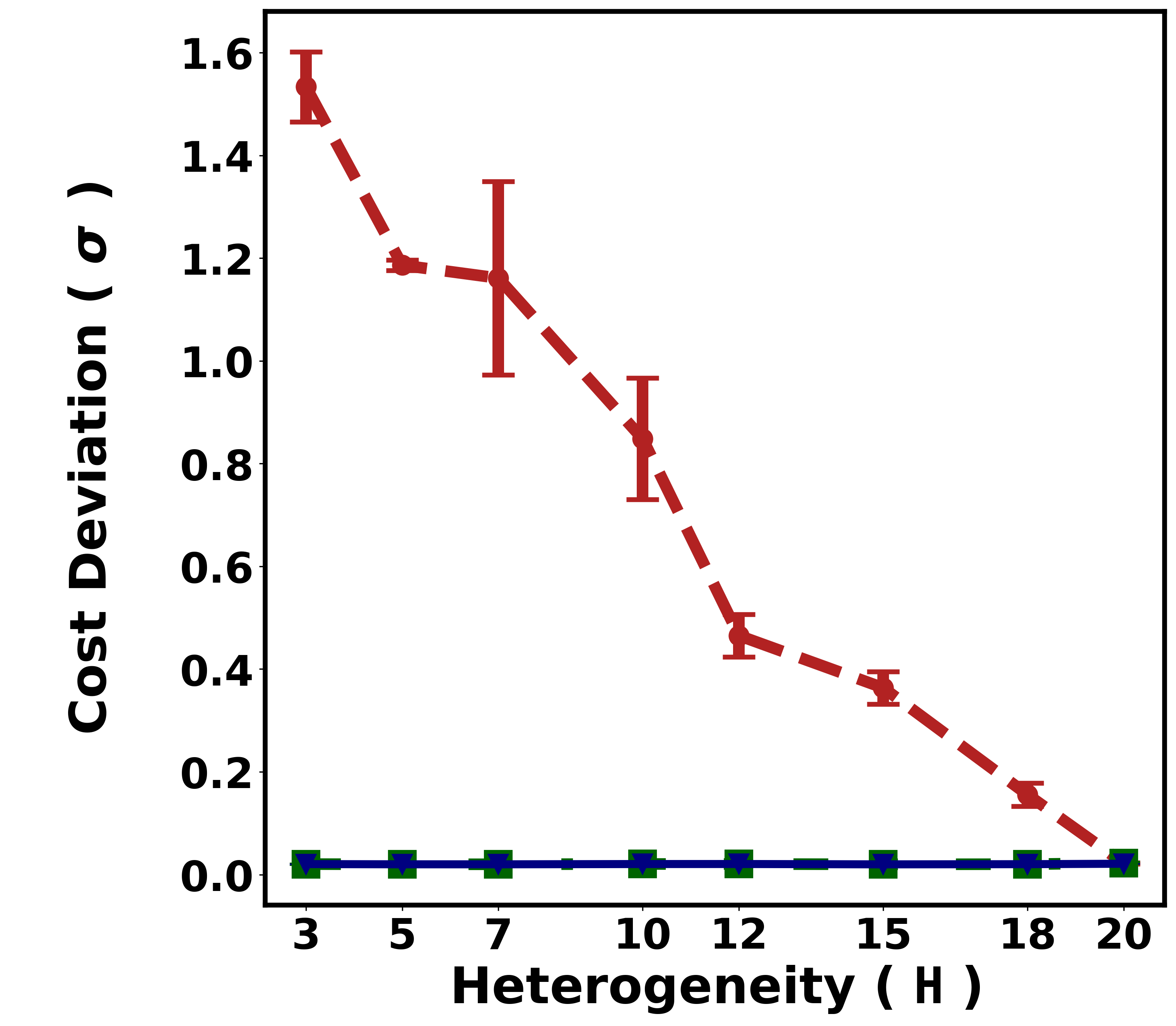}&\includegraphics[width=0.24\textwidth]{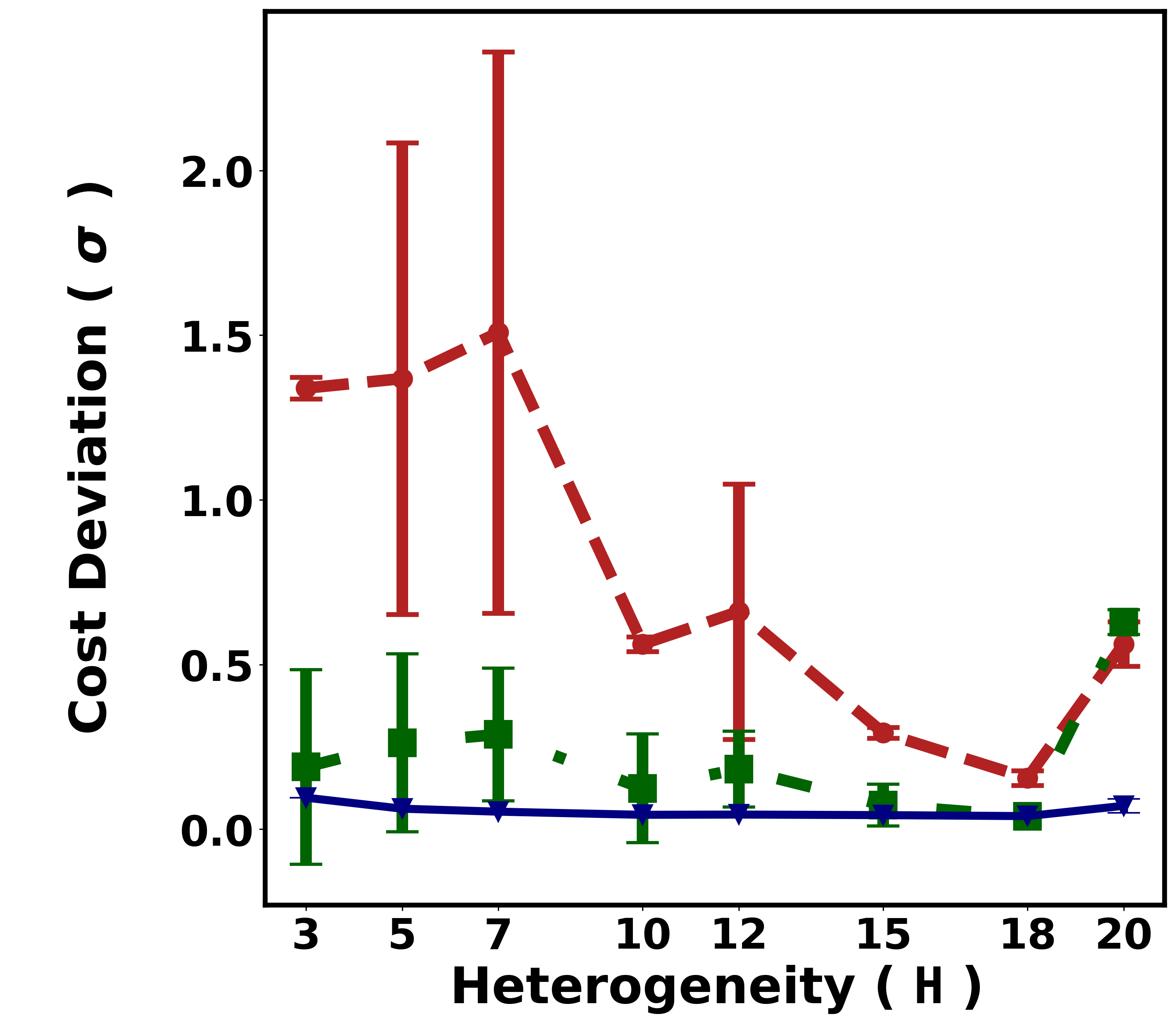}&\includegraphics[width=0.25\textwidth]{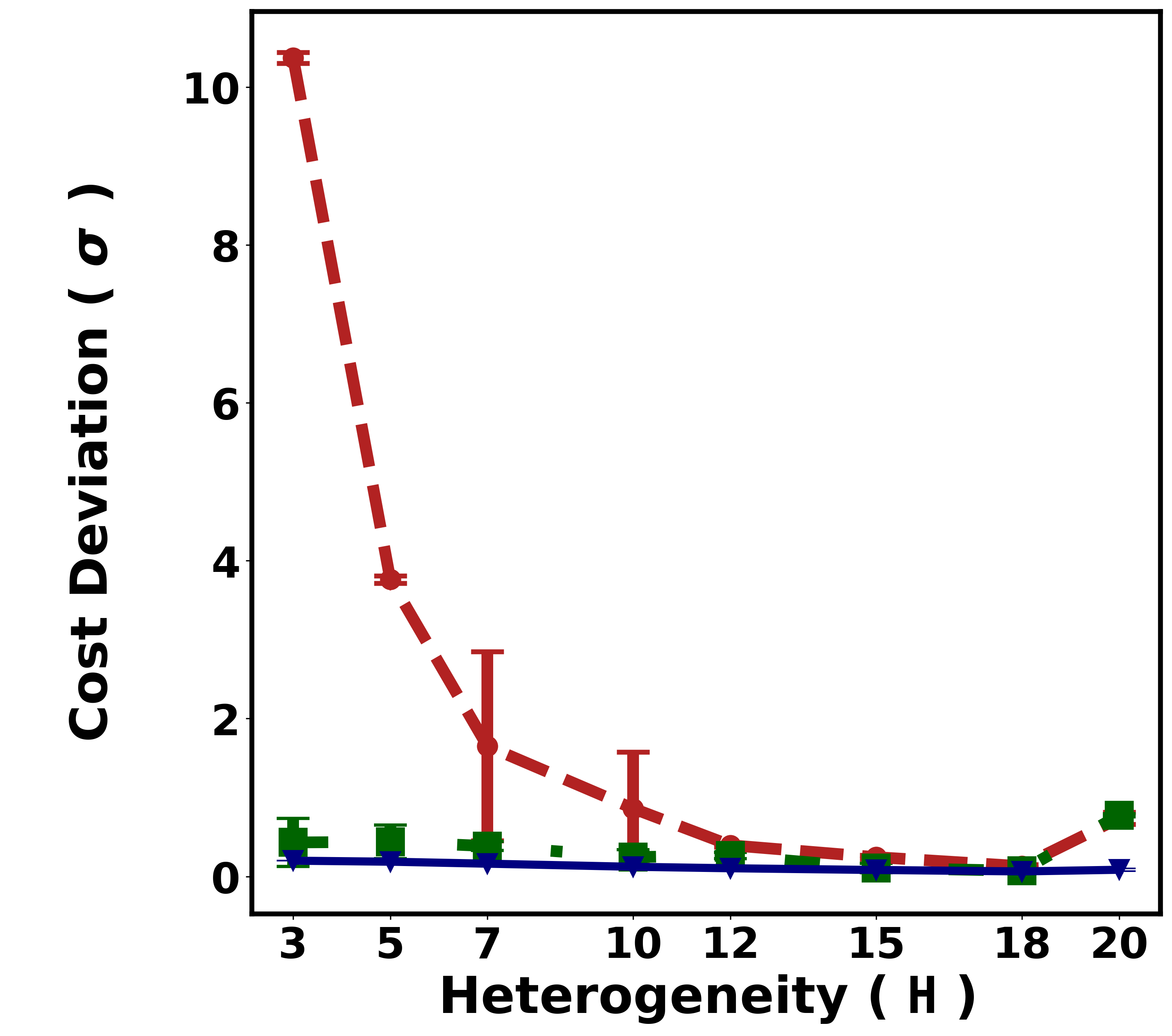}\\
    \includegraphics[width=0.24\textwidth]{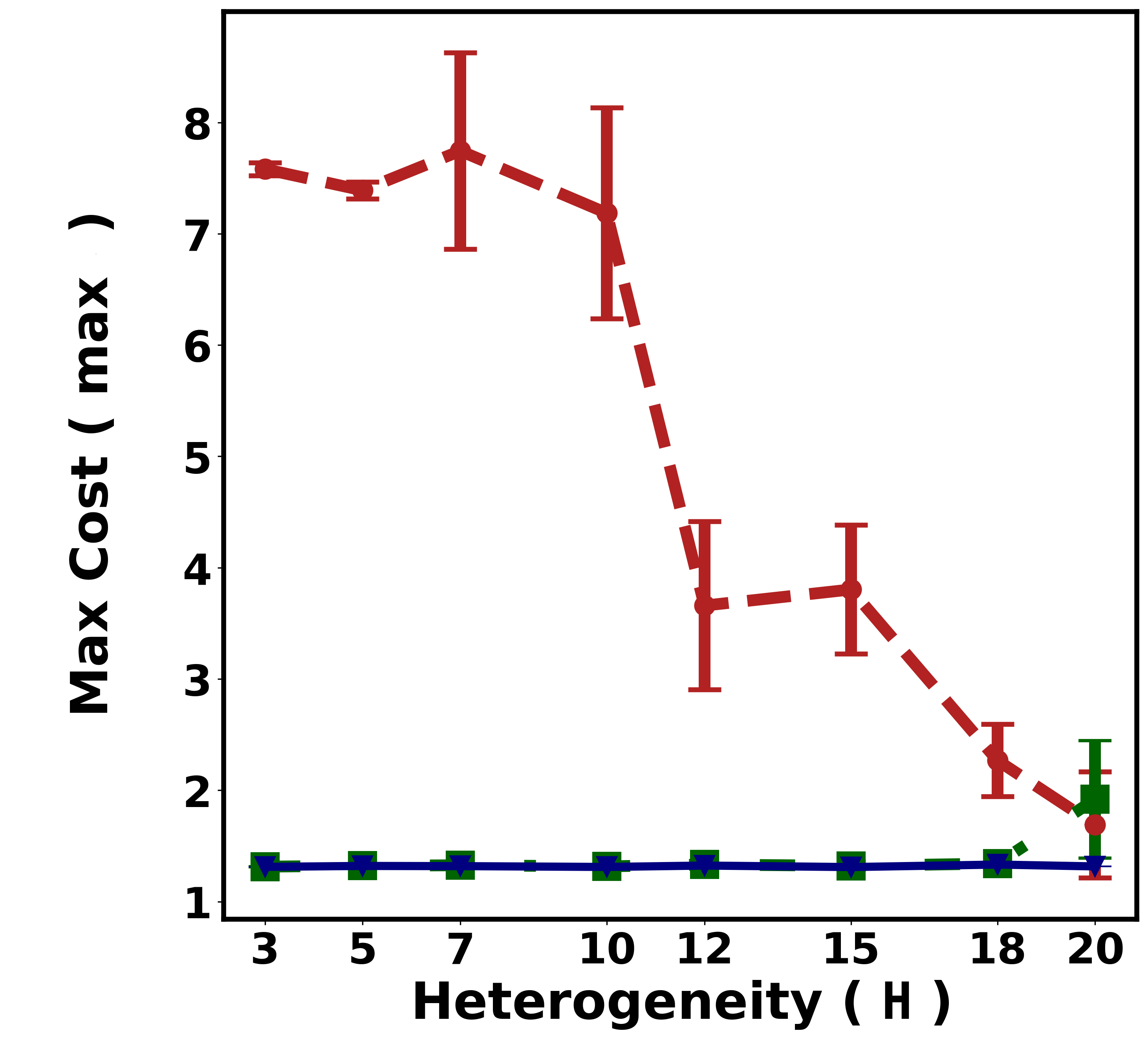}&\includegraphics[width=0.24\textwidth]{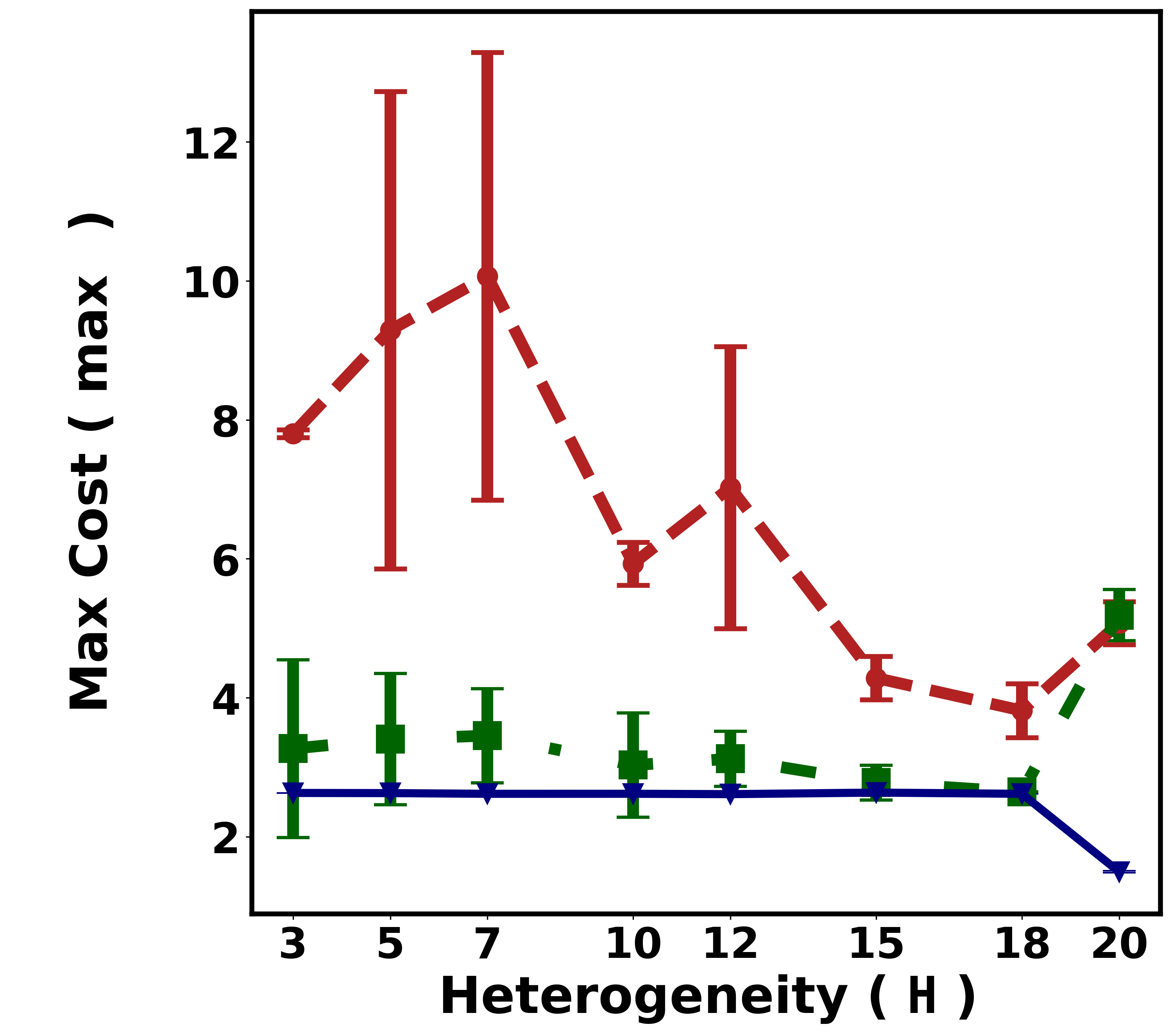}&\includegraphics[width=0.25\textwidth]{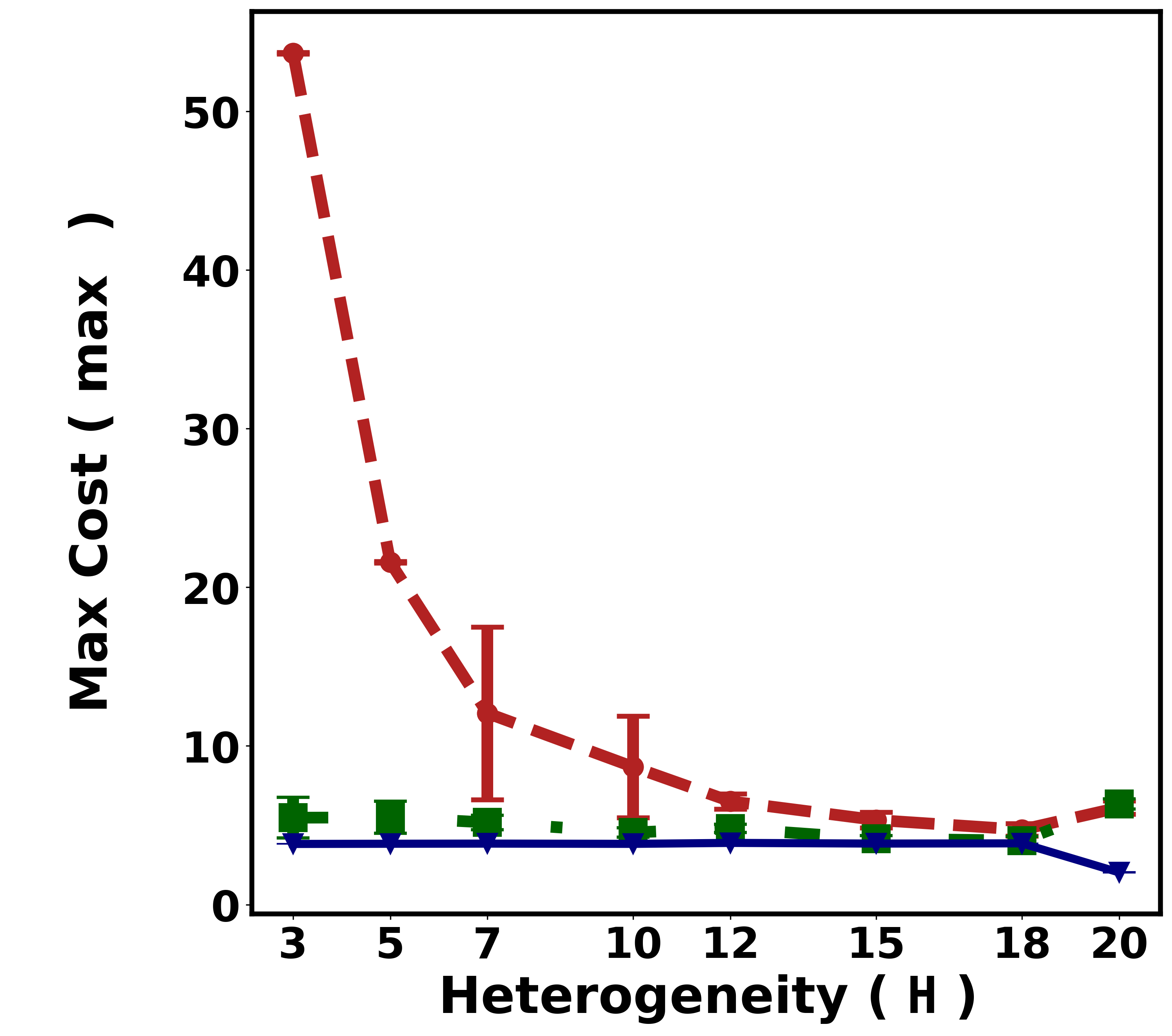}\\
      \multicolumn{3}{l}{\centering\includegraphics[width=0.55\textwidth]{images_final/label_long.png}}
    \end{tabular}
      \caption{The plot shows the variation in evaluation metrics against proposed \ouralgo\ and \sota\ on $k$-means objective for varying heterogeneity levels on a \texttt{Balanced} data split across $1000$ clients. Each column represents a specific Synthetic dataset (\texttt{Syn}) in sequence: \texttt{Syn-NO}, \texttt{Syn-LO}, \texttt{Syn-O} respectively, and each row represents one metric under evaluation. (Best viewed in color).}
        \label{fig:Balanced1000clientsSynthe}
    \end{figure*}


\noindent Within each of the above settings, we further consider the level of heterogeneity ($\mathtt{H}$) as $2, 5, 7$ and  $10$ (Here, $10$ is the maximum number of distributions in non-federated datasets under consideration).  This implies that if, for instance, the $\mathtt{H}$ = $5$, then every client will contain data points only from any of the $5\ (\le k)$ distributions. 

  We now begin by validating our \ouralgo\ against \sota\ on a \texttt{Balanced} distribution setting and non-federated datasets. Later, we will explore the scenario where clients can have an \texttt{Unequal} data distribution and then on intrinsic or fixed heterogeneity federated datasets (\texttt{FEMNIST} and \texttt{WISDM}).

\subsection{Analysis on \texttt{Balanced} Data Distribution among Clients on $k$-means Objective}
This subsection delves into the results of the balanced (or equal) data distribution setting in the $k$-means objective ($\ell\text{-norm }= 2$). The results are illustrated in Fig. \ref{fig:Balanced100clients} (Real-world dataset, $100$ clients),  Fig.  \ref{fig:Balanced1000clients} (Real-world dataset, $1000$ clients) and  Fig. \ref{fig:Balanced1000clientsSynthe} (\texttt{Syn} dataset)
. We first provide a brief overview of the observations for each dataset in the subsections below and then conclude the overall results in this setting after the last subsection.

   \begin{figure*}[ht!]
    \centering
    \begin{tabular}{@{}c@{}c@{}c@{}c@{}}
    \includegraphics[width=0.25\textwidth]{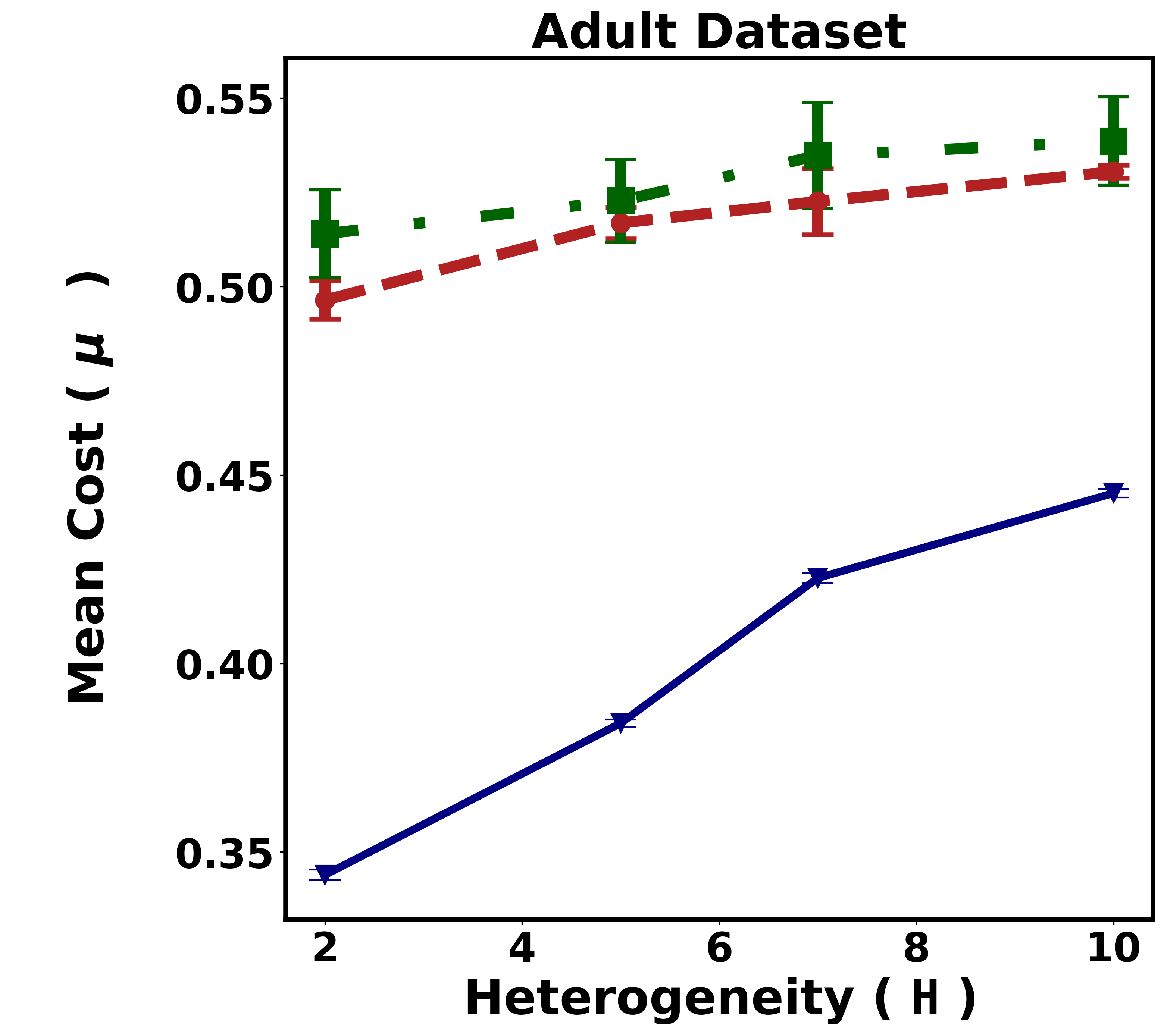}&\includegraphics[width=0.25\textwidth]{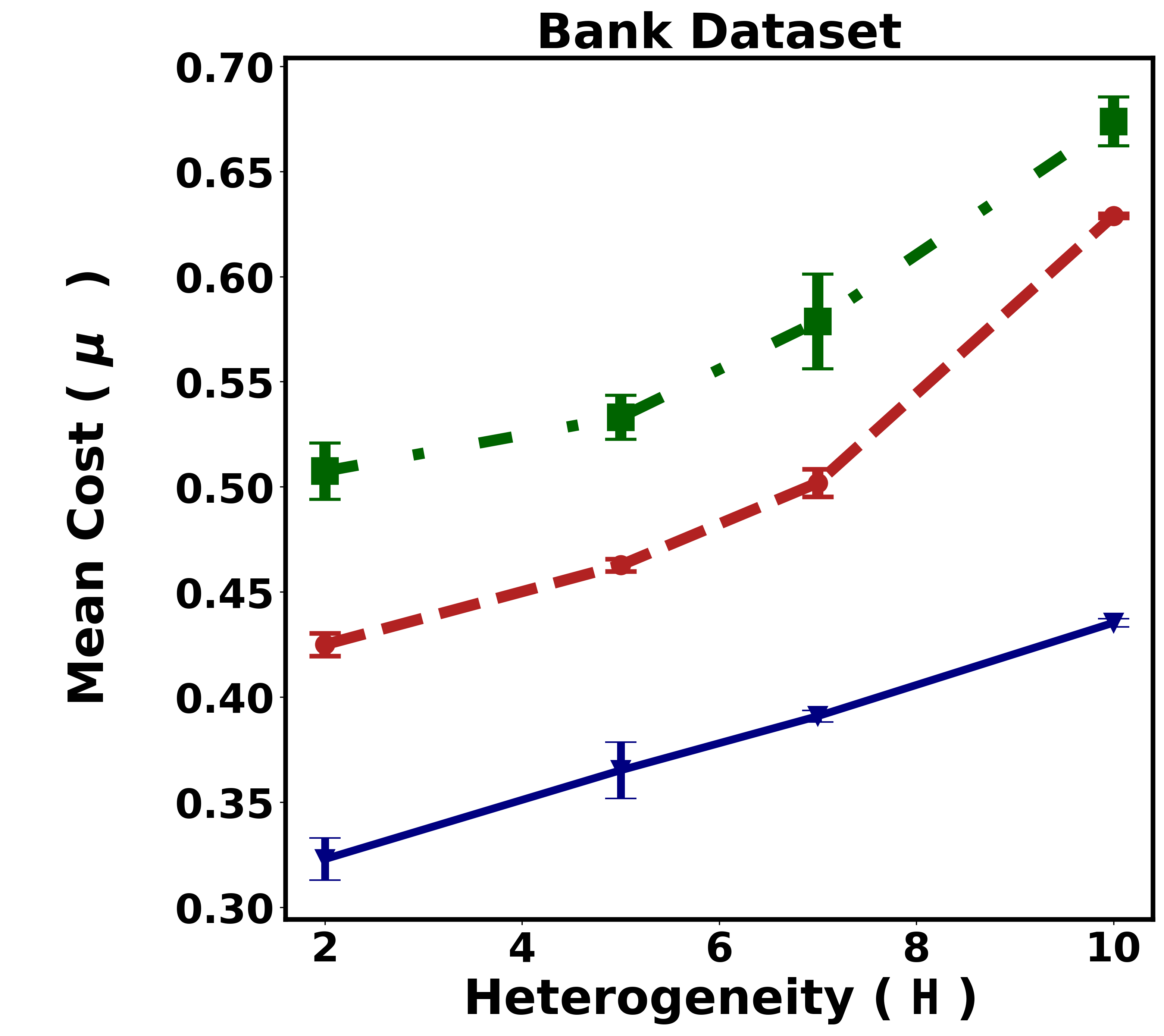}&\includegraphics[width=0.25\textwidth]{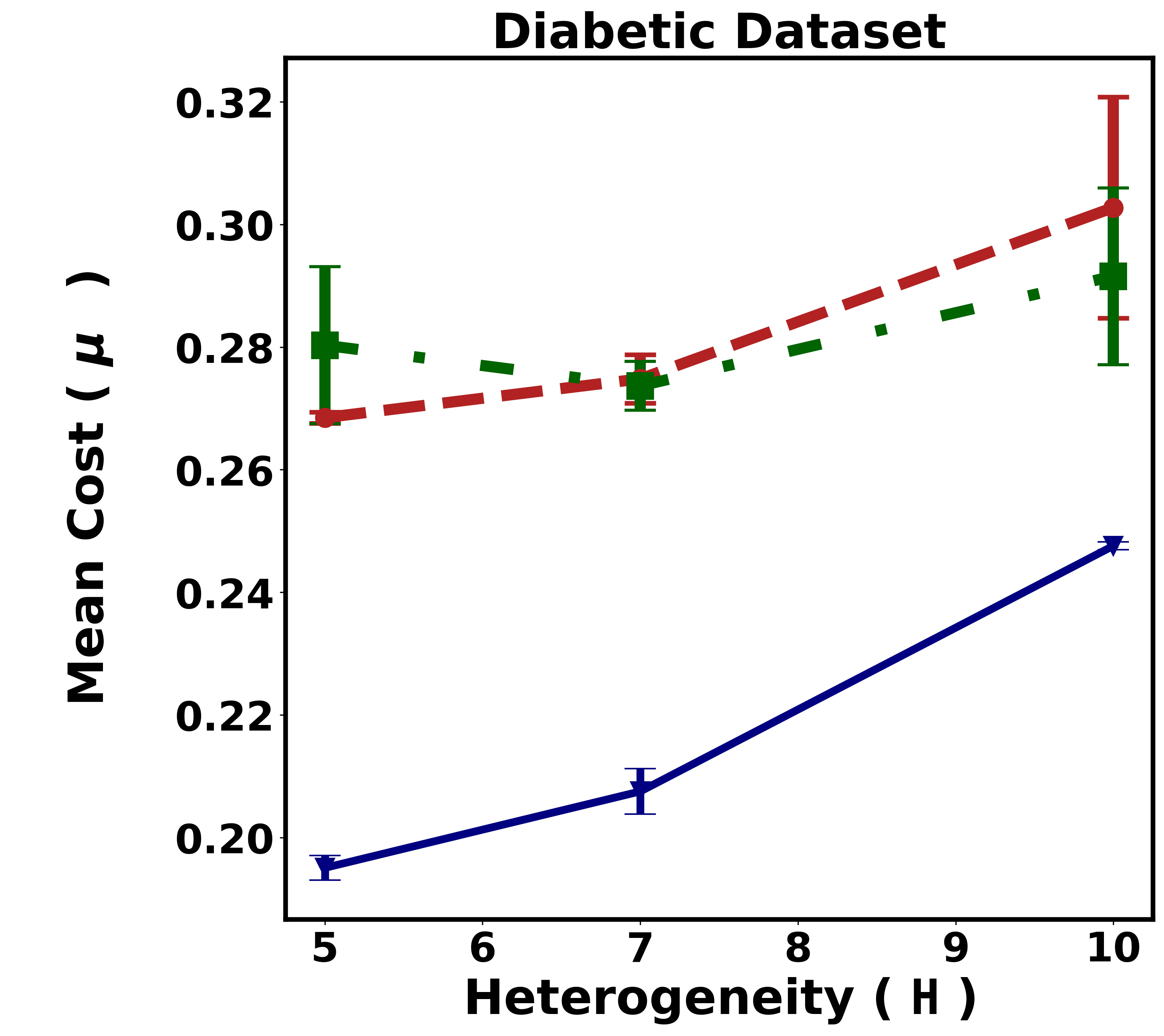}\\
    \includegraphics[width=0.25\textwidth]{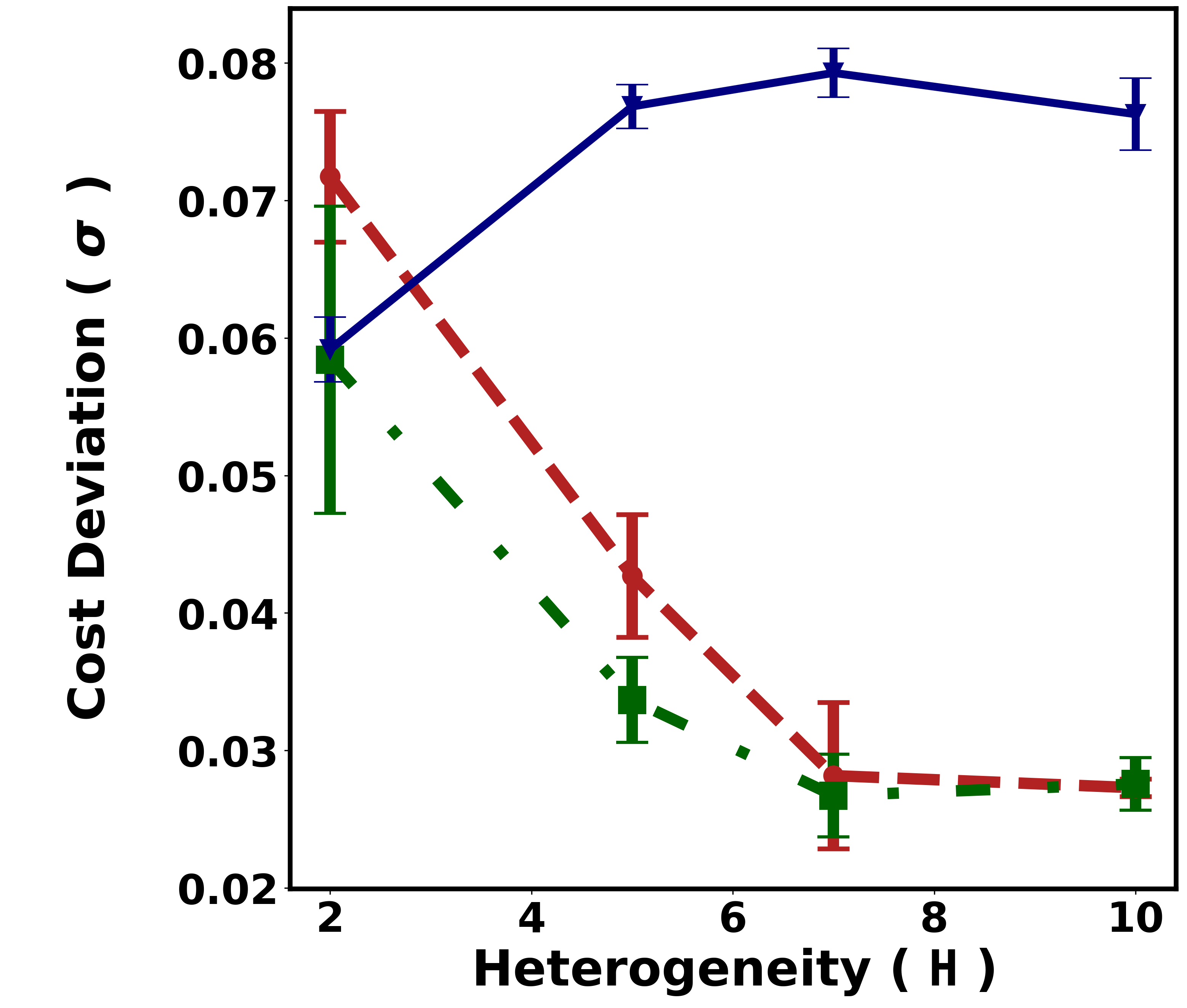}&\includegraphics[width=0.25\textwidth]{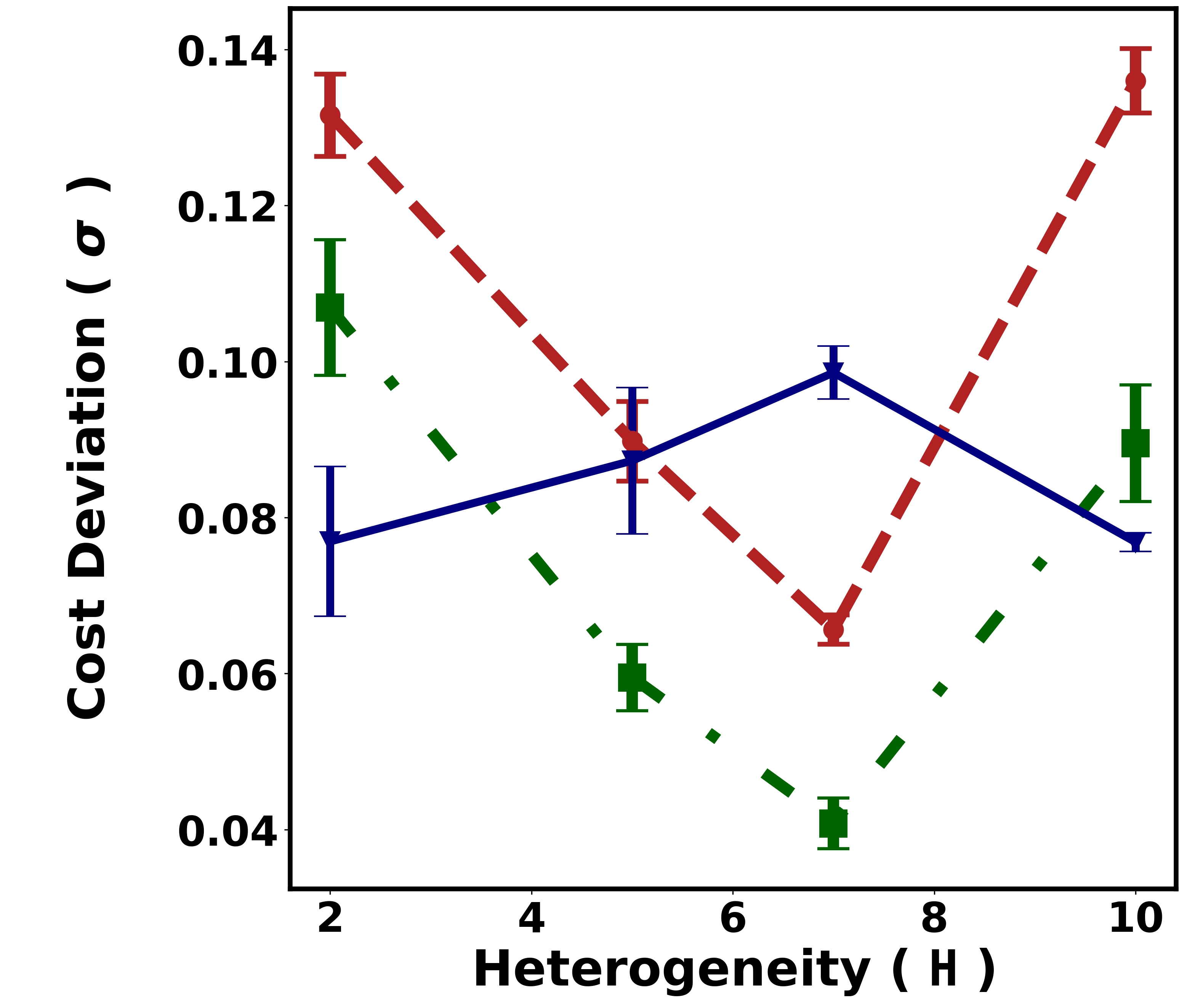}&\includegraphics[width=0.25\textwidth]{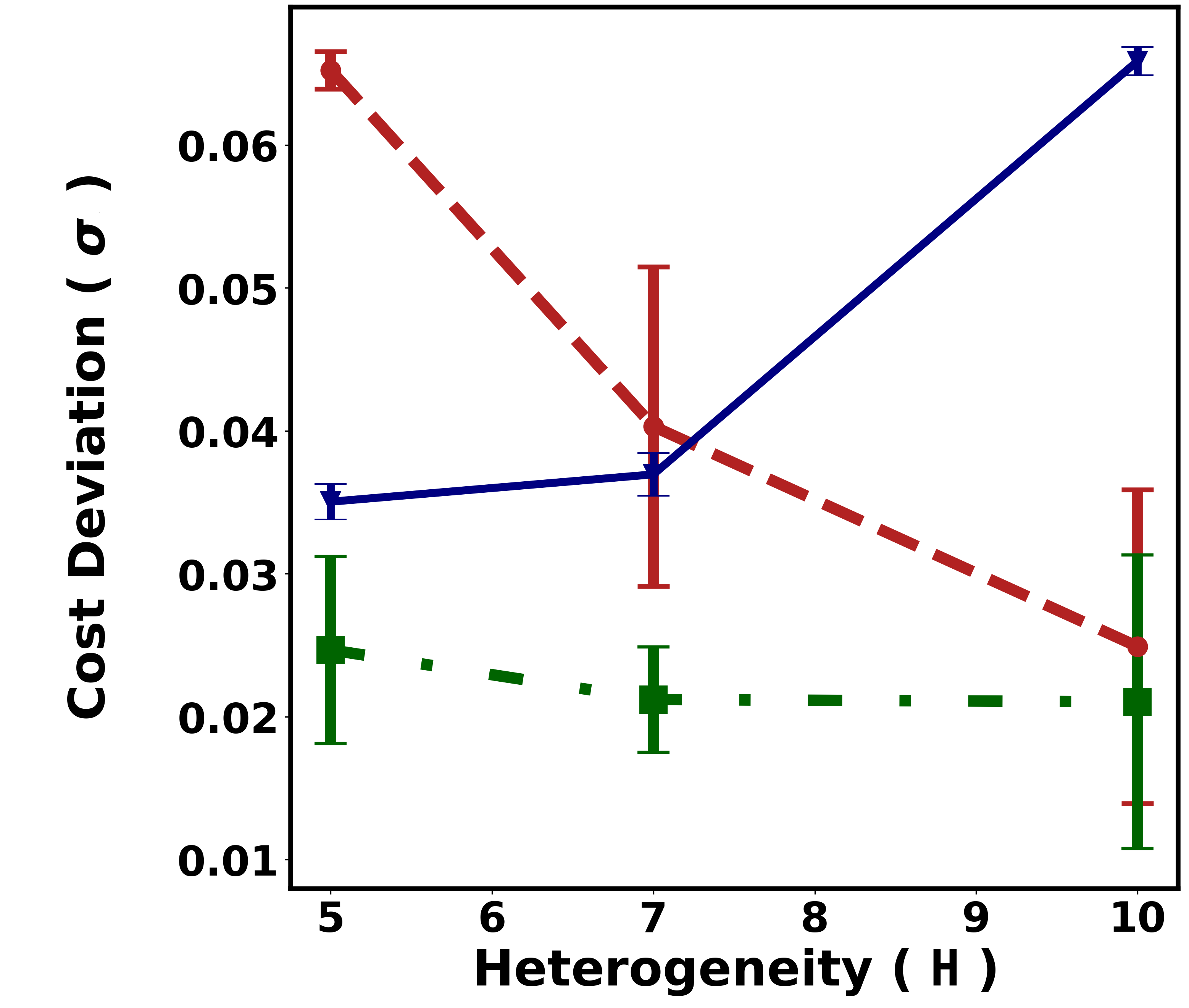}\\
    \includegraphics[width=0.25\textwidth]{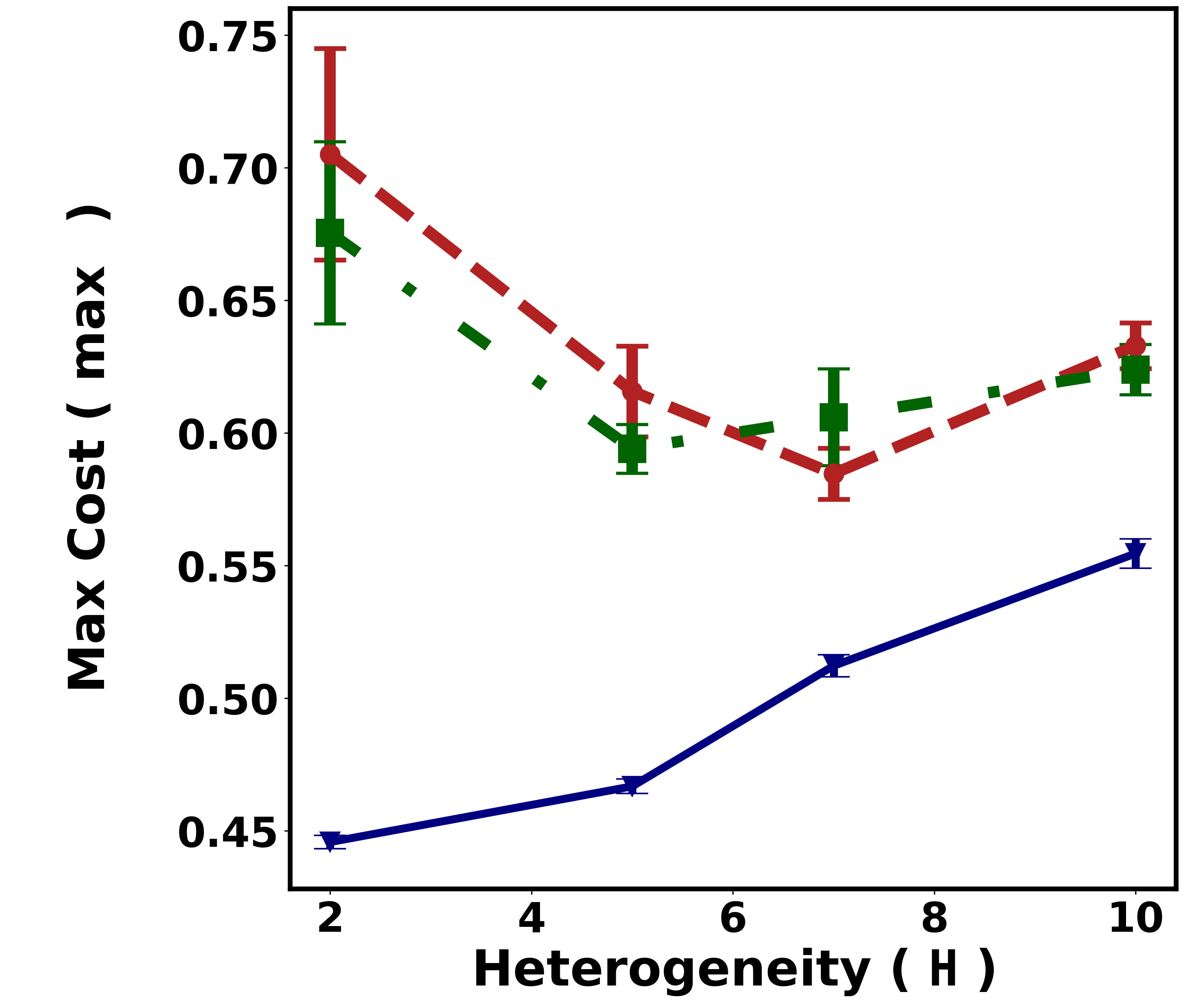}&\includegraphics[width=0.25\textwidth]{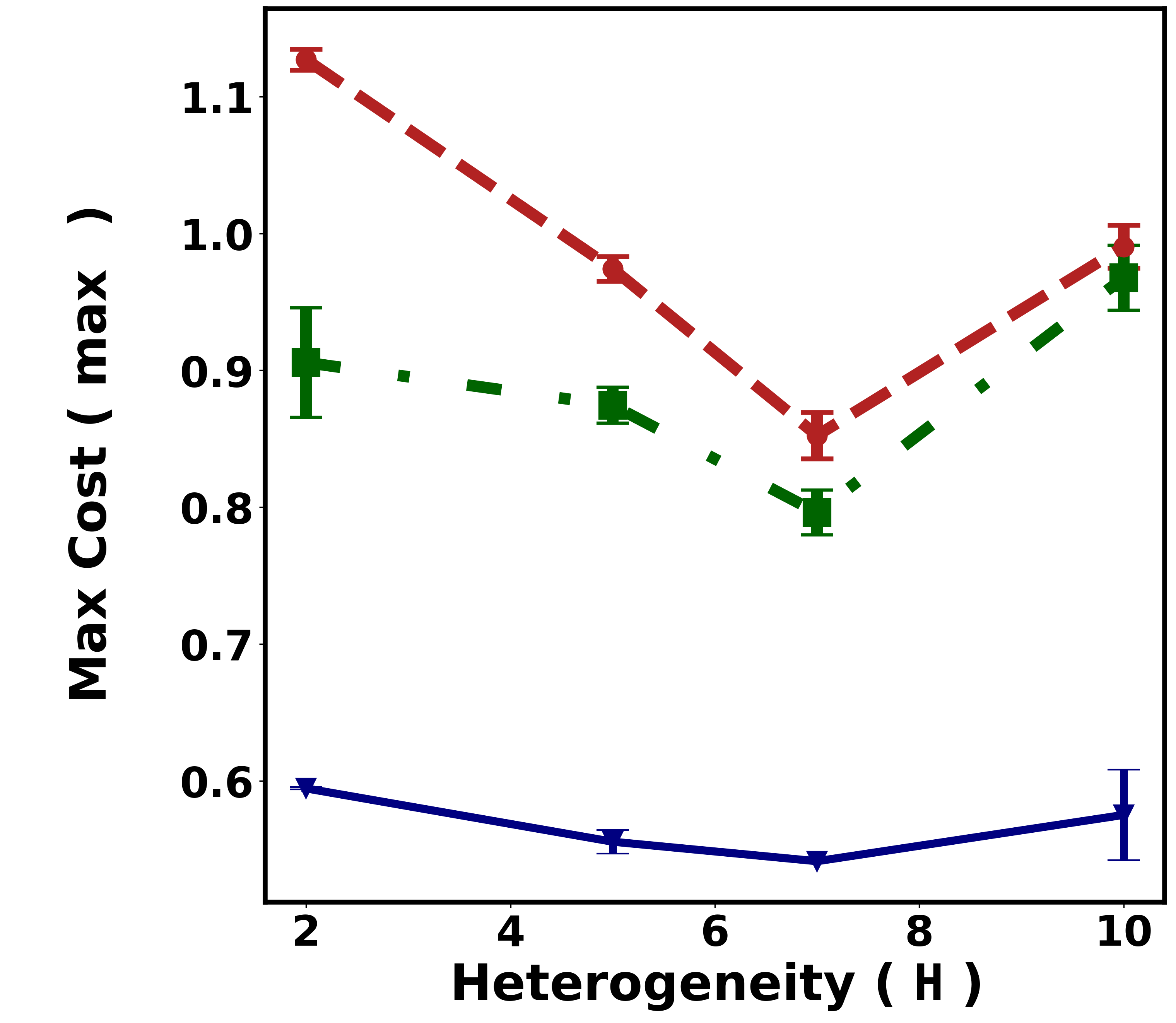}&\includegraphics[width=0.25\textwidth]{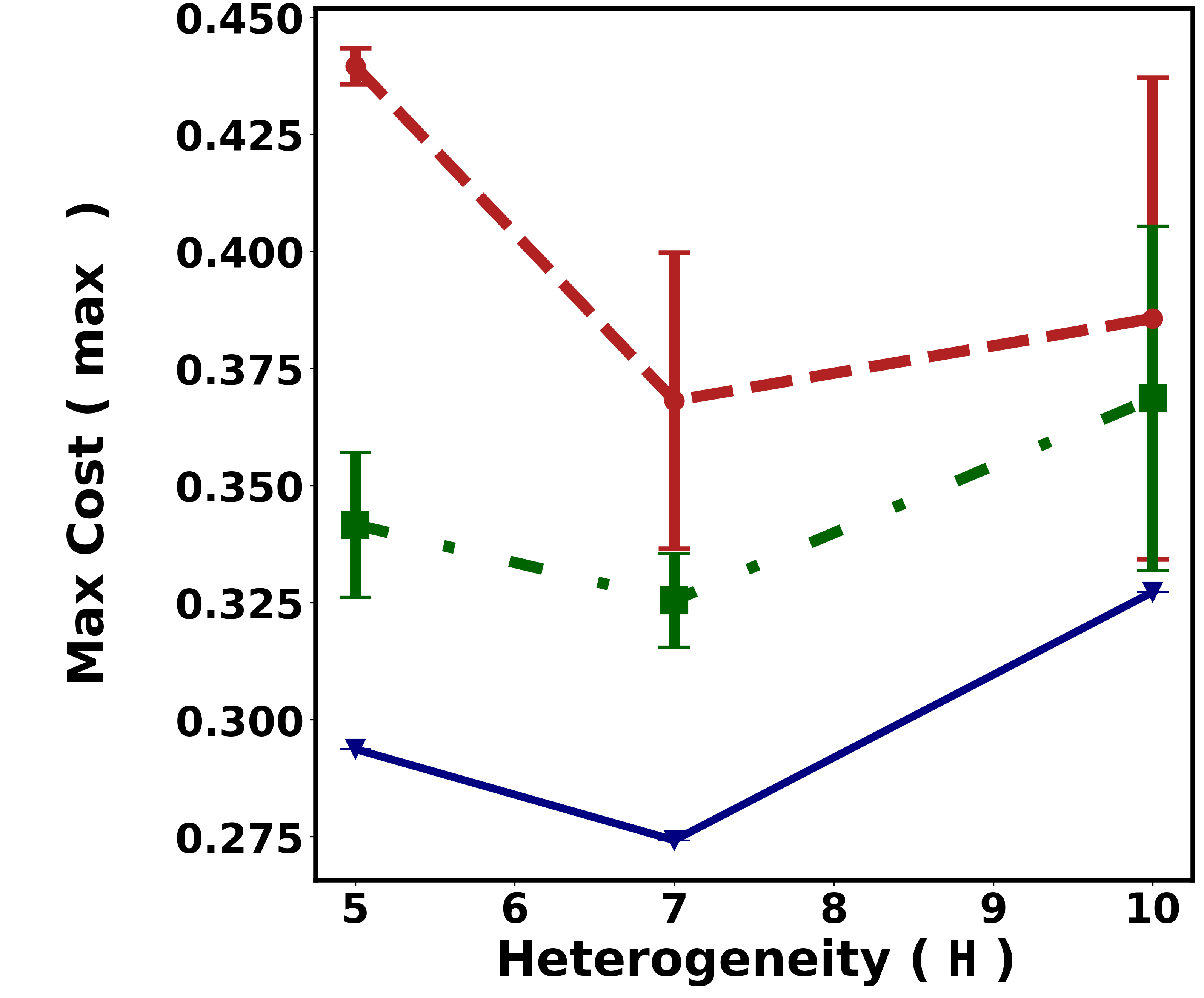}\\
      \multicolumn{3}{l}{\centering\includegraphics[width=0.55\textwidth]{images_final/label_long.png}}
    \end{tabular}
      \caption{The plot shows the variation in evaluation metrics against proposed \ouralgo\ and \sota\ on $k$-means objective for varying heterogeneity levels on a \texttt{Unequal} data split across $100$ clients. Each column represents a specific dataset as specified at the top, and each row represents one metric under evaluation. (Best viewed in color).}

        \label{fig:Unequal100clients}
    \end{figure*}

    \begin{figure*}[ht!]
    \centering
    \begin{tabular}{@{}c@{}c@{}c@{}c@{}}
    \includegraphics[width=0.25\textwidth]{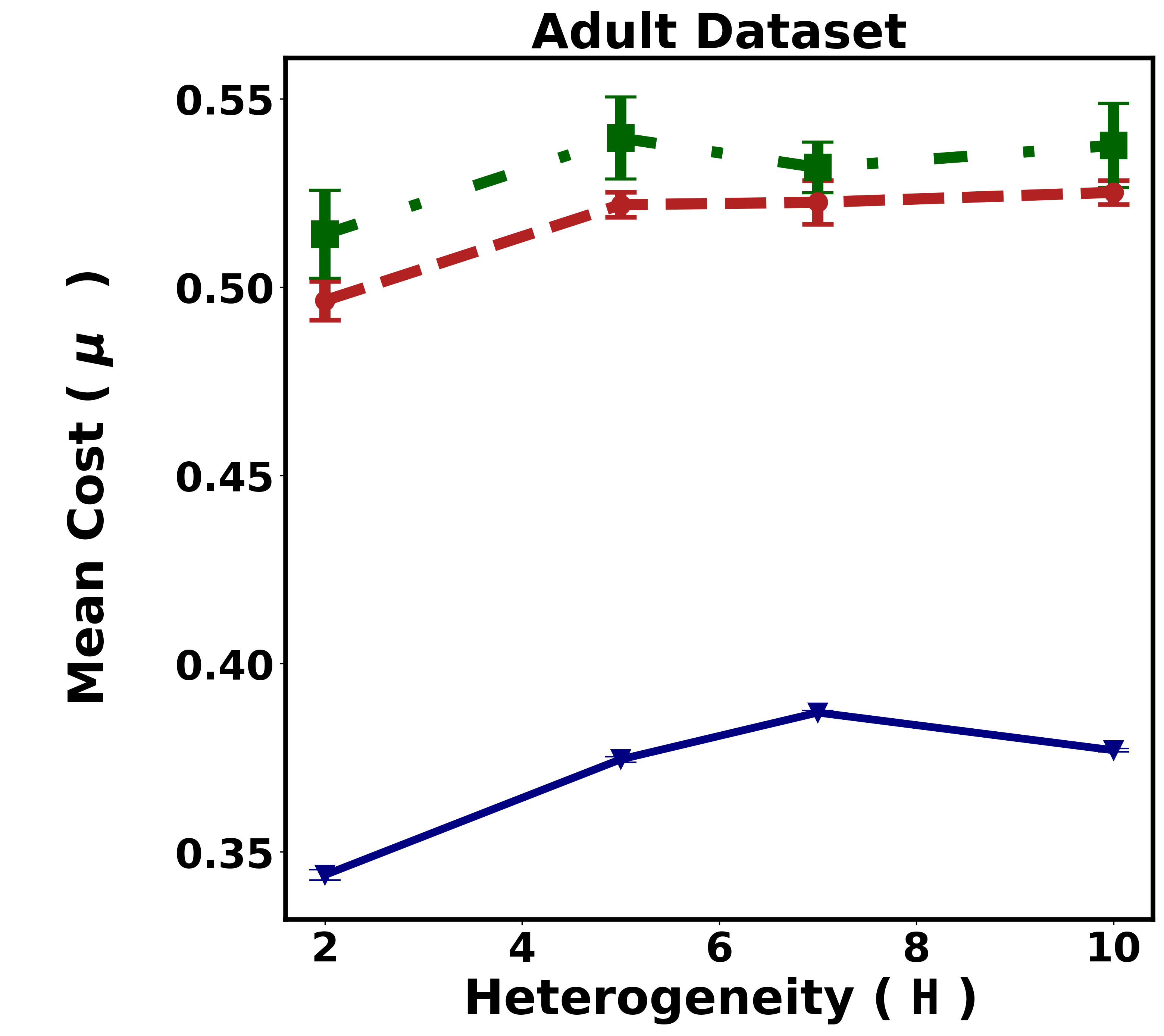}&\includegraphics[width=0.25\textwidth]{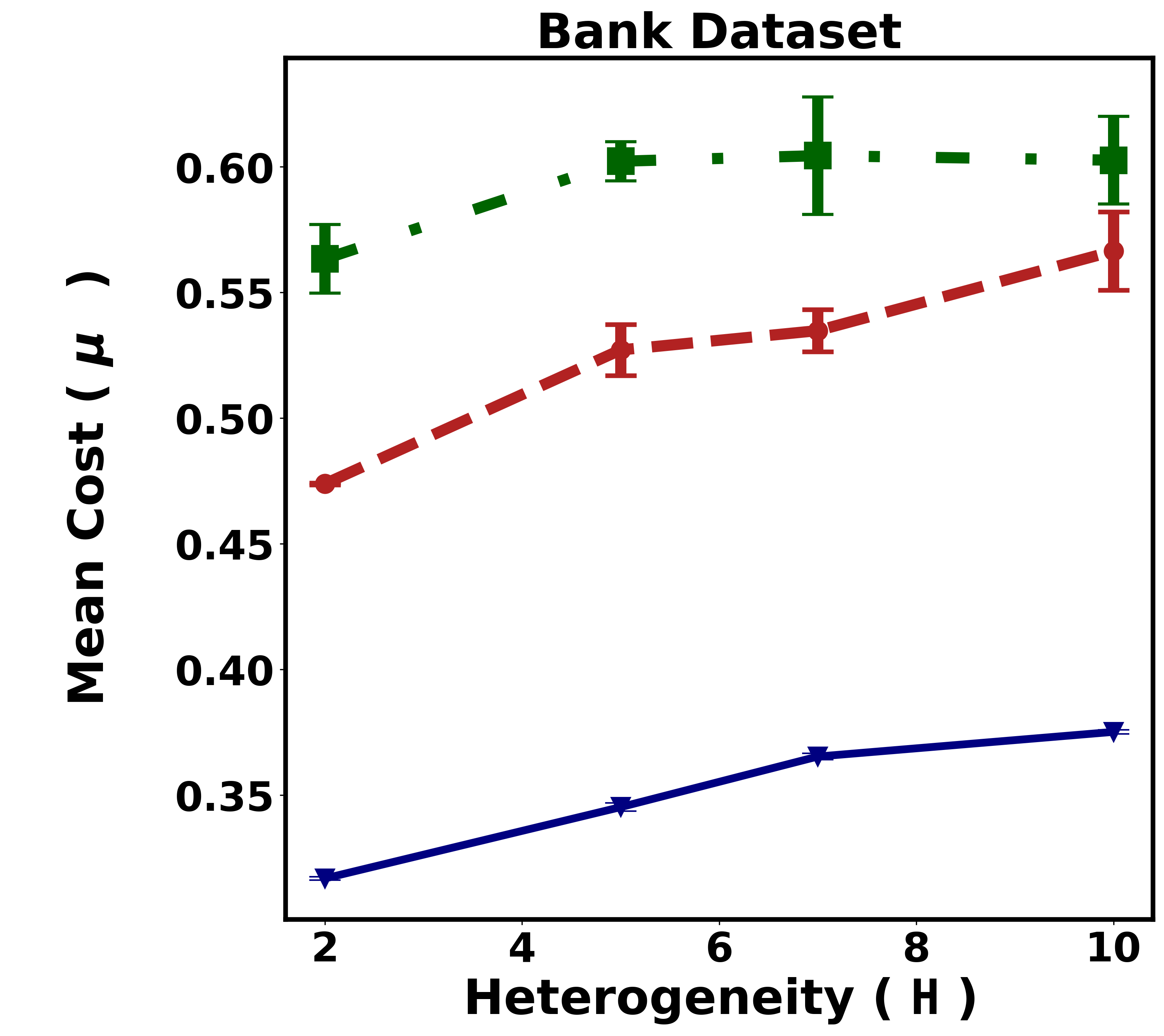}&\includegraphics[width=0.25\textwidth]{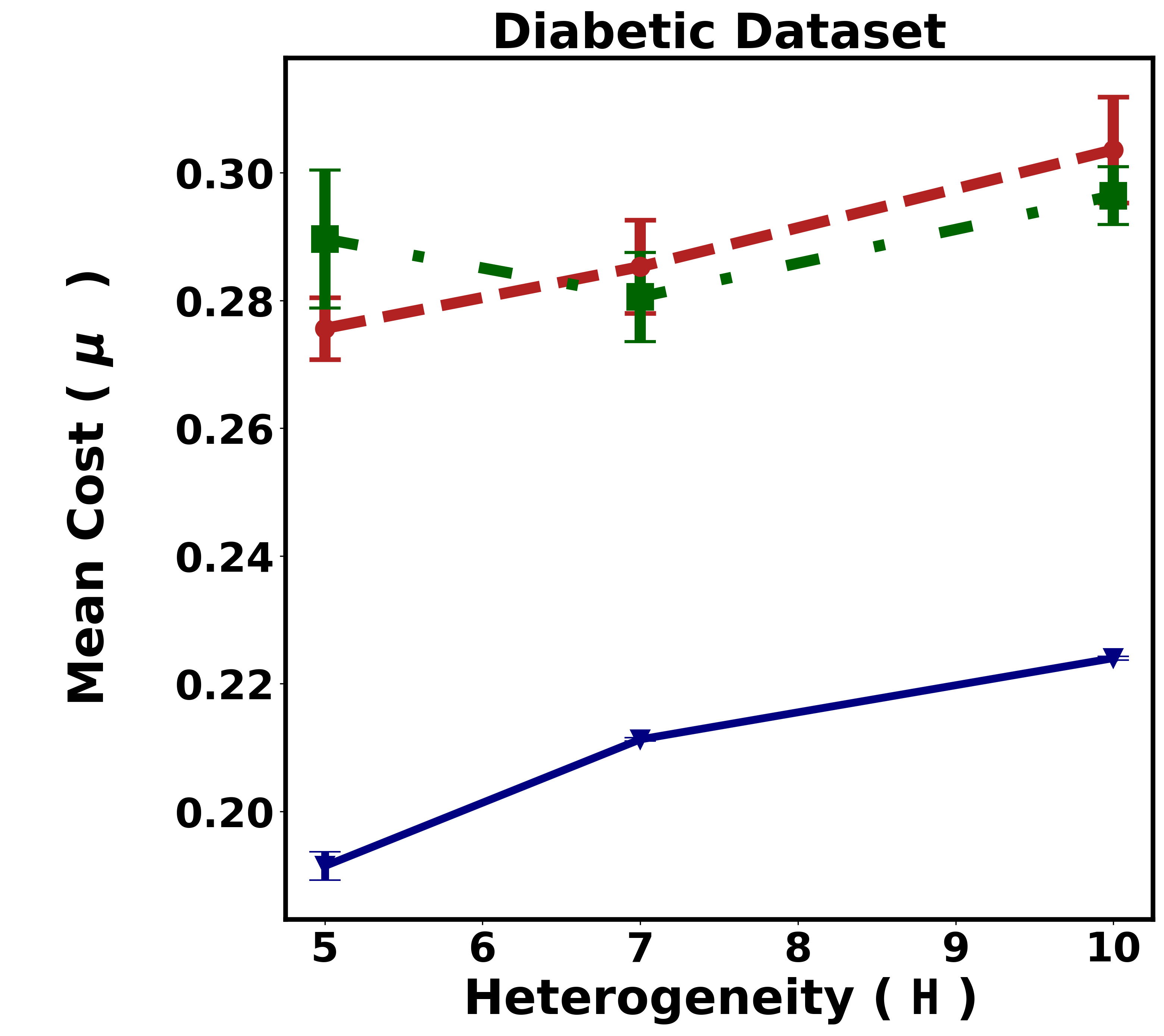}\\
    \includegraphics[width=0.25\textwidth]{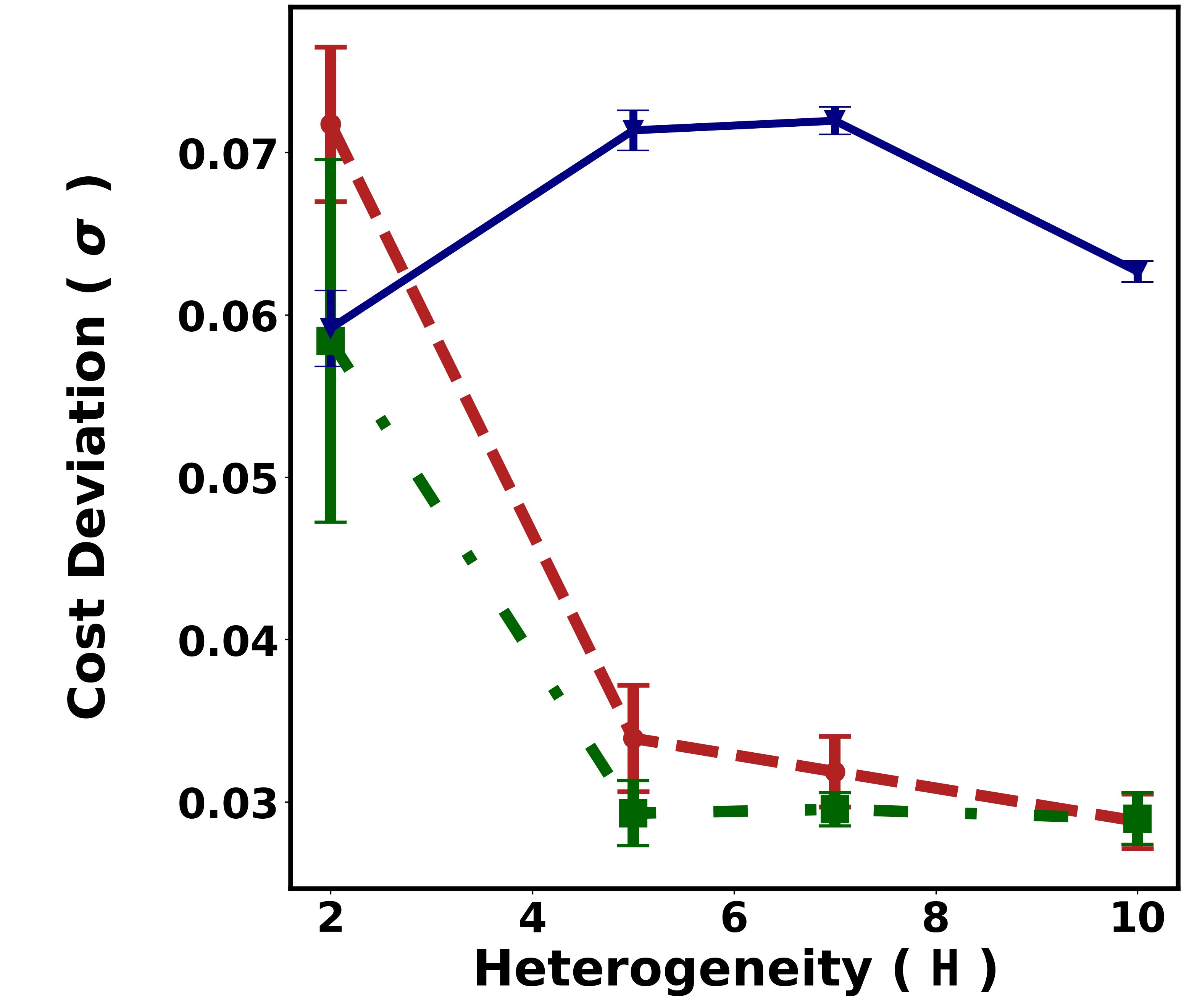}&\includegraphics[width=0.25\textwidth]{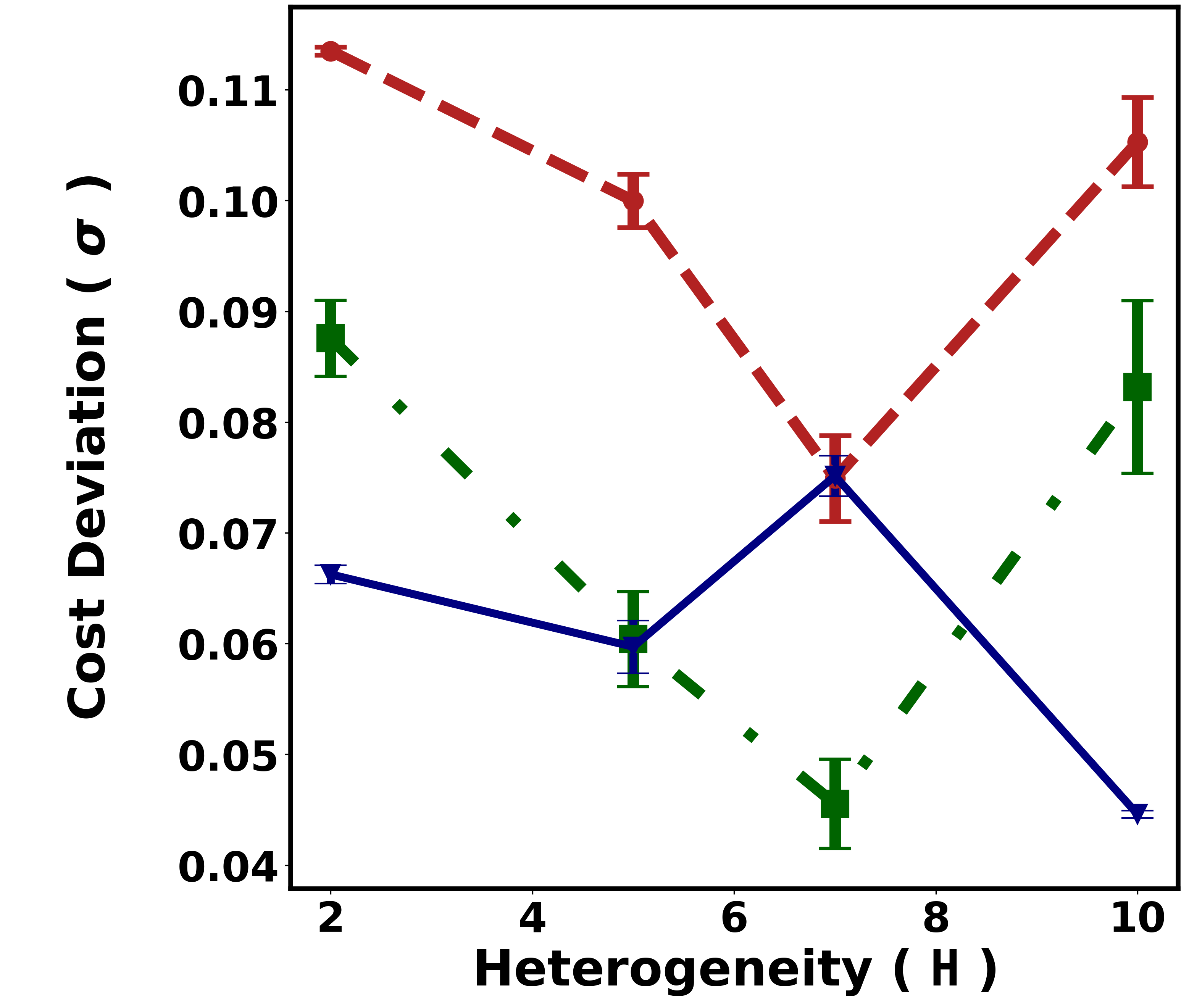}&\includegraphics[width=0.25\textwidth]{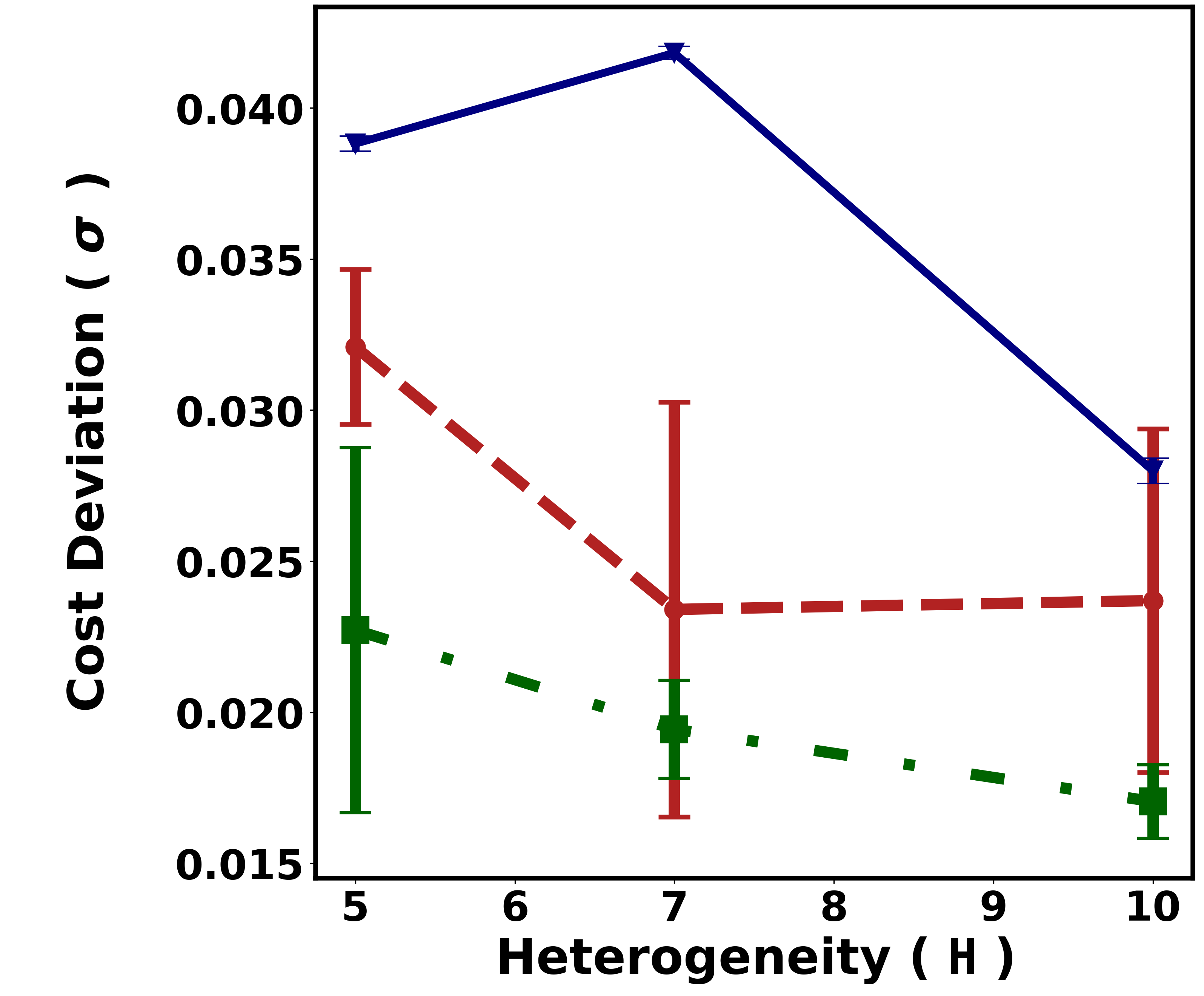}\\
    \includegraphics[width=0.25\textwidth]{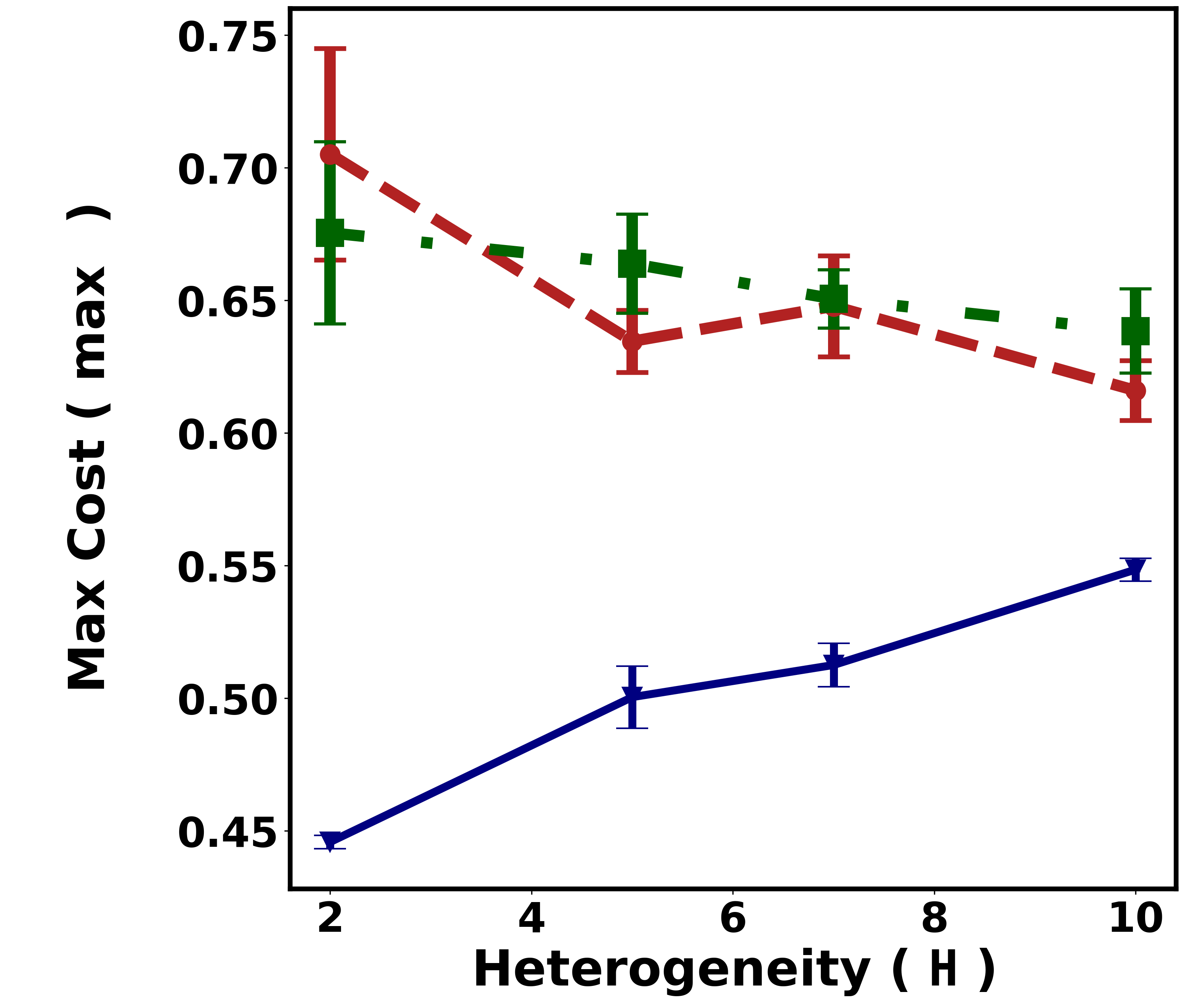}&\includegraphics[width=0.25\textwidth]{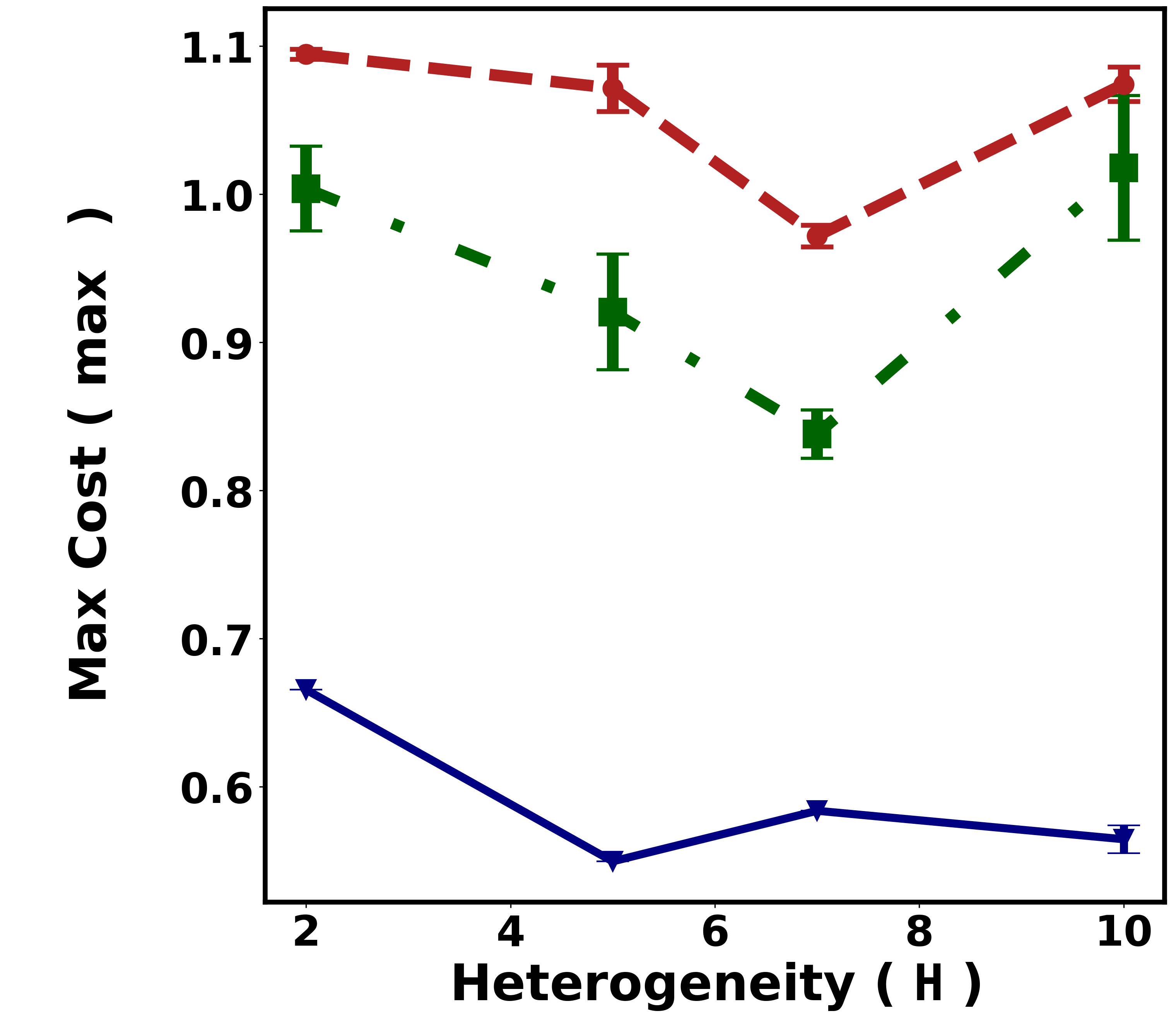}&\includegraphics[width=0.25\textwidth]{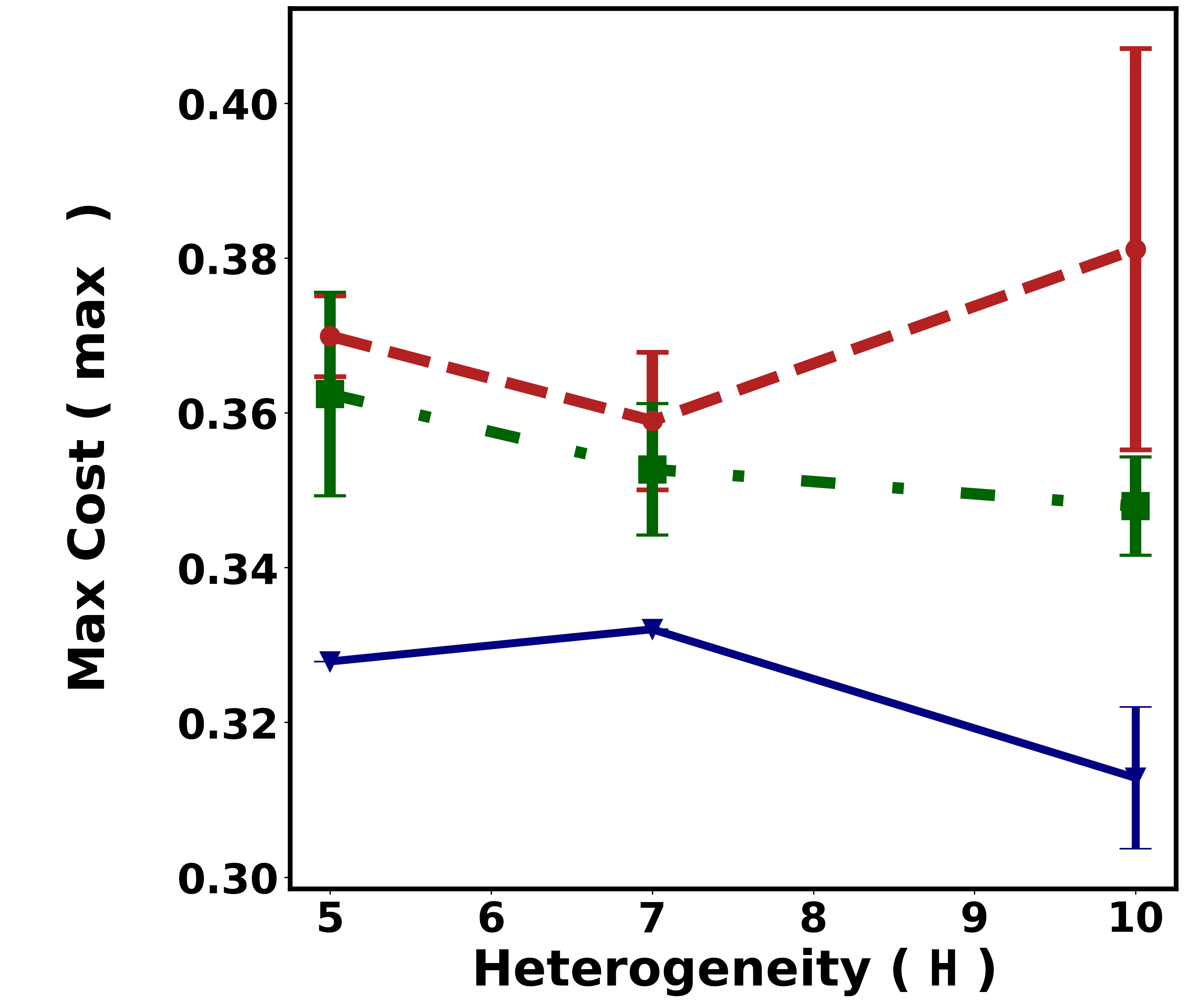}\\
  \multicolumn{3}{l}{\centering\includegraphics[width=0.55\textwidth]{images_final/label_long.png}}
    \end{tabular}
      \caption{The plot shows the variation in evaluation metrics against proposed \ouralgo\ and \sota\ on $k$-means Objective for varying heterogeneity levels on a \texttt{Unequal} data split across $500$ clients. Each column represents a specific dataset as specified at the top, and each row represents one metric under evaluation. (Best viewed in color).}

        \label{fig:Unequal1000clients}
    \end{figure*}


\subsubsection{\textbf{Observations for \texttt{Adult}}}
It can be observed from Fig. \ref{fig:Balanced100clients} and Fig. \ref{fig:Balanced1000clients} that for both $100$ and $1000$ clients we have the mean per-point cost ($\Mu$) for \ouralgo\ significantly lower than that of \sota, especially in more challenging settings of  lower heterogeneity. Furthermore, the variance of $\Mu$ is lower compared to both \kFed\ and \MFC, showing the efficacy of \ouralgo\ in achieving a lower  objective cost for all number of clients settings. Additionally, the fairness metric $\Var$ for \kFed\ and \MFC\ is higher at lower levels. In contrast, \ouralgo\ helps achieve a lower $\Mu$ for all clients  and stays within a fixed confidence region by fine-tuning using local data to bring the global centers close to local ones without deviating significantly from the global model. This shows that \ouralgo\ is not sensitive to changes in heterogeneity levels.  Approaches such as \kFed\ and \MFC\ can achieve better $\Mu$ and $\Var$ at higher heterogeneity, as in such scenarios, they can capture good estimates of global centers mainly due to sufficient data availability from all distributions across clients. Notably, the maximum per-point cost ($\Max$) remains consistently high for \sota\ methods compared to \ouralgo, showcasing the presence of clients that might be willing to not contribute and leave the federated system. Though there is an increasing trend for \ouralgo, it remains at a significant gap from \sota. Therefore, in the \texttt{Adult} dataset,  \ouralgo\ performs considerably well.

  \subsubsection{\textbf{Observations for \texttt{Bank}}}
  The observations are quite aligned with the \texttt{Adult} dataset.  $\Mu$, $\Max$ is considerably lower than \sota\ even on varying heterogeneity ($\mathtt{H}$) and number of clients. Also, the fairness metric, $\Var$ for \ouralgo\, remains below \sota\ for most heterogeneity ($\mathtt{H}$) levels.

  \subsubsection{\textbf{Observations for \texttt{Diabetes}}}
  The additional comments to observe is that though in \texttt{Diabetes} dataset \kFed\ and \MFC\ have lower $\Var$ but suffer high variance compared to \ouralgo. This is primarily due to local optima in the dataset (\cite{gupta2023efficient}).   Also, the $\Max$ for \ouralgo\ is either lower or comparable to \sota\ methods, showcasing the robustness of our algorithm (\ouralgo) to local optima's.

   \subsubsection{\textbf{Observations for \texttt{FMNIST}}}
   The $\Var$ of all approaches are quite close enough, but \ouralgo\ has considerably better performance on other metrics, namely $\Mu$ and $\Max$.

   \subsubsection{\textbf{Observations for Synthetic (\texttt{Syn}) }}
   We can observe from the Fig. \ref{fig:Balanced1000clientsSynthe} that when the dataset is bearing no overlaps between different clusters, both \MFC\ and \ouralgo\ have comparable costs, but as the overlap increases in \texttt{Syn-LO} and \texttt{Syn-O}, the \MFC\ method slightly deviates and achieves higher mean per point cost ($\Mu$) and higher fairness metrics i.e., deviation ($\Var$) and maximum cost ($\Max$). On the other hand, across all different settings, \kFed\ always exhibits significantly poorer performance in terms of mean cost and fairness metrics. 
   
   \noindent \textbf{Overall Insight}: In a \texttt{Balanced} data distribution, \ouralgo\ achieves a lower per-point cost ($\Mu$, owing to personalization) and a more fair solution ($\Var$) across clients, making it a reliable choice when information about the level of heterogeneity in the network is unknown or unstable. Additionally, one can not neglect the robustness (or consistency) of performance to heterogeneity levels and the number of participating clients.


\subsection{Analysis on \texttt{Unequal} Data Distribution among Clients on $k$-means Objective}
This subsection now delves into the results of the unequal data distribution setting in the $k$-means objective ($\ell = 2$). The results are illustrated in Fig. \ref{fig:Unequal100clients} to \ref{fig:Unequal1000clients}. We first provide a brief overview of the observations per dataset and then summarize the overall results in this setting.


    \begin{figure*}[ht!]
    \centering
    \begin{tabular}{@{}c@{}c@{}c@{}c@{}}
    \includegraphics[width=0.25\textwidth]{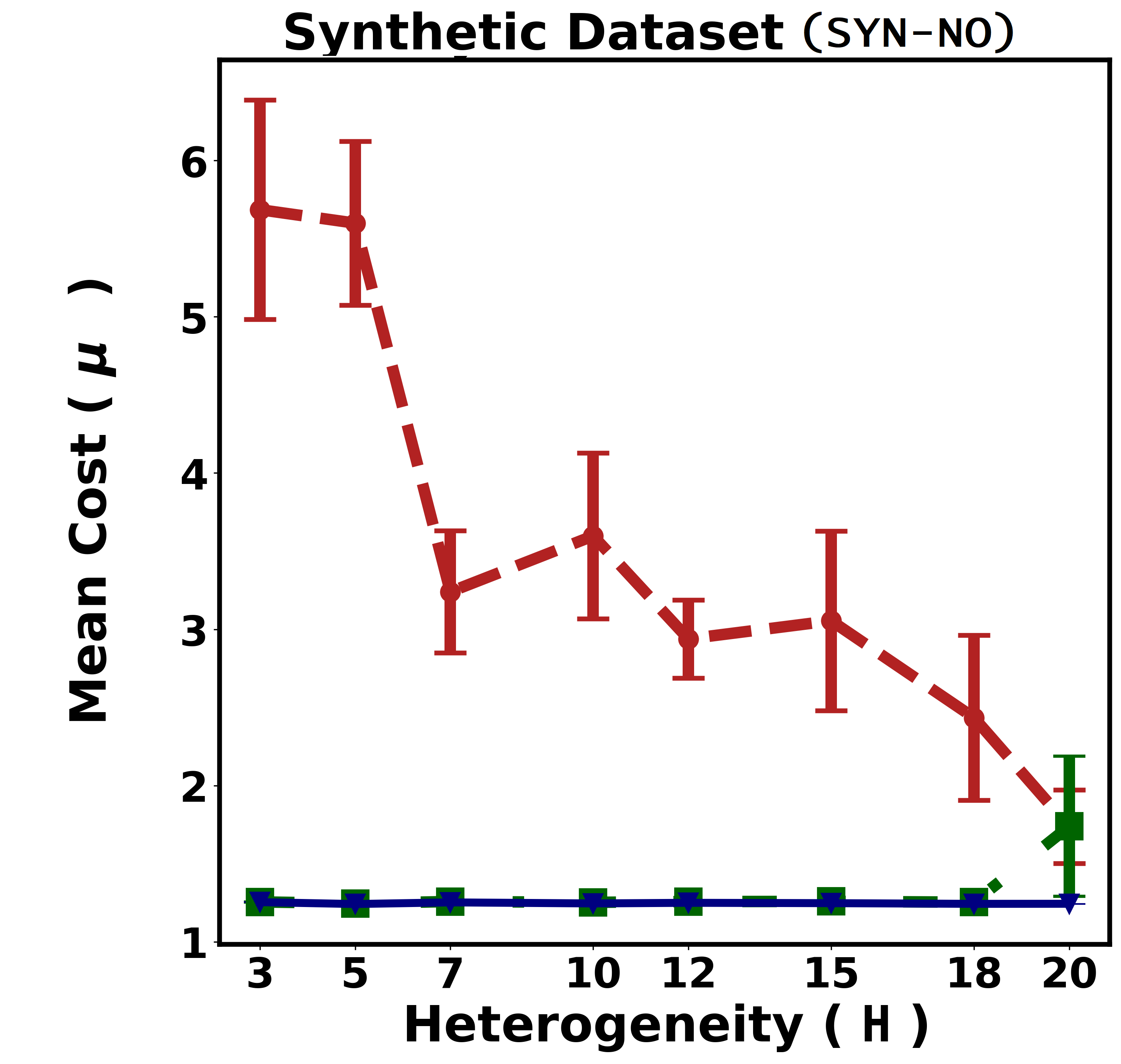}&\includegraphics[width=0.25\textwidth]{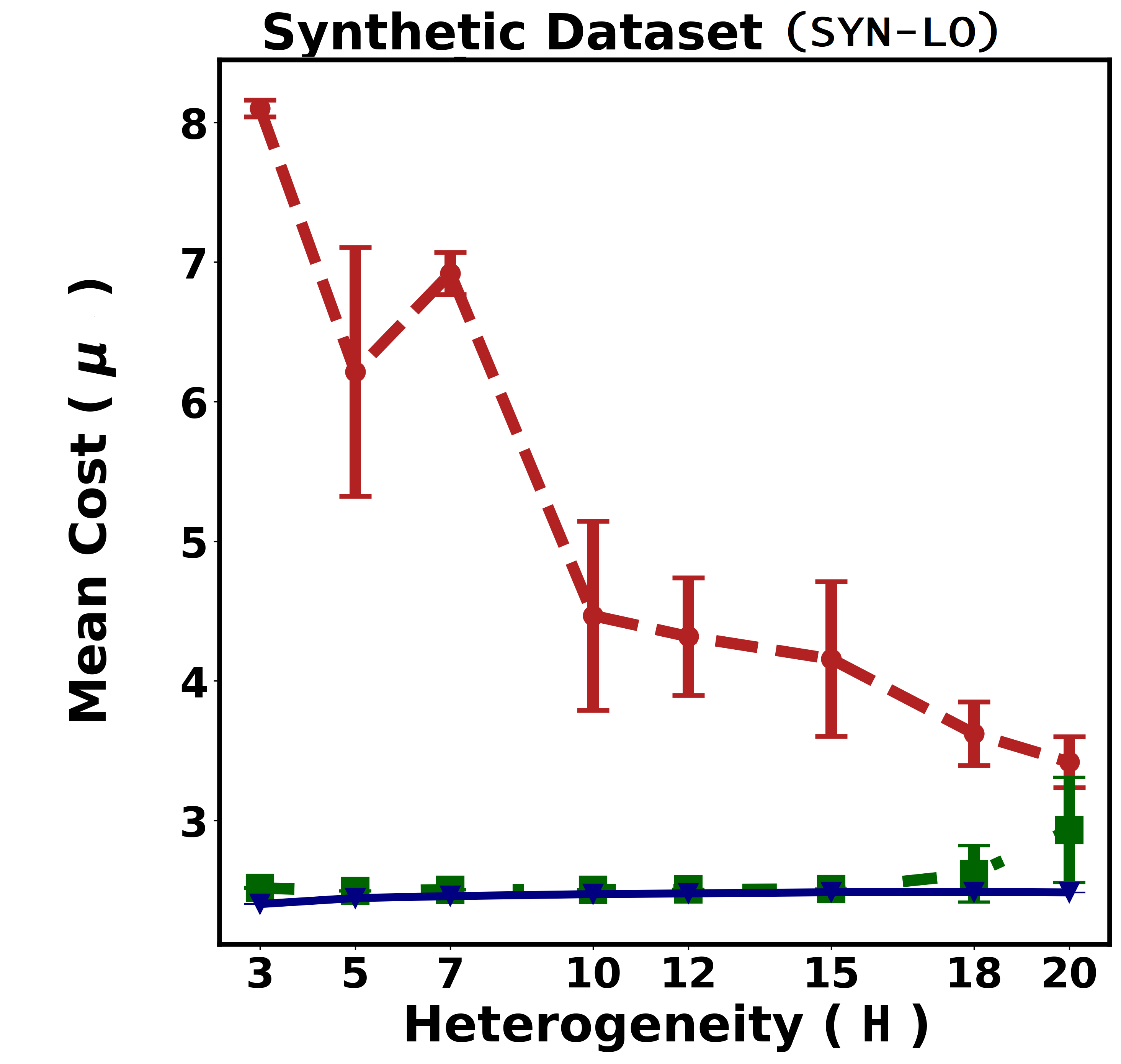}&\includegraphics[width=0.25\textwidth]{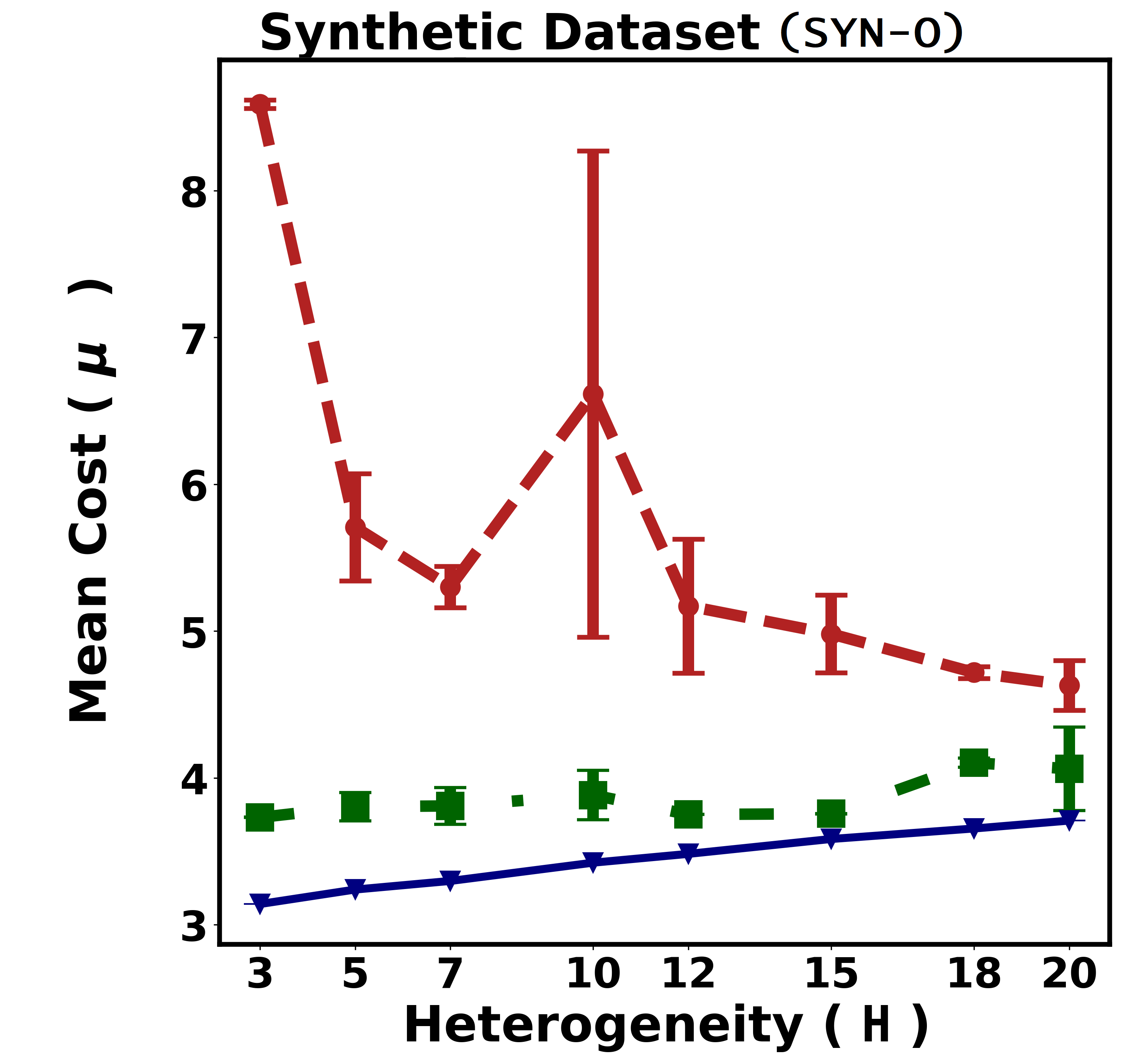}\\
    \includegraphics[width=0.25\textwidth]{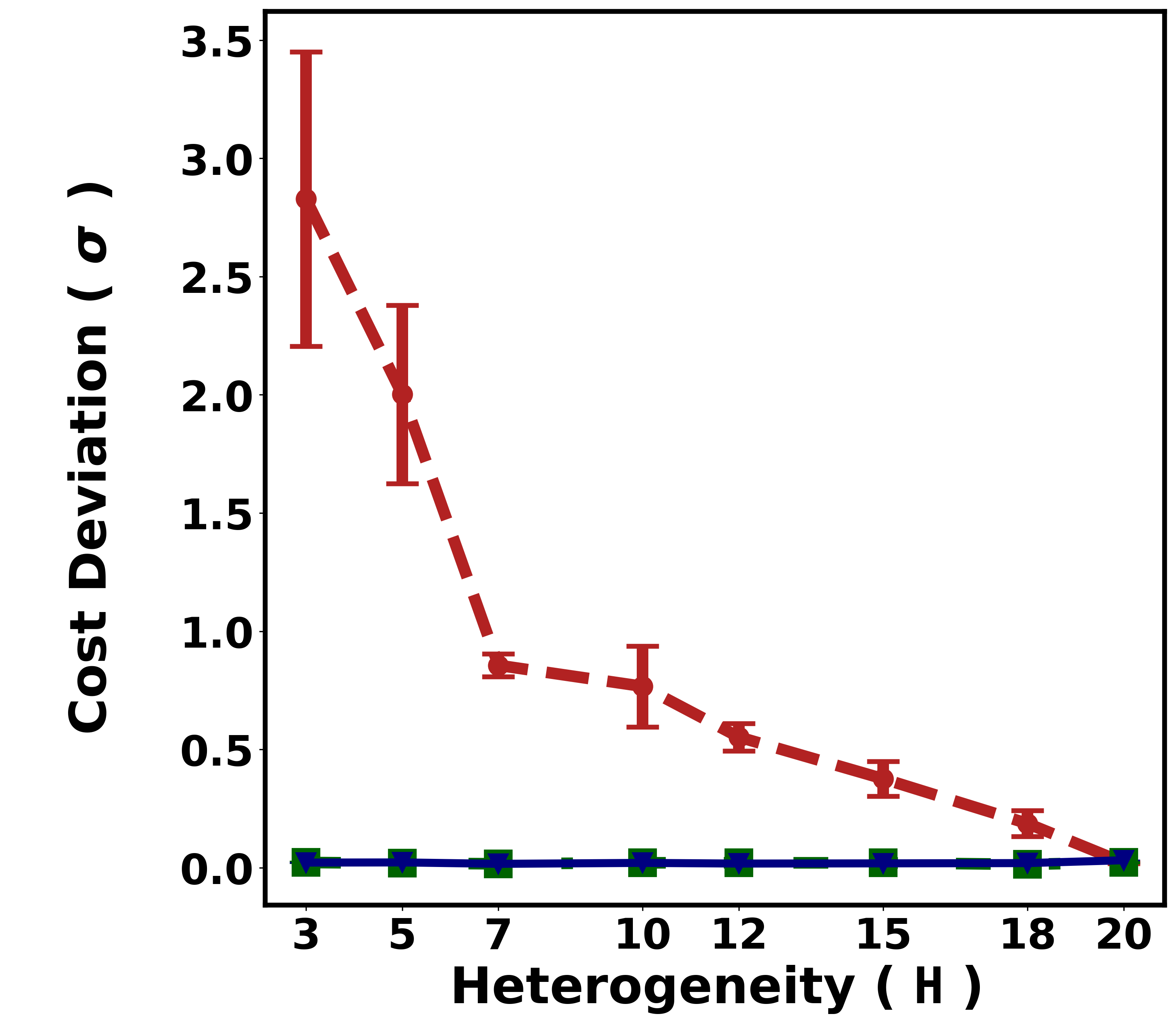}&\includegraphics[width=0.25\textwidth]{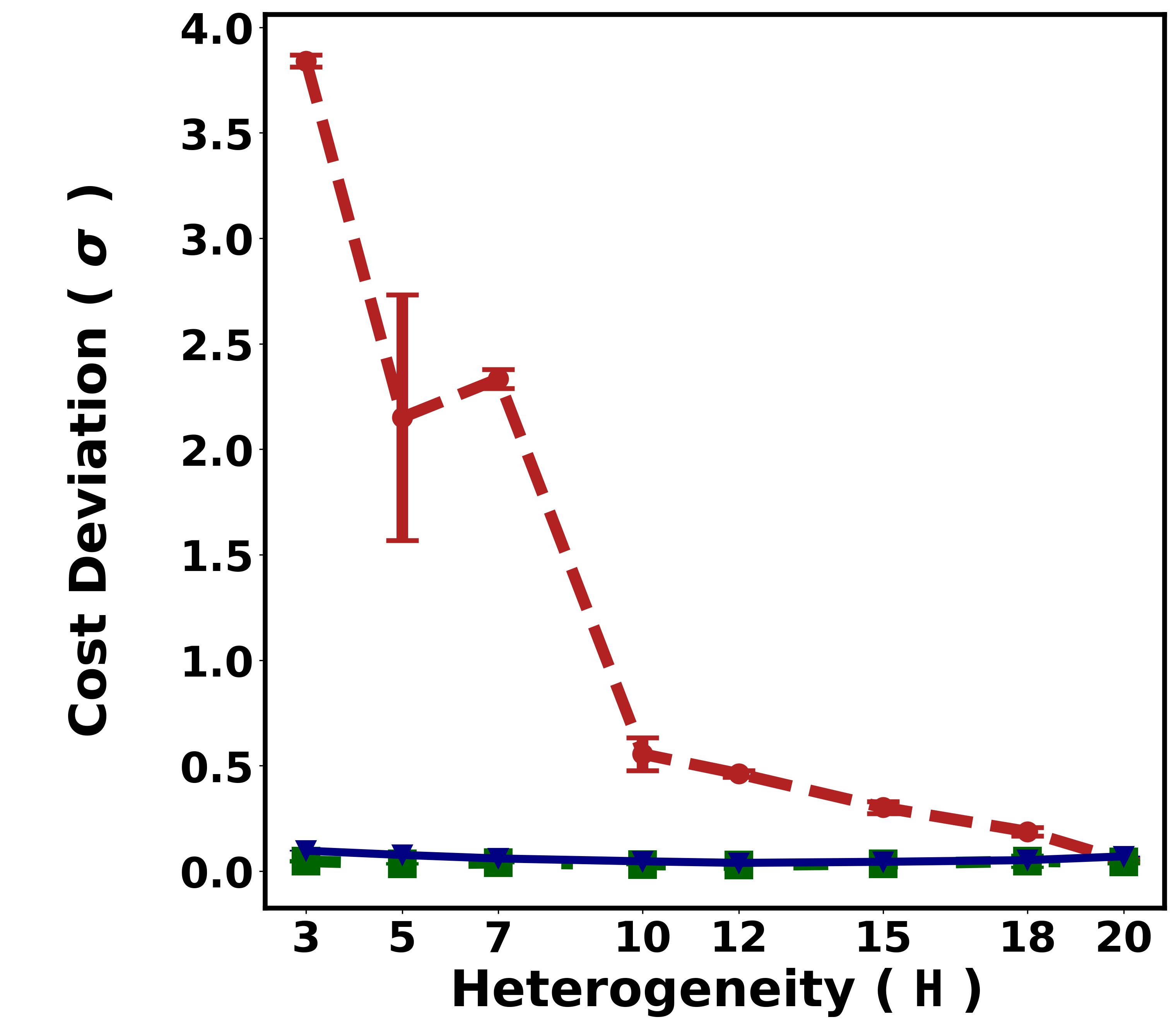}&\includegraphics[width=0.25\textwidth]{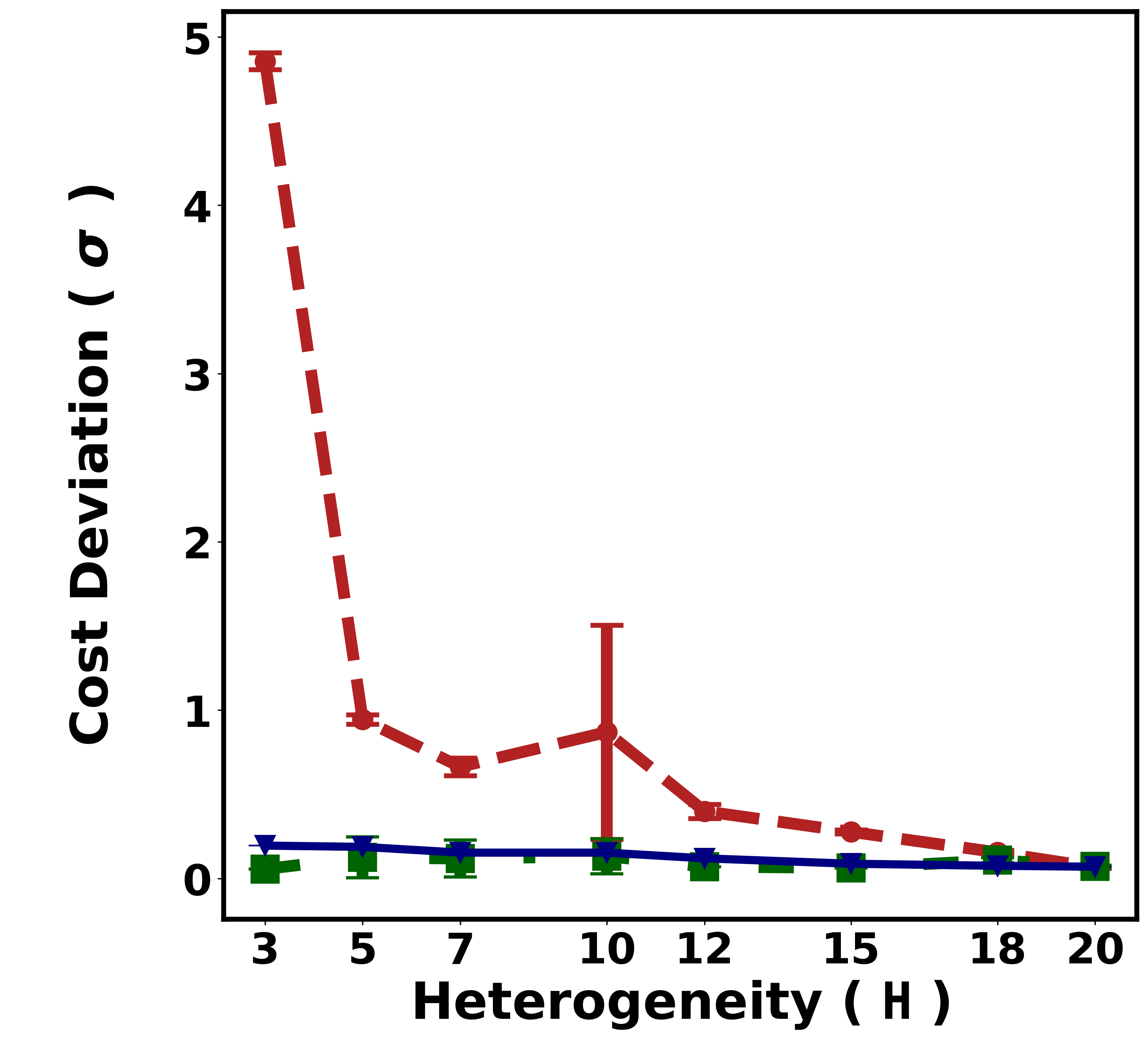}\\
    \includegraphics[width=0.25\textwidth]{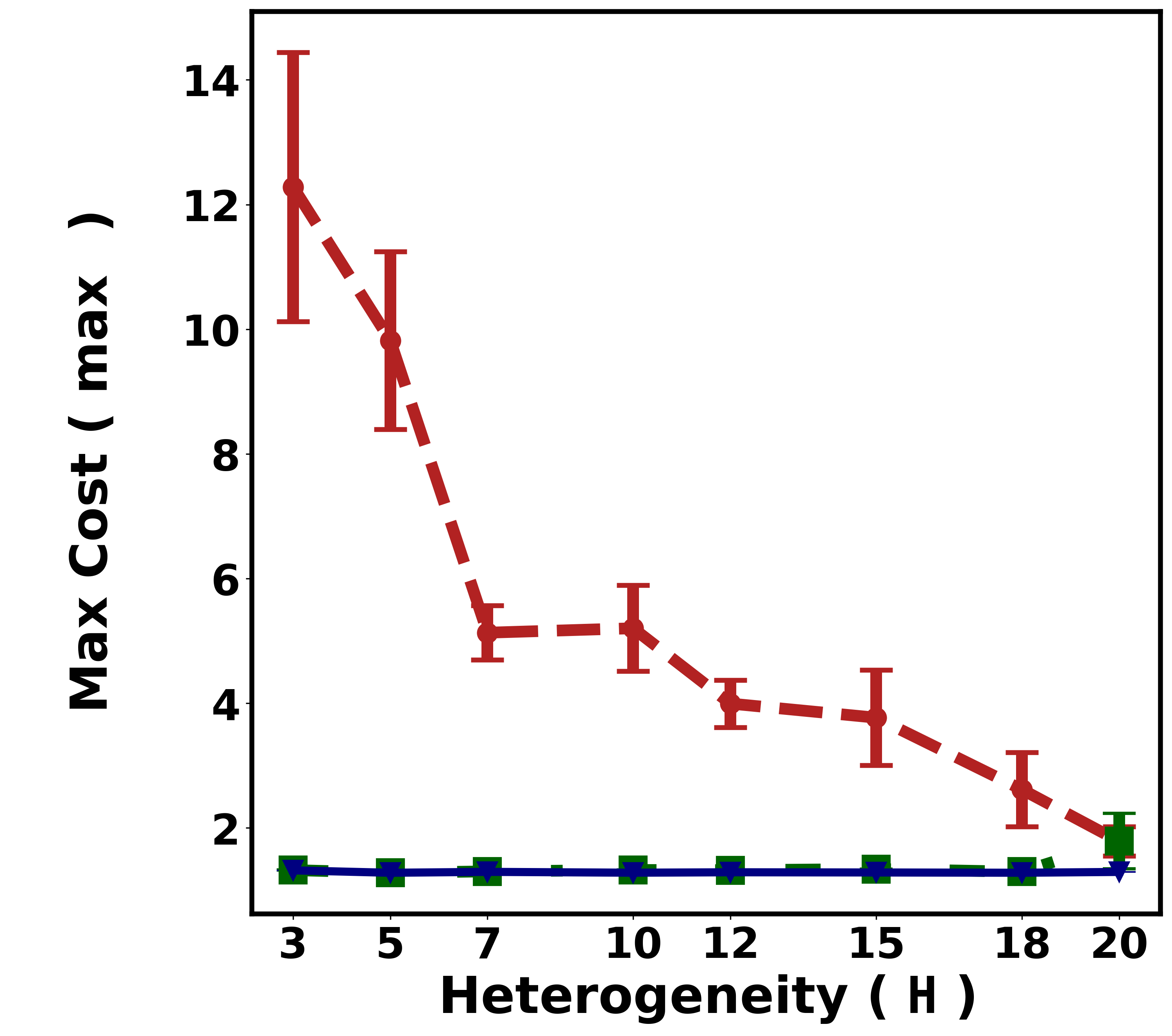}&\includegraphics[width=0.25\textwidth]{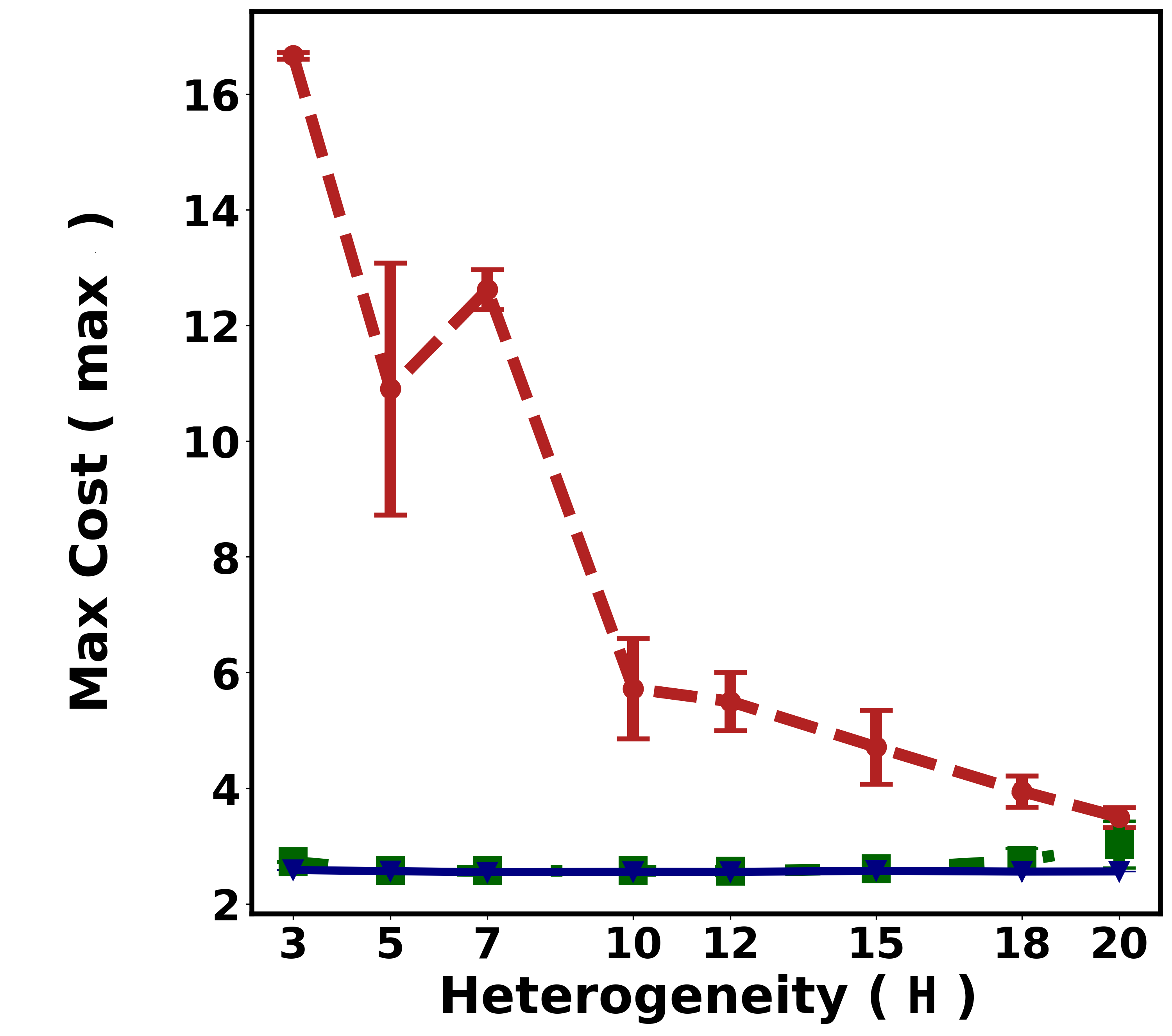}&\includegraphics[width=0.25\textwidth]{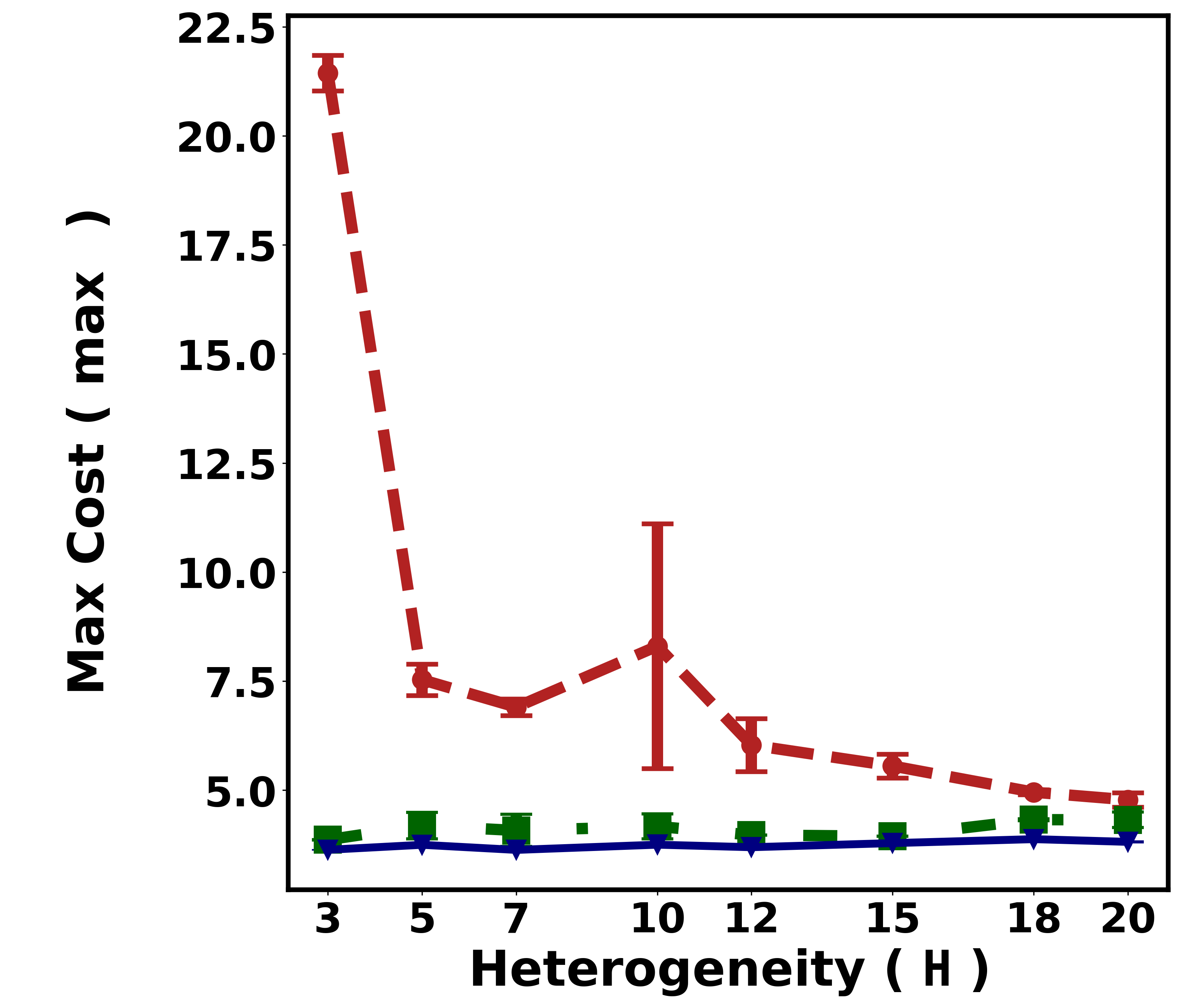}\\
    \multicolumn{3}{l}{\includegraphics[width=0.55\textwidth]{images_final/label_long.png}}
    \end{tabular}
      \caption{The plot shows the variation in evaluation metrics against proposed \ouralgo\ and \sota\ on $k$-means Objective for varying heterogeneity levels on a \texttt{Unequal} data split across $50$ clients. Each column represents a specific Synthetic dataset (\texttt{Syn}) in sequence:  \texttt{Syn-NO}, \texttt{Syn-LO}, \texttt{Syn-O} respectively, and each row represents one metric under evaluation. (Best viewed in color).}

        \label{fig:Unequal1000clients}
    \end{figure*}

 \subsubsection{\textbf{Observations for \texttt{Adult}}}
 The $\Mu$ is considerably lower for \ouralgo\ compared to \sota\ methods. However, the performance of \ouralgo\ becomes slightly questionable when it comes to the $\Var$ metric since $\Var$ is higher for \kFed\ and \MFC\ at smaller $\mathtt{H}$ values but drops below \ouralgo\ as $\mathtt{H}$ increases. However, one should not overlook the observation that there is high variability in $\Var$ for \sota\ methods, whereas the proposed \ouralgo\ maintains a reasonable confidence interval. Further, it is least affected by the number of total clients and the heterogeneity level that are unknown in most scenarios. Thus showcasing the adaptability of \ouralgo\ seamlessly to many real-world applications without worrying about the system's heterogeneity level. Also, it should be noted that the slightly higher $\Var$ is for a significantly reduced mean cost. Thus, when $\Mu$ and $\Var$ are considered together, it demonstrates the efficacy of \ouralgo. 

\renewcommand{\arraystretch}{1.2}
\begin{table*}[h!]
\centering
\scalebox{0.8}{
\begin{tabular}{|l|l|c|c|c|}
\hline
Dataset                  & Method    & $\Mu$ ($\downarrow$)       & $\Var$ ($\downarrow$)   & $\Max$ ($\downarrow$)  \\ \hline
\multirow{3}{*}{\texttt{WISDM}}   & \kFed\   & $5.71\times10^{12}$ $\pm$ $8.02\times10^{10}$ & $1.32\times10^{13}$ $\pm$  $1.88\times10^{9}$ & $7.21\times10^{13}$ $\pm$ $1.32\times10^{10}$ \\ \cline{2-5} 
                         & \MFC     & $3.35\times10^{12}$ $\pm$ $8.32\times10^{6}$ & $2.51\times10^{12}$ $\pm$ $8.68\times10^{6}$ & $1.10\times10^{13}$ $\pm$ $8.17\times10^{7}$\\ \cline{2-5} 
                         & \textbf{\ouralgo} & {5.64$\times10^{11}$ $\pm$ $5.28\times10^{5}$} & {1.96 $\times$$10^{11}$ $\pm$ $8.05\times10^{5}$} & {1.19$\times10^{12}$ $\pm$ $8.33\times10^{6}$}  \\ \hline
\multirow{3}{*}{\texttt{FEMNIST}} & \kFed\   & {$2.806$ $\pm$ $0.0022$}     & $0.3518  \pm 0.0008 $   & $4.4780 \pm 0.0130 $     \\ \cline{2-5} 
                         & \MFC     & $3.158 \pm 0.0341 $     & {$0.3110$ $\pm$ $0.0186$ }    & $4.3710 \pm 0.0427 $     \\ \cline{2-5} 
                         & \textbf{\ouralgo} & $3.103  \pm 0.0048 $     & $0.3640  \pm 0.0035 $   & {$4.1700$ $\pm $ $0.0513$  }    \\ \hline
                         
\end{tabular}
}
\caption{The table summarizes mean and deviation of evaluation metrics against proposed \ouralgo\ and \sota\ on $k$-means for the Intrinsic datasets.}
\label{tab:intrisic}
\end{table*}

  \subsubsection{\textbf{Observations for \texttt{Bank}}}
  The trend for $\Var$ is arbitrary in Fig.\ref{fig:Unequal100clients} to \ref{fig:Unequal1000clients}; there is no perfect demarcation indicating after how much heterogeneity level the \sota\ methods will start performing considerably well on fairness metrics such as $\Var$ and $\Max$. In contrast, \ouralgo\ is least affected by heterogeneity level perturbations and the system's number of clients.

 \renewcommand{\arraystretch}{1.3}
\begin{table}[h!]
\scalebox{0.52}{  
\begin{tabular}{|ccccccc|}
\hline
\multicolumn{1}{|c|}{\multirow{2}{*}{Dataset}}  & \multicolumn{1}{c|}{\multirow{2}{*}{Method}}    & \multicolumn{1}{c|}{\multirow{2}{*}{Metric}}   & \multicolumn{4}{c|}{Heterogeneity ($\mathtt{H}$)}                                                                                                               \\ \cline{4-7} 
\multicolumn{1}{|c|}{}                          & \multicolumn{1}{c|}{}                         & \multicolumn{1}{c|}{}                          & \multicolumn{1}{c|}{2}            & \multicolumn{1}{c|}{5}             & \multicolumn{1}{c|}{7}             & \multicolumn{1}{l|}{10}            \\ \hline

                          \multicolumn{7}{|c|}{$100$ Clients}                 \\ \hline
\multicolumn{1}{|c|}{\multirow{6}{*}{\texttt{Adult}}}    & \multicolumn{1}{c|}{\textbf{\ouralgo}}                  & \multicolumn{1}{c|}{\multirow{2}{*}{$\Mu$($\downarrow$)}}    & \multicolumn{1}{c|}{$1.21\pm 0.00$}  & \multicolumn{1}{c|}{$1.31\pm 0.00$}   & \multicolumn{1}{l|}{$1.31\pm 0.00$}   & \multicolumn{1}{l|}{$1.31\pm 0.00$}   \\ \cline{2-2} \cline{4-7} 
\multicolumn{1}{|c|}{}                          & \multicolumn{1}{c|}{\Clus}                  & \multicolumn{1}{c|}{}                          & \multicolumn{4}{c|}{$1.64\pm 0.00$}                                                                                                                 \\ \cline{2-7} 
\multicolumn{1}{|c|}{}                          & \multicolumn{1}{c|}{\textbf{\ouralgo}}                  & \multicolumn{1}{c|}{\multirow{2}{*}{$\Var$($\downarrow$)}} & \multicolumn{1}{c|}{$0.12\pm 0.00$}  & \multicolumn{1}{c|}{$0.06\pm 0.00$} & \multicolumn{1}{l|}{$0.07\pm 0.00$} & \multicolumn{1}{l|}{$0.03\pm 0.00$} \\ \cline{2-2} \cline{4-7} 
\multicolumn{1}{|c|}{}                          & \multicolumn{1}{c|}{\Clus}                  & \multicolumn{1}{c|}{}                          & \multicolumn{4}{c|}{$0.03\pm 0.00$}                                                                                                                \\ \cline{2-7} 
\multicolumn{1}{|c|}{}                          & \multicolumn{1}{c|}{\textbf{\ouralgo}}                  & \multicolumn{1}{c|}{\multirow{2}{*}{$\Max$($\downarrow$)}}   & \multicolumn{1}{c|}{$1.39\pm 0.00$}   & \multicolumn{1}{c|}{$1.45\pm 0.00$}   & \multicolumn{1}{l|}{$1.50\pm 0.00$}   & \multicolumn{1}{l|}{$1.36\pm 0.00$}   \\ \cline{2-2} \cline{4-7} 
\multicolumn{1}{|c|}{}                          & \multicolumn{1}{c|}{\Clus}                  & \multicolumn{1}{c|}{}                          & \multicolumn{4}{c|}{$0.60\pm 0.00$}                                                                                                                \\ \hline
\multicolumn{1}{|c|}{\multirow{6}{*}{\texttt{Bank}}}     & \multicolumn{1}{c|}{\textbf{\ouralgo}}                  & \multicolumn{1}{c|}{\multirow{2}{*}{$\Mu$($\downarrow$)}}    & \multicolumn{1}{c|}{$1.09\pm 0.00$}  & \multicolumn{1}{c|}{$1.54\pm 0.00$}   & \multicolumn{1}{c|}{$1.61\pm 0.00$}   & $1.58\pm 0.00$                        \\ \cline{2-2} \cline{4-7} 
\multicolumn{1}{|c|}{}                          & \multicolumn{1}{c|}{\Clus}                  & \multicolumn{1}{c|}{}                          & \multicolumn{4}{c|}{$0.61\pm 0.00$}                                                                                                                \\ \cline{2-7} 
\multicolumn{1}{|c|}{}                          & \multicolumn{1}{c|}{\textbf{\ouralgo}}                  & \multicolumn{1}{c|}{\multirow{2}{*}{$\Var$($\downarrow$)}} & \multicolumn{1}{c|}{$0.21\pm 0.00$} & \multicolumn{1}{c|}{$0.12\pm 0.00$}  & \multicolumn{1}{c|}{$0.13\pm 0.00$}  & $0.036 \pm 0.00$                    \\ \cline{2-2} \cline{4-7} 
\multicolumn{1}{|c|}{}                          & \multicolumn{1}{l|}{\Clus}                  & \multicolumn{1}{c|}{}                          & \multicolumn{4}{c|}{$0.14\pm 0.00$}                                                                                                                \\ \cline{2-7} 
\multicolumn{1}{|c|}{}                          & \multicolumn{1}{l|}{\textbf{\ouralgo}}                  & \multicolumn{1}{l|}{\multirow{2}{*}{$\Max$($\downarrow$)}}   & \multicolumn{1}{c|}{$1.53\pm 0.00$}  & \multicolumn{1}{c|}{$1.82\pm 0.00$}   & \multicolumn{1}{c|}{$2.42\pm 0.00$}   & $1.64 \pm 0.00$                       \\ \cline{2-2} \cline{4-7} 
\multicolumn{1}{|c|}{}                          & \multicolumn{1}{l|}{\Clus}                  & \multicolumn{1}{l|}{}                          & \multicolumn{4}{c|}{$0.94\pm 0.00$}                                                                                                                \\ \hline

                          \multicolumn{7}{|c|}{$500$ Clients}                \\ \hline
\multicolumn{1}{|c|}{\multirow{6}{*}{\texttt{Adult}}}    & \multicolumn{1}{c|}{\textbf{\ouralgo}}                  & \multicolumn{1}{c|}{\multirow{2}{*}{$\Mu$($\downarrow$)}}    & \multicolumn{1}{c|}{$1.08\pm 0.00$}  & \multicolumn{1}{c|}{$1.28\pm 0.00$}   & \multicolumn{1}{l|}{$1.27\pm 0.00$}   & \multicolumn{1}{l|}{$1.29\pm 0.00$}   \\ \cline{2-2} \cline{4-7} 
\multicolumn{1}{|c|}{}                          & \multicolumn{1}{c|}{\Clus}                  & \multicolumn{1}{c|}{}                          & \multicolumn{4}{c|}{$1.64\pm 0.00$}                                                                                                                 \\ \cline{2-7} 
\multicolumn{1}{|c|}{}                          & \multicolumn{1}{c|}{\textbf{\ouralgo}}                  & \multicolumn{1}{c|}{\multirow{2}{*}{$\Var$($\downarrow$)}} & \multicolumn{1}{c|}{$0.17\pm 0.00$}  & \multicolumn{1}{c|}{$0.10\pm 0.00$} & \multicolumn{1}{l|}{$0.11\pm 0.00$} & \multicolumn{1}{l|}{$0.05\pm 0.00$} \\ \cline{2-2} \cline{4-7} 
\multicolumn{1}{|c|}{}                          & \multicolumn{1}{c|}{\Clus}                  & \multicolumn{1}{c|}{}                          & \multicolumn{4}{c|}{$0.03\pm 0.00$}                                                                                                                \\ \cline{2-7} 
\multicolumn{1}{|c|}{}                          & \multicolumn{1}{c|}{\textbf{\ouralgo}}                  & \multicolumn{1}{c|}{\multirow{2}{*}{$\Max$($\downarrow$)}}   & \multicolumn{1}{c|}{$1.49\pm 0.00$}   & \multicolumn{1}{c|}{$1.51\pm 0.00$}   & \multicolumn{1}{l|}{$1.45\pm 0.00$}   & \multicolumn{1}{l|}{$1.43\pm 0.00$}   \\ \cline{2-2} \cline{4-7} 
\multicolumn{1}{|c|}{}                          & \multicolumn{1}{c|}{\Clus}                  & \multicolumn{1}{c|}{}                          & \multicolumn{4}{c|}{$0.60\pm 0.00$}                                                                                                                \\ \hline
\multicolumn{1}{|c|}{\multirow{6}{*}{\texttt{Bank}}}     & \multicolumn{1}{c|}{\textbf{\ouralgo}}                  & \multicolumn{1}{c|}{\multirow{2}{*}{$\Mu$($\downarrow$)}}    & \multicolumn{1}{c|}{$1.16\pm 0.00$}  & \multicolumn{1}{c|}{$1.55\pm 0.00$}   & \multicolumn{1}{c|}{$1.60\pm 0.00$}   & $1.58\pm 0.00$                        \\ \cline{2-2} \cline{4-7} 
\multicolumn{1}{|c|}{}                          & \multicolumn{1}{c|}{\Clus}                  & \multicolumn{1}{c|}{}                          & \multicolumn{4}{c|}{$0.61\pm 0.00$}                                                                                                                \\ \cline{2-7} 
\multicolumn{1}{|c|}{}                          & \multicolumn{1}{c|}{\textbf{\ouralgo}}                  & \multicolumn{1}{c|}{\multirow{2}{*}{$\Var$($\downarrow$)}} & \multicolumn{1}{c|}{$0.23\pm 0.00$} & \multicolumn{1}{c|}{$0.10\pm 0.00$}  & \multicolumn{1}{c|}{$0.07\pm 0.00$}  & $0.04\pm 0.00$                    \\ \cline{2-2} \cline{4-7} 
\multicolumn{1}{|c|}{}                          & \multicolumn{1}{l|}{\Clus}                  & \multicolumn{1}{c|}{}                          & \multicolumn{4}{c|}{$0.14\pm 0.00$}                                                                                                                \\ \cline{2-7} 
\multicolumn{1}{|c|}{}                          & \multicolumn{1}{l|}{\textbf{\ouralgo}}                  & \multicolumn{1}{l|}{\multirow{2}{*}{$\Max$($\downarrow$)}}   & \multicolumn{1}{c|}{$1.79\pm 0.00$}  & \multicolumn{1}{c|}{$1.87\pm 0.00$}   & \multicolumn{1}{c|}{$1.76\pm 0.00$}   & $1.69 \pm 0.00$                       \\ \cline{2-2} \cline{4-7} 
\multicolumn{1}{|c|}{}                          & \multicolumn{1}{l|}{\Clus}                  & \multicolumn{1}{l|}{}                          & \multicolumn{4}{c|}{$0.94\pm 0.00$}                                                                                                                \\ \hline
\end{tabular}
}
\caption{The table summarizes mean and deviation of evaluation metrics against proposed \ouralgo\ and \Clus\ on $k$-medoids for varying heterogeneity levels on \texttt{Balanced} data split across $100$ and $1000$ clients.}
\label{tab:KmediodEqual100and1000}
\end{table}
  
  \subsubsection{\textbf{Observations for \texttt{Diabetes}}}
  \ouralgo\ has wide gap in $\Mu$ and $\Max$ metrics. Further, it has a lower variance in $\Var$ compared to \sota, exhibiting the efficacy of the approach.  

  \subsubsection{\textbf{Observations for Synthetic (\texttt{Syn})}}
  In unequal data distribution for \texttt{Syn} dataset, it appears like both \MFC\ and \ouralgo\ are quite similar and fairer approaches and only when there is a lot of overlap, i.e., \texttt{Syn-O}, one can see a slight increase in mean cost for \MFC, but as seen throughout the section, the results do not follow the similar trend on other datasets, especially real-world datasets. Thus, this helps us lead to the following overall analysis:

   \noindent \textbf{Overall Insight}: Similar to a balanced (or equal) data split, \ouralgo\ is more likely to be chosen for real-world deployments due to its consistent reliability in terms of the range of cost deviations that different clients may experience.  This is because if some clients face high costs, they may lose incentives to stay in the system and could opt to leave. Nonetheless, the performance of \ouralgo\ is also not subject to heterogeneity levels and number of clients. The key factor driving the performance of the proposed method is its personalized approach through fine-tuning steps.

\subsection{Analysis on Intrinsic Federated Datasets on $k$-means Objective}
This subsection delves into datasets with a pre-captured level of heterogeneity ($\mathtt{H}$). This experiment directly compares the \sota\ with \ouralgo\ on performance metrics.

  \subsubsection{\texttt{WISDM}}
    The results for the \texttt{WISDM} dataset are summarized in Table. \ref{tab:intrisic}. It can be observed that \MFC\ and \ouralgo\ against \kFed\ have a wide gap in $\Var$ and $\Max$, owing to achieving a fair solution as a byproduct or through fine-tuning steps, respectively. The performance of $\Mu$ is also significantly reduced by an order of $10^5$ times, indicating that our \ouralgo\ is the best available fair federated solution.
    

   \subsubsection{\texttt{FEMNIST}}
   The results for the \texttt{FEMNIST} dataset are summarized in Table. \ref{tab:intrisic}. The performance of all methods is quite similar, possibly due to the nature of the dataset. However, \ouralgo\ is considerably close in $\Mu, \Var$, but it has a lower $\Max$ cost a client has to suffer, thus overtaking \sota\ methods.

\renewcommand{\arraystretch}{1.3}
\begin{table}[h!]
\scalebox{0.6}{ 
\begin{tabular}{|ccccccc|}
\hline
\multicolumn{1}{|c|}{\multirow{2}{*}{Dataset}}    & \multicolumn{1}{c|}{\multirow{2}{*}{Method}} & \multicolumn{1}{c|}{\multirow{2}{*}{Metric}}   & \multicolumn{4}{c|}{Heterogeneity ($\mathtt{H}$)}                                                                                                               \\ \cline{4-7} 
\multicolumn{1}{|c|}{}                          & \multicolumn{1}{c|}{}                         & \multicolumn{1}{c|}{}                          & \multicolumn{1}{c|}{2}            & \multicolumn{1}{c|}{5}             & \multicolumn{1}{c|}{7}             & \multicolumn{1}{l|}{10}            \\ \hline

\multicolumn{7}{|c|}{$100$ Clients}         \\ \hline
\multicolumn{1}{|c|}{\multirow{6}{*}{\texttt{Adult}}}    & \multicolumn{1}{l|}{\textbf{\ouralgo}}                  & \multicolumn{1}{c|}{\multirow{2}{*}{$\Mu$($\downarrow$)}}    & \multicolumn{1}{c|}{$1.17\pm 0.00$}  & \multicolumn{1}{c|}{$1.31\pm 0.00$}   & \multicolumn{1}{l|}{$1.32\pm 0.00$}   & \multicolumn{1}{l|}{$1.29\pm 0.00$}   \\ \cline{2-2} \cline{4-7} 
\multicolumn{1}{|c|}{}               & \multicolumn{1}{l|}{\Clus}                             & \multicolumn{1}{c|}{}                          & \multicolumn{4}{c|}{$1.64\pm 0.00$}                                                                                                                 \\ \cline{2-7} 
\multicolumn{1}{|c|}{}               & \multicolumn{1}{l|}{\textbf{\ouralgo}}                           & \multicolumn{1}{c|}{\multirow{2}{*}{$\Var$($\downarrow$)}} & \multicolumn{1}{c|}{$0.11\pm 0.00$}  & \multicolumn{1}{c|}{$0.11\pm 0.00$} & \multicolumn{1}{l|}{$0.07\pm 0.00$} & \multicolumn{1}{l|}{$0.16\pm 0.00$} \\ \cline{2-2} \cline{4-7} 
\multicolumn{1}{|c|}{}                    & \multicolumn{1}{l|}{\Clus}                         & \multicolumn{1}{c|}{}                          & \multicolumn{4}{c|}{$0.03\pm 0.00$}                                                                                                                \\ \cline{2-7} 
\multicolumn{1}{|c|}{}                        & \multicolumn{1}{l|}{\textbf{\ouralgo}}               & \multicolumn{1}{c|}{\multirow{2}{*}{$\Max$($\downarrow$)}}   & \multicolumn{1}{c|}{$1.46\pm 0.00$}   & \multicolumn{1}{c|}{$1.48\pm 0.00$}   & \multicolumn{1}{l|}{$1.47\pm 0.00$}   & \multicolumn{1}{l|}{$1.65\pm 0.00$}   \\ \cline{2-2} \cline{4-7} 
\multicolumn{1}{|c|}{}                     & \multicolumn{1}{l|}{\Clus}                  & \multicolumn{1}{c|}{}                          & \multicolumn{4}{c|}{$0.60\pm 0.00$}                                                                                                                \\ \hline
\multicolumn{1}{|c|}{\multirow{6}{*}{\texttt{Bank}}}       & \multicolumn{1}{l|}{\textbf{\ouralgo}}               & \multicolumn{1}{c|}{\multirow{2}{*}{$\Mu$($\downarrow$)}}    & \multicolumn{1}{c|}{$1.21\pm 0.00$}  & \multicolumn{1}{c|}{$1.50\pm 0.00$}   & \multicolumn{1}{c|}{$1.63\pm 0.00$}   & $1.71\pm 0.00$                        \\ \cline{2-2} \cline{4-7} 
\multicolumn{1}{|c|}{}                  & \multicolumn{1}{l|}{\Clus}                            & \multicolumn{1}{c|}{}                          & \multicolumn{4}{c|}{$0.61\pm 0.00$}                                                                                                                \\ \cline{2-7} 
\multicolumn{1}{|c|}{}                           & \multicolumn{1}{l|}{\textbf{\ouralgo}}           & \multicolumn{1}{c|}{\multirow{2}{*}{$\Var$($\downarrow$)}} & \multicolumn{1}{c|}{$0.14\pm 0.00$} & \multicolumn{1}{c|}{$0.11\pm 0.00$}  & \multicolumn{1}{c|}{$0.06\pm 0.00$}  & $0.10\pm 0.00$                    \\ \cline{2-2} \cline{4-7} 
\multicolumn{1}{|c|}{}                   & \multicolumn{1}{l|}{\Clus}                        & \multicolumn{1}{c|}{}                          & \multicolumn{4}{c|}{$0.14\pm 0.00$}                                                                                                                \\ \cline{2-7} 
\multicolumn{1}{|c|}{}                          & \multicolumn{1}{l|}{\textbf{\ouralgo}}               & \multicolumn{1}{l|}{\multirow{2}{*}{$\Max$($\downarrow$)}}   & \multicolumn{1}{c|}{$1.55\pm 0.00$}  & \multicolumn{1}{c|}{$1.80\pm 0.00$}   & \multicolumn{1}{c|}{$1.75\pm 0.00$}   & $2.04 \pm 0.00$                       \\ \cline{2-2} \cline{4-7} 
\multicolumn{1}{|c|}{}                  & \multicolumn{1}{l|}{\Clus}                          & \multicolumn{1}{l|}{}                          & \multicolumn{4}{c|}{$0.94\pm 0.00$}                                                                                                              \\ \hline

                          \multicolumn{7}{|c|}{$500$ Clients}                                                                                                              \\ \hline

\multicolumn{1}{|c|}{\multirow{6}{*}{\texttt{Adult}}}    & \multicolumn{1}{l|}{\textbf{\ouralgo}}                  & \multicolumn{1}{c|}{\multirow{2}{*}{$\Mu$($\downarrow$)}}    & \multicolumn{1}{c|}{$1.17\pm 0.00$}  & \multicolumn{1}{c|}{$1.30\pm 0.00$}   & \multicolumn{1}{l|}{$1.29\pm 0.00$}   & \multicolumn{1}{l|}{$1.27\pm 0.00$}   \\ \cline{2-2} \cline{4-7} 
\multicolumn{1}{|c|}{}               & \multicolumn{1}{l|}{\Clus}                             & \multicolumn{1}{c|}{}                          & \multicolumn{4}{c|}{$1.64\pm 0.00$}                                                                                                                 \\ \cline{2-7} 
\multicolumn{1}{|c|}{}               & \multicolumn{1}{l|}{\textbf{\ouralgo}}                           & \multicolumn{1}{c|}{\multirow{2}{*}{$\Var$($\downarrow$)}} & \multicolumn{1}{c|}{$0.11\pm 0.00$}  & \multicolumn{1}{c|}{$0.09\pm 0.00$} & \multicolumn{1}{l|}{$0.07\pm 0.00$} & \multicolumn{1}{l|}{$0.14\pm 0.00$} \\ \cline{2-2} \cline{4-7} 
\multicolumn{1}{|c|}{}                    & \multicolumn{1}{l|}{\Clus}                         & \multicolumn{1}{c|}{}                          & \multicolumn{4}{c|}{$0.03\pm 0.00$}                                                                                                                \\ \cline{2-7} 
\multicolumn{1}{|c|}{}                        & \multicolumn{1}{l|}{\textbf{\ouralgo}}               & \multicolumn{1}{c|}{\multirow{2}{*}{$\Max$($\downarrow$)}}   & \multicolumn{1}{c|}{$1.38\pm 0.00$}   & \multicolumn{1}{c|}{$1.50\pm 0.00$}   & \multicolumn{1}{l|}{$1.48\pm 0.00$}   & \multicolumn{1}{l|}{$1.65\pm 0.00$}   \\ \cline{2-2} \cline{4-7} 
\multicolumn{1}{|c|}{}                     & \multicolumn{1}{l|}{\Clus}                  & \multicolumn{1}{c|}{}                          & \multicolumn{4}{c|}{$0.60\pm 0.00$}                                                                                                                \\ \hline
\multicolumn{1}{|c|}{\multirow{6}{*}{\texttt{Bank}}}       & \multicolumn{1}{l|}{\textbf{\ouralgo}}               & \multicolumn{1}{c|}{\multirow{2}{*}{$\Mu$ ($\downarrow$)}}    & \multicolumn{1}{c|}{$1.17\pm 0.00$}  & \multicolumn{1}{c|}{$1.53\pm 0.00$}   & \multicolumn{1}{c|}{$1.63\pm 0.00$}   & $1.54\pm 0.00$                        \\ \cline{2-2} \cline{4-7} 
\multicolumn{1}{|c|}{}                  & \multicolumn{1}{l|}{\Clus}                            & \multicolumn{1}{c|}{}                          & \multicolumn{4}{c|}{$0.61\pm 0.00$}                                                                                                                \\ \cline{2-7} 
\multicolumn{1}{|c|}{}                           & \multicolumn{1}{l|}{\textbf{\ouralgo}}           & \multicolumn{1}{c|}{\multirow{2}{*}{$\Var$ ($\downarrow$)}} & \multicolumn{1}{c|}{$0.18\pm 0.00$} & \multicolumn{1}{c|}{$0.12\pm 0.00$}  & \multicolumn{1}{c|}{$0.08\pm 0.00$}  & $0.19\pm 0.00$                    \\ \cline{2-2} \cline{4-7} 
\multicolumn{1}{|c|}{}                   & \multicolumn{1}{l|}{\Clus}                        & \multicolumn{1}{c|}{}                          & \multicolumn{4}{c|}{$0.14\pm 0.00$}                                                                                                                \\ \cline{2-7} 
\multicolumn{1}{|c|}{}                          & \multicolumn{1}{l|}{\textbf{\ouralgo}}               & \multicolumn{1}{l|}{\multirow{2}{*}{$\Max$($\downarrow$)}}   & \multicolumn{1}{c|}{$1.75\pm 0.00$}  & \multicolumn{1}{c|}{$1.90\pm 0.00$}   & \multicolumn{1}{c|}{$1.79\pm 0.00$}   & $2.11 \pm 0.00$                       \\ \cline{2-2} \cline{4-7} 
\multicolumn{1}{|c|}{}                  & \multicolumn{1}{l|}{\Clus}                          & \multicolumn{1}{l|}{}                          & \multicolumn{4}{c|}{$0.94\pm 0.00$}                                                                                                              \\ \hline

\end{tabular}
}
\caption{The table summarizes mean and deviation of evaluation metrics against proposed \ouralgo\ and \Clus\ on $k$-medoids for varying heterogeneity levels on \texttt{Unequal} data split across $100$ and $500$ clients.}
\label{tab:KmediodUnequal100and500}
\end{table}

   \subsection{Analysis on different Dataset for $k$-mediod Objective}

   In $k$-mediod objective, we limit the comparison only to \Clus\ because other baselines are not intrinsically designed to handle objectives except $k$-means. The results for \texttt{Balanced}  data split for real-world non-federated datasets are reported in Table \ref{tab:KmediodEqual100and1000} for $100, 1000$ clients. We can clearly observe that the mean per point cost ($\Mu$), and correspondingly max cost ($\Max$) for \ouralgo\ is quite close to centralized clustering, showcasing that \ouralgo\  efficiently captures the centers in a federated setting.  Furthermore, it not only showcases its efficacy in cost but also in the fairness metric, i.e., $\Var$, which is also considerably low across clients, thus resulting in a fair aka personalized clustering.

\begin{table}[h!]
\centering
\scalebox{0.7}{
\begin{tabular}{|c|l|c|c|c|c|c|c|c|}
\hline
Dataset     & Method   & Metric    & Value   &  & Dataset  & Value\\ \hline

\multirow{6}{*}{\texttt{WISDM}} & \textbf{\ouralgo}  & \multirow{2}{*}{$\Mu$($\downarrow$)}      & $1.08\times10^{13} \pm 0.00$ & & \multirow{6}{*}{\texttt{FEMNIST}}     & $6.87 \pm 0.00$    \\ \cline{2-2} \cline{4-4} \cline{7-7}

   & \Clus &   & (refer caption) &  & &  $2.91 \pm 0.00$                                  \\ \cline{2-4} \cline{7-7}
                         
  & \textbf{\ouralgo}  & \multirow{2}{*}{$\Var$($\downarrow$)}   & $2.64\times10^{13} \pm 0.00$   &   & & $0.96 \pm 0.00$            \\ \cline{2-2} \cline{4-4} \cline{7-7}

  & \Clus &  & (refer caption)   & & & $0.65 \pm 0.00$   \\ \cline{2-4} \cline{7-7}

 & \textbf{\ouralgo}  & \multirow{2}{*}{$\Max$($\downarrow$)}  & $1.30\times10^{14} \pm 0.00$  & & & $9.69 \pm 0.00$    \\ \cline{2-2} \cline{4-4} \cline{7-7}
&  \Clus &   & (refer caption) & & & $4.32 \pm 0.00$   \\ \hline
\end{tabular}
}
\caption{The table summarizes mean and deviation of evaluation metrics against proposed \ouralgo\ and \Clus\ on $k$-medoids for Intrinsic datasets. The \texttt{WISDM} dataset is not evaluated on \Clus\ due to $8$ terabytes of main memory requirements.}
\label{tab:KmediodIntrin}
\end{table}

   The results for \texttt{Unequal}  data split are reported in Table \ref{tab:KmediodUnequal100and500} for 100 and 500 clients.  In this setting also, the observations are similar to the previous setting, showcasing the benefit of \ouralgo\ in both scenarios. The results for intrinsic federated datasets, namely \texttt{WISDM} and \texttt{FEMNIST}, are reported in Table \ref{tab:KmediodIntrin}.  We do not report the \Clus\ results for the \texttt{WISDM} dataset as the main memory requirement for such a large dataset is nearly $8$TB and thus can only be processed in streaming or federated (distributed) settings.

\section{Conclusion and Future Directions}
\label{sec:concl}
In this paper, we focus on solving the problem of handling unlabelled data in a federated setting. We propose a first-of-its-kind personalized data clustering algorithm, \ouralgo\, which operates in three sub-phases within a single round of to-and-fro communication between the server and clients. The algorithm does not require a small amount of labelled data (or synthetic labels) and is a completely unsupervised method without any significant computational overheads. Furthermore, through a series of rigorous experimental analysis, we observe that \ouralgo's performance does not suffer across varying levels of heterogeneity and clients, showcasing its data distribution independence. Additionally, it achieves a lower mean per-point objective cost in most scenarios compared to state-of-the-art (\sota) methods while ensuring small deviations in cost across clients (fairness). The maximum cost any client incurs using \ouralgo\ is significantly lower than \sota\ methods in almost all settings, enabling clients to have personalized models and incentivising them to continue contributing to the federated setup. Moreover, the method is reliable and applicable to any finite $\ell$-norm objectives, including $k$-means and $k$-medoids. However, the current version of the algorithm assumes that all clients are benign and not malicious. Studying a robust personalized data clustering method in the presence of malicious clients (\cite{gupta2023performance}) and noisy data (\cite{rehman2022divide}) is a potential direction for future study. Additionally, an interesting future direction is investigating scenarios where clients might strategically report their features, thus hampering the quality of generated local centers (\cite{stoica2018strategic}). Since clients in federated learning can join and leave the system, analyzing the effects of unlearning in clustering is another promising direction (\cite{pan2022machine}). We leave these as future works.

\section*{Acknowledgement}
The authors thank the Prime Minister Research Fellowship for generously funding Shivam Gupta (ID: 2901481) for this work. We would also like to acknowledge the support of the Science and Engineering Research Board (Anusandhan National Research Foundation), Government of India under grant number CRG/2022/004980. 

\section*{Declaration of Interest Statement} No potential competing interest was reported by the authors.  The paper is the authors' own original work, which has not been previously published elsewhere. The paper is not currently being considered for publication elsewhere. The paper reflects the author's own research and analysis truthfully and completely. The paper properly credits the meaningful contributions of co-authors and co-researchers.

\bibliographystyle{elsarticle-harv} 
\bibliography{reference}



\end{document}